\documentclass[10pt,twocolumn,letterpaper]{article}

\usepackage[pagenumbers]{wacv} 

\usepackage{graphicx}
\usepackage{amsmath}
\usepackage{amssymb}
\usepackage{booktabs}

\usepackage[pagebackref,breaklinks,colorlinks]{hyperref}

\usepackage[capitalize]{cleveref}
\crefname{section}{Sec.}{Secs.}
\Crefname{section}{Section}{Sections}
\Crefname{table}{Table}{Tables}
\crefname{table}{Tab.}{Tabs.}

\newcommand{\method}{\textsc{ReEdit}\xspace}
\newcommand{\xhdr}[1]{\vspace{0em}\noindent{{\bf #1.}}}

\newif\ifcomments
\commentstrue
\ifcomments
    \providecommand{\abhinav}[2][]{{\protect\color{magenta}{[\textbf{A}:\textbf{#1} #2]}}}
    \providecommand{\tarun}[2][]{{\protect\color{red}{[\textbf{T}:\textbf{#1} #2]}}}
    \providecommand{\silky}[2][]{{\protect\color{violet}{[\textbf{S}:\textbf{#1} #2]}}}
\else
    \providecommand{\abhinav}[2][]{}
     \providecommand{\tarun}[2][]{}
     \providecommand{\silky}[2][]{}
\fi

\def\wacvPaperID{2609} 
\def\confName{WACV}
\def\confYear{2025}

\begin{document}

\title{ReEdit: Multimodal Exemplar-Based Image Editing with Diffusion Models}


\author{
Ashutosh Srivastava$^{1}\thanks{Equal Contribution} ~~\thanks{Work done during internship at Adobe MDSR}$ 
\qquad
Tarun Ram Menta$^{2*}$
\qquad
Abhinav Java$^{3*}\thanks{Work done while at Adobe}$
\qquad
Avadhoot Jadhav$^{4\dag}$\\
Silky Singh$^{5\ddag}$
\qquad
Surgan Jandial$^{6\ddag}$
\qquad
Balaji Krishnamurthy$^{2}$
 \\
$^{1}$ Indian Institute of Technology, Roorkee \quad $^{2}$ Adobe MDSR \quad $^{3}$ Microsoft Research \\
\quad $^{4}$ Indian Institute of Technology, Bombay \quad $^{5}$ Stanford University \quad $^{6}$ Carnegie Mellon University
}

\maketitle


\begin{abstract}

Modern Text-to-Image (T2I) Diffusion models have revolutionized image editing by enabling the generation of high-quality photorealistic images. While the de facto method for performing edits with T2I models is through text instructions, this approach non-trivial due to the complex many-to-many mapping between natural language and images. In this work, we address \textit{exemplar-based image editing} -- the task of transferring an edit from an exemplar pair to a content image(s). We propose \method, a modular and efficient end-to-end framework that captures edits in both text and image modalities while ensuring the fidelity of the edited image. We validate the effectiveness of \method through extensive comparisons with state-of-the-art baselines and sensitivity analyses of key design choices. Our results demonstrate that \method consistently outperforms contemporary approaches both qualitatively and quantitatively. Additionally, \method boasts high practical applicability, as it does not require any task-specific optimization and is four times faster than the next best baseline.

\end{abstract}

\section{Introduction}
\label{sec:intro}


Image editing~\cite{karras2019style, alaluf2022third, liu2022self, gal2022stylegan, nichol2021glide} is a rapidly growing research area, with a wide range of practical applicability in domains like multimedia, cinema, advertising, etc. Recent advancements in text-based diffusion models~\cite{ho2020denoising, saharia2022photorealistic, nichol2021glide, podell2023sdxl} have accelerated the progress in the field of image editing, yet diffusion models remain limited in their practical viability to real world applications. For example, if a practitioner is making detailed edits—such as transforming a scene from daytime to nighttime—and wants to apply the same adjustments to multiple images, they would face a considerable challenge, since crafting each image individually can be time consuming. In such cases, simple textual prompts might not be sufficient to achieve the desired consistency and efficiency.

Notably, an ideal editing application should be \textbf{fast}, have the ability to understand the exact \textbf{user intent} and produce \textbf{high fidelity} outputs. Most existing work in this domain leverages textual descriptions to perform image editing~\cite{brooks2023instructpix2pix, tumanyan2023plug, hertz2022prompt, kawar2023imagic, mokady2023null, huang2023reversion,  kim2022diffusionclip, parmar2023zero}, however, text is inherently limited in its ability to adequately describe edits. These challenges motivate us to focus on a relatively unexplored field of \emph{exemplar based image editing}. This formulation is motivated by `visual prompting' proposed in~\cite{bar2022visual}. 

Existing works in this area typically optimize a text embedding during inference to capture each edit~\cite{nguyen2024visual, jeong2024visual} which is time taking. Other methods like~\cite{hertzmann2023image, vsubrtova2023diffusion} utilize sophisticated models trained specifically for the task of editing like InstructPix2Pix~\cite{brooks2023instructpix2pix} (IP2P), which requires a large labelled training dataset. These datasets can be extremely difficult to obtain due to the nature of the problem. Further, recent approaches like VISII~\cite{nguyen2024visual} can only capture a limited type of edits (performs well only for \emph{global style transfer} type edits) as a result of the way its text embedding is optimized. 

Unlike existing approaches, we propose an efficient end-to-end optimization-free framework for exemplar based image editing - \textbf{\method}. The proposed framework consists of three primarily components - \emph{first} we capture the edit from the exemplar in the image embedding space using pretrained adapter modules~\cite{rombach2022high}, \emph{second}, we capture the edit in natural language by incorporating multimodal VLMs like~\cite{liu2024llava} capable of detailed reasoning, and \emph{last} we ensure that the content and structure of the test image is maintained and only the relevant parts are edited by conditioning the image generator on the features and self attention maps~\cite{tumanyan2023plug} of the test image. \textbf{Overall, none of the components of our approach are explicitly trained for image editing, do not require inference time optimization, and easily generalize to a wider variety of edit types while being independent of the base diffusion model and prove to be extremely efficient}. To summarize, the contributions of our work are listed below:

\noindent{\xhdr{1}} We propose an inference-time approach for \emph{exemplar-based image editing} that does not require finetuning or optimizing any part of the pipeline. Compared to the most optimal baseline, the runtime of our method is $\sim$4x faster.

\noindent{\xhdr{2}} We collate a dataset of ~1500 exemplar pairs ($x, x_{\text{edit}}$), and corresponding test images with ground truth $(y, y_\text{edit})$, covering a wide range of edits. Due to a lack of standardized datasets, our dataset paves towards a standardized evaluation of \emph{exemplar-based image editing} approaches.

\noindent{\xhdr{3}} Our rigorous qualitative and quantitative analysis shows that our method performs well on a variety of edits while preserving the structure of the original image. These observations are corroborated by significant improvements in quantitative scores over baselines. We plan on open sourcing the dataset and code.

\section{Related Work}
\label{sec:related}
\noindent \textbf{Diffusion Models.} Prior to diffusion models, GANs~\cite{goodfellow2020generative, zhang2017stackgan, zhu2016generative} were the de-facto generative models used for (conditional) image synthesis and editing. However, training GAN networks is prone to instability and mode collapse, among many issues. Recently, large-scale text-to-image generative models~\cite{saharia2022photorealistic, ramesh2022hierarchical, ramesh2021zero, yu2022scaling, ding2021cogview, gafni2022make, rombach2022high} have benefitted from superior model architectures~\cite{vaswani2017attention} and large-scale training data available on the internet. Of particular interest is diffusion models~\cite{nichol2021glide, rombach2022high, saharia2022photorealistic, podell2023sdxl, ramesh2021zero, song2020denoising, ho2020denoising}, that are trained to denoise random gaussian noise resulting in high-fidelity and highly diverse images. These models are typically trained on millions of text-image pairs. In this work, we use a pretrained Stable Diffusion~\cite{rombach2022high} model which operates in the latent space instead of the image pixel space.

\vspace{0.1cm}
\noindent \textbf{Multimodal Vision-Language Models (VLMs).} Multimodal VLMs~\cite{li2022blip, radford2021learning, liu2024visual, liu2024llava, liu2024improved, singh2022flava} have the remarkable capability to understand and process both texts and images. Two particularly useful works fall in the scope of this paper: CLIP~\cite{radford2021learning} and LLaVA~\cite{liu2024improved, liu2024llava, liu2024visual}. CLIP represents both images and texts in a shared embedding space. It was trained on 400M image-text pairs in a contrastive manner -- maximizing the similarity between related image-text embeddings, while minimizing the similarity between unrelated image-text embeddings. LLaVA combines a visual encoder with Vicuna~\cite{vicuna2023} to provide powerful language and visual comprehension capabilities. It has impressive capacity to follow user instructions based on visual cues.

\vspace{0.1cm}
\noindent \textbf{Text-based Image Editing.} Diffusion models, with their impressive generative capabilites, have also been adapted for image editing~\cite{zhang2023adding, mokady2023null, gal2022image, zhang2023sine, huang2023reversion, ruiz2023dreambooth, kim2022diffusionclip, parmar2023zero, tumanyan2023plug, yang2023object, kwon2022diffusion, couairon2022diffedit}. Multimodal models like CLIP~\cite{radford2021learning}, and cross-attention mechanisms~\cite{vaswani2017attention} have enabled conditioning a diffusion model to directly edit an image with a text input~\cite{avrahami2022blended, nichol2021glide}. SDEdit~\cite{meng2021sdedit} takes an image as input along with a user guide, and subsequently denoises it using SDE prior to increase its realism. Other related works~\cite{choi2021ilvr, song2020score} guide the generative process conditioned on some user input, for e.g., a reference image. Imagic~\cite{kawar2023imagic} finetunes a diffusion model on a single image to perform image editing. Prompt-to-prompt~\cite{hertz2022prompt} attempts to edit an image while preserving its structure by modifying the attention maps in a pretrained diffusion model. Similarly, pix2pix-zero~\cite{parmar2023zero} preserves the content and structure of the original image while editing via cross-attention guidance. Instruct-pix2pix~\cite{brooks2023instructpix2pix} first collected a huge dataset of (image, edit text, edited image) triplets, and trained a diffusion model to follow edit instructions provided by a user. Plug-and-Play~\cite{tumanyan2023plug} aims to preserve the semantic layout of an image during an edit by manipulating spatial features and self-attention in a pretrained text-to-image diffusion model. Although these approaches produce plausible edits to an image, there still exist limitations where either the edit instruction/text is completely ignored, or the structure of the original image is drastically modified. Additionally, our work differs from this line of work since we get rid of text-based instructions altogether.

\vspace{0.1cm}
\noindent \textbf{Exemplar-based Image Editing.} In the field of Computer Vision, `visual prompting' was first proposed in \cite{bar2022visual}. Later works~\cite{wang2023images, wang2023seggpt} build a generalist model based on visual in-context learning to solve multiple vision tasks, including segmentation. Exemplar-based editing methods~\cite{jeong2024visual, yang2023paint, nguyen2024visual, hertzmann2023image, vsubrtova2023diffusion} are an extension of "visual prompting", where the focus is to edit an image conditioned on a visual input, called \textit{exemplar}. This can include insertion of the exemplar object in a given image to produce a photo-realistic output as in Paint-by-Example~\cite{yang2023paint}, or transfer of overall style from an exemplar image to a given image~\cite{jeong2024visual}. The concept of \textit{image analogies} was proposed in~\cite{hertzmann2023image} and later used in~\cite{liao2017visual} for visual attribute transfer from one image to another, for e.g., color, tone, texture, style. It has also been extended for example-based editing using diffusion models~\cite{vsubrtova2023diffusion}. The present work is closest to VISII~\cite{nguyen2024visual} and ImageBrush~\cite{yang2024imagebrush} -- both explore the idea of using exemplar pair as visual instruction for image editing. Differently from our work, VISII~\cite{nguyen2024visual} relies on optimization-based inversion for capturing the edit in CLIP~\cite{radford2021learning} text space, while ImageBrush~\cite{yang2024imagebrush} benefits from training a diffusion model on the revised task of conditional inpainting. At the same time, we achieve superior edit outputs without any optimization or training required.

\section{Methodology}
\label{sec:method}

\begin{figure*}[h!]
    \centering
    \includegraphics[width=0.9\textwidth]{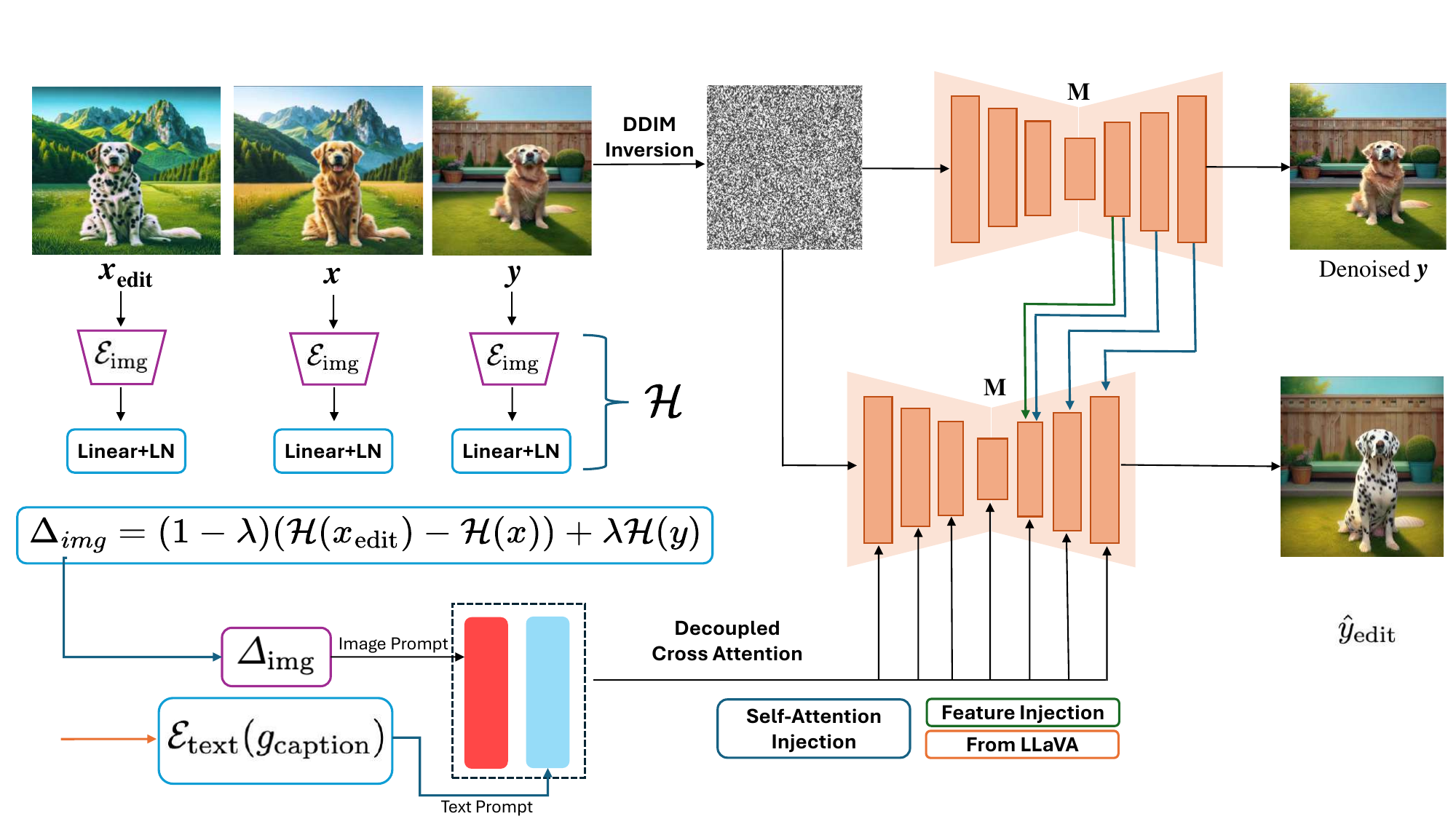}
    \small{\caption{Overview of our framework \method. For details, please refer Section~\ref{sec:method}.\label{fig:overview}}}
\end{figure*}

In this section, we first introduce some preliminaries and describe the notation. We then, introduce our proposed framework, \method which comprises two key steps: \textbf{(a)} capturing the edit ($g$) from the given exemplar pair in both text and image space, followed by \textbf{(b)} conditioning the diffusion model ($M$) to apply this edit on a test image ($y$) without any optimization. The overview of our framework is illustrated in Fig.~\ref{fig:overview}.

\noindent{\xhdr{Problem Setting and Notation}} 
Given a pair of exemplar images $(x, x_{\text{edit}})$, where $x$ denotes the original image, and $x_{\text{edit}}$ denotes the edited image respectively. Our objective is to capture the edit (say $g$, such that $x_{\text{edit}} = g(x)$), and apply the \emph{same edit} ($g$) on a test image $y$ to obtain the corresponding edited image $\hat{y}_{\text{edit}}$. Let $M(\theta)$ denote a pretrained diffusion model (here, SD1.5~\cite{rombach2022high}) parameterized by $\theta$, where $\theta$ remains frozen. And let $\mathcal{E}_{\text{img}}$ and $\mathcal{E}_{\text{text}}$ denote pretrained CLIP image and text encoders respectively, with a hidden dimension of $768$.

\noindent{\xhdr{Background}} Recent work~\cite{ye2023ip-adapter} proposes to utilize simple adapter modules to generate high quality images with images as prompts. Unlike typical T2I models whose cross attention parameters are only conditioned on text-embeddings, IP-Adapter~\cite{ye2023ip-adapter} adds newly initialized linear and cross attention layers and finetunes these additional parameters ($\sim$22M), which directly allow the introduction of image embeddings to pretrained T2I models. As motivated in Sec~\ref{sec:intro}, text alone often falls short in capturing the edit from exemplar pairs, so we propose a strategy that enables us to capture the edits from the exemplar pairs both in the image space (using simple adapters) and in the text space.

\subsection{Capturing Edits from exemplars}
We posit that \emph{textual descriptions are necessary but not sufficient} to generate $\hat{y}_{\text{edit}}$ from ($x, x_{\text{edit}}, y$). Consequently, we capture edits in both \textit{text} and \textit{image} space. 

\xhdr{Edits in natural language} Firstly, we leverage a multimodal VLM (LLaVA~\cite{liu2024improved, liu2024llava, liu2024visual}) to verbalize the edits in the exemplar pair ($x$, $x_{\text{edit}}$). We pass these images as a grid, along with a detailed prompt $p_1$ that instructs LLaVA to generate a comprehensive description of the edits, denoted by $g_{\text{text}}$. Additionally, to provide the context of the test image $y$, we curate another prompt $p_2$ instructing LLaVA to describe $\hat{y}_{\text{edit}}$ in text after applying the edit $g_{\text{text}}$ on $y$. As a result, we obtain a final text description of $\hat{y}_{\text{edit}}$, denoted by $g_{\text{caption}}$. To reduce verbosity and token length, we limit $g_{\text{caption}}$ to $40$ words. Refer to Appendix~\ref{sec:addn-llava} for the exact prompts $p_1$ and $p_2$, and a an overview of the caption generation process. 

\xhdr{Edits in image space} Natural language can't capture the specific style, intensity, hue, saturation, exact shape, or other detailed attributes of the objects in the image. Therefore, we also capture the edits from $(x, x_{\text{edit}})$ and the original image ($y$) directly in CLIP's embedding space. 
Specifically, we apply a pretrained linear layer and layer norm~\cite{ye2023ip-adapter} on the clip embeddings of $x$, $x_{\text{edit}}$ to make the embeddings compatible with $M$. The edit is captured as follows - $\Delta_{img} =  \lambda(\mathcal{H}(x_{\text{edit}}) -  \mathcal{H}(x)) +  (1-\lambda)\mathcal{H}(y)$; where $\mathcal{H}(x) = \operatorname{LN}(\operatorname{Lin}(\mathcal{E}_{\text{img}}(x)))$ and $\operatorname{LN}$, $\operatorname{Lin}$ are the layer norm and linear projection operators respectively. The \textbf{edit weight} slider is denoted by $\lambda$. $\lambda$ weighs the contributions of the edit and the target image while generating the final result $y_\text{edit}$. 
Our final edit embedding is hence given by the pair $g := (\Delta_{\text{img}},~ \mathcal{E}_{\text{text}}(g_{\text{caption}}))$. Both the image and text conditioning in $g$ work in tandem to provide nuanced  guidance for precise edits. As shown in Fig~\ref{fig:overview}, the edit embeddings in $g$ are processed by their respective decoupled cross attention parameters and propagated through $M$ to generate the final image.


\vspace{0.15cm}
\subsection{Conditioning Stable Diffusion on $(g, y)$}
A crucial requirement of image editing approaches is that they preserve the content and structure of the original image in the edited output. Thus, we aim to condition $M$ on $g$ such that only the relevant parts of $y$ are edited, while the rest of the image remains intact. To achieve this, we introduce the approach of attention and feature injection motivated by~\cite{tumanyan2023plug}. Specifically, we invert $y$ using DDIM inversion~\cite{song2020denoising}, and run vanilla denoising on the inverted noise ($y_{\text{noise}}$). The features (say $f$) and attention matrices ($Q, K$) of the upsampling blocks while sampling the noise unconditionally contain the overall structure information for $y$~\cite{tumanyan2023plug}. In a parallel run, starting with $y_{\text{noise}}$, we condition the denoising process on edit $g$ (through cross-attention), inject the features ($f$) at the fourth layer and modify the keys and queries ($Q, K$) in the self-attention layers from layers $4$ to $11$ of $M$ to obtain $\hat{y}_{\text{edit}}$.

\section{Dataset Creation}
\label{sec:dataset}


    \begin{table}[b]  
        \centering
        \begin{tabular}{@{}cc@{}}
        \toprule
        \setlength{\tabcolsep}{2pt}
        \textbf{Type of Edit} & \textbf{Number of Examples} \\ \midrule
        Global Style Transfer & 428 \\
        Background Change & 212 \\
        Localized Style Transfer & 290  \\
        Object Replacement & 366 \\
        Motion Edit & 14 \\
        Object Insertion & 164 \\ \midrule
        \textbf{Total} & \textbf{1474} \\ \bottomrule
        \end{tabular}
        \caption{Summary and statistics of the types of edits in the evaluation dataset. Special care was taken to ensure diversity of edit categories.}
        \label{tab:edit-types}
    \end{table}
    \begin{figure}[b]
        \centering
        \includegraphics[width=0.95\linewidth]{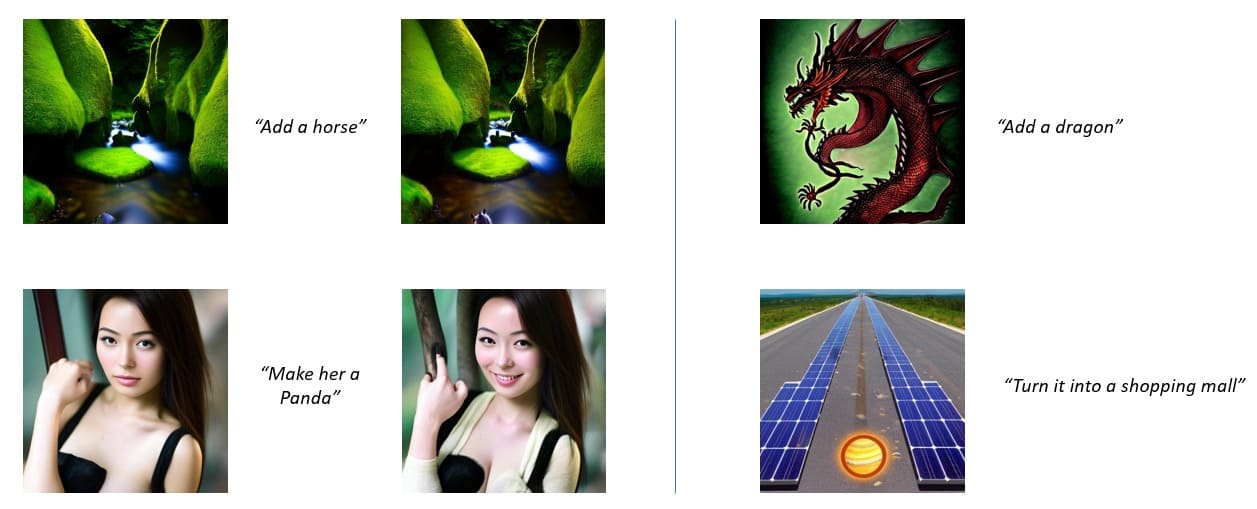}
        \caption{Examples of ambiguous samples in the InstructPix2Pix dataset, motivating the need for manual curation. Additional examples can be found in Appendix~\ref{sec:ip2p-failure}}
        \label{fig:ip2p-failure}
    \end{figure}

Our method is an inference-time approach, and is directly applicable to an arbitrary set of $(x, x_{\text{edit}}, y)$ images. However, there are no existing evaluation datasets for exemplar-based image editing in the current literature. Hence, we curate a dataset from the existing image editing dataset. Specifically, the exemplar pairs are taken from the InstructPix2Pix dataset. This is a dataset for text-based image editing, containing $450,000$ $(x, x_{\text{edit}}, g_{\text{edit}})$, where $x_{\text{edit}}$ is the image obtained after applying the edit instruction $g_{\text{edit}}$ on input image $x$. This dataset was generated by applying Prompt-to-Prompt~\cite{hertz2022prompt} on a Stable Diffusion model. We found two common issues with this dataset - i) the edit pair $(x, x_{\text{edit}}$ did not adhere to the edit instruction $g_{\text{edit}}$, and ii) The edit instruction $g_{\text{edit}}$ did not apply to the input image $x$. Refer to Fig.~\ref{fig:ip2p-failure} for examples of these failure cases. As a result, we carefully curate a dataset of $(x, x_{\text{edit}}, y, y_{\text{edit}})$ where $x$, $y$, and the corresponding edited images are taken from IP2P samples with the same edit instruction $g_{\text{edit}}$. Through visual inspection, we manually ensure that the two aforementioned issues do not creep into our dataset, resulting in a high-quality dataset of $\sim$~1500 samples, across a diverse set of edit types. We provide the exact statistics of our dataset, including the different types of edits in Table~\ref{tab:edit-types}


\section{Experiments and Results}
\label{sec:experiments}

\begin{figure*}[!th]
    \centering

    \begin{minipage}[t]{0.91\textwidth}
        \centering
        \textbf{Style Transfer}
    \end{minipage}%
    \par
    
    \begin{minipage}[t]{0.13\textwidth}
        \centering
        \subcaption[]{$x$}{}
        \includegraphics[width=\textwidth]{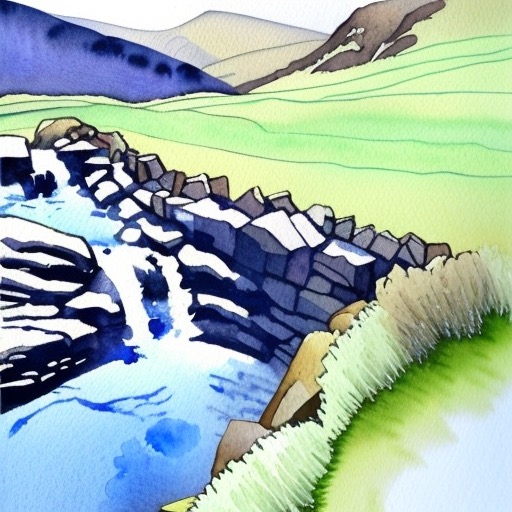}
    \end{minipage}%
    \begin{minipage}[t]{0.13\textwidth}
        \centering
        \subcaption[]{$x_{\text{edit}}$}{}
        \includegraphics[width=\textwidth]{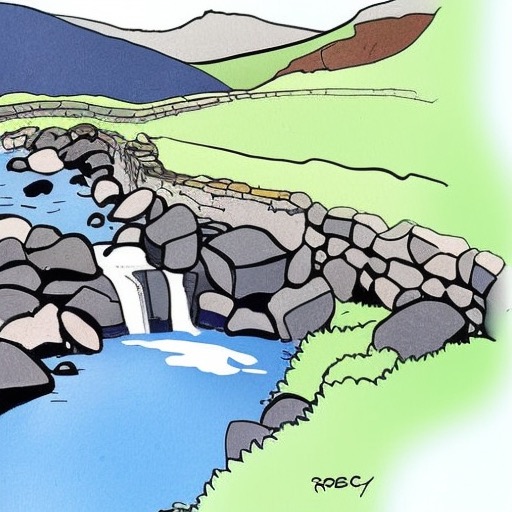}
    \end{minipage}%
    \begin{minipage}[t]{0.13\textwidth}
        \centering
        \subcaption[]{$y$}
        \includegraphics[width=\textwidth]{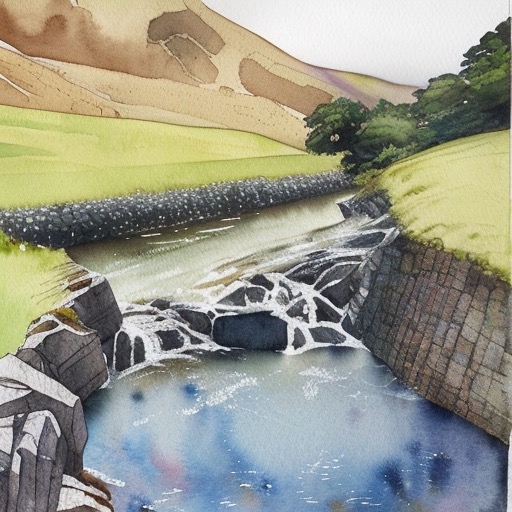}
    \end{minipage}%
    \begin{minipage}[t]{0.13\textwidth}
        \centering
        \subcaption[]{\textbf{\method}}{}
        \includegraphics[width=\textwidth]{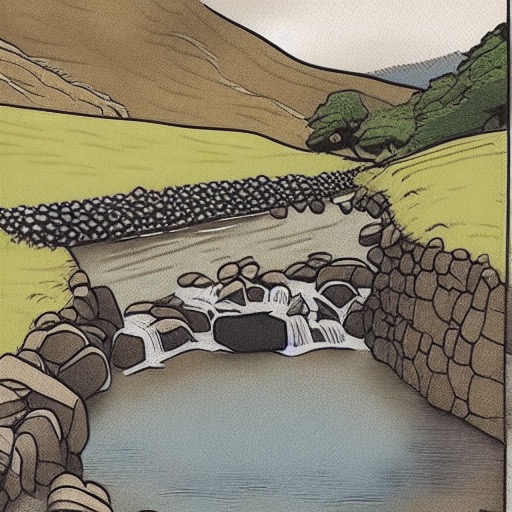}
    \end{minipage}%
    \begin{minipage}[t]{0.13\textwidth}
        \centering
        \subcaption[]{VISII}{}
        \includegraphics[width=\textwidth]{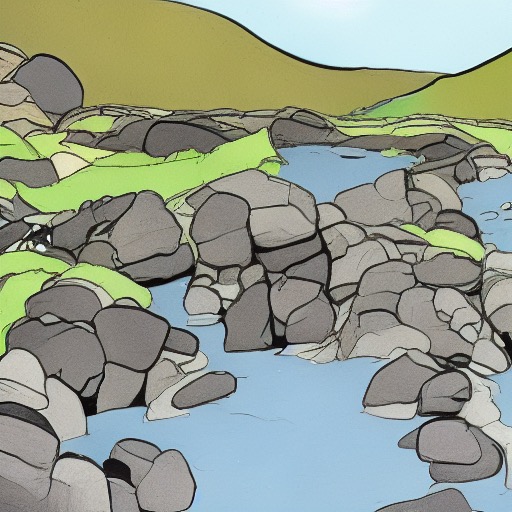}
    \end{minipage}%
    \begin{minipage}[t]{0.13\textwidth}
        \centering
        \subcaption[]{VISII w/ Text}{}
        \includegraphics[width=\textwidth]{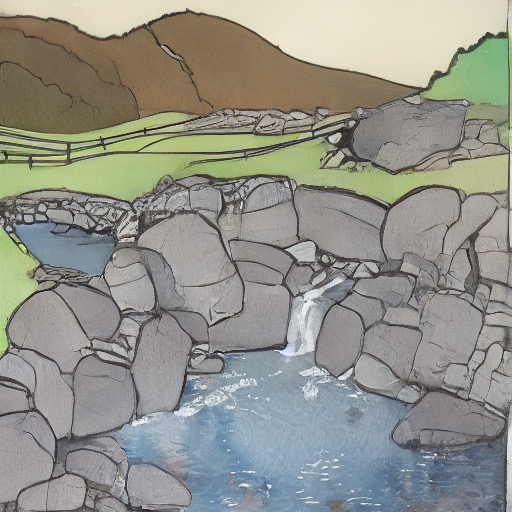}
    \end{minipage}%
    \begin{minipage}[t]{0.13\textwidth}
        \centering
        \subcaption[]{IP2P w/ Text}{}
        \includegraphics[width=\textwidth]{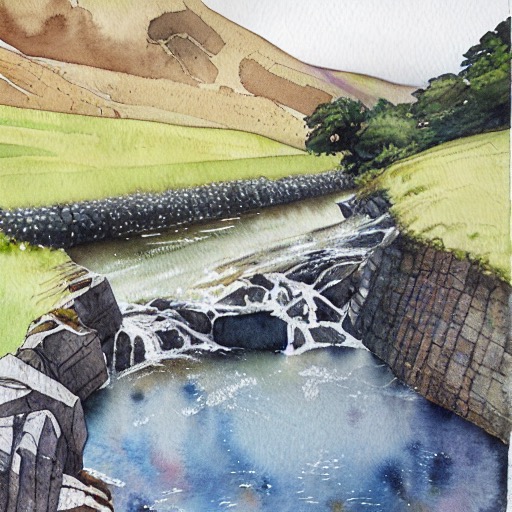}
    \end{minipage}%
    \par

       \begin{minipage}[t]{0.13\textwidth}
        \centering
        \includegraphics[width=\textwidth]{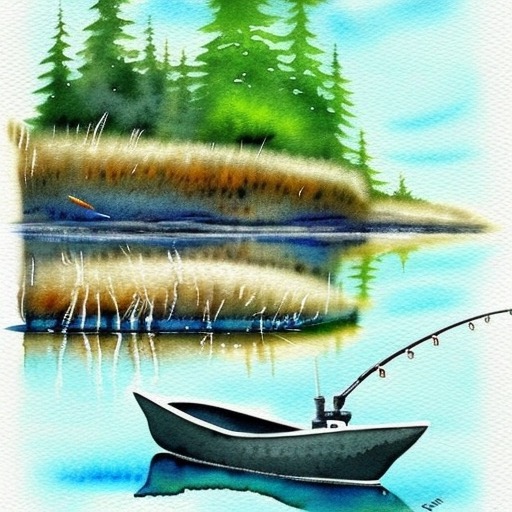}
    \end{minipage}%
    \begin{minipage}[t]{0.13\textwidth}
        \centering
        \includegraphics[width=\textwidth]{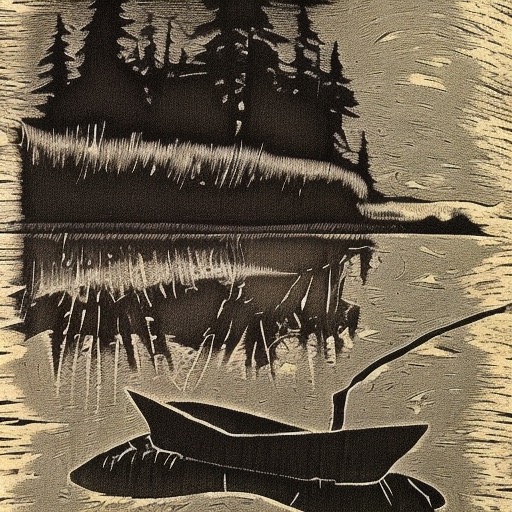}
    \end{minipage}%
    \begin{minipage}[t]{0.13\textwidth}
        \centering
        \includegraphics[width=\textwidth]{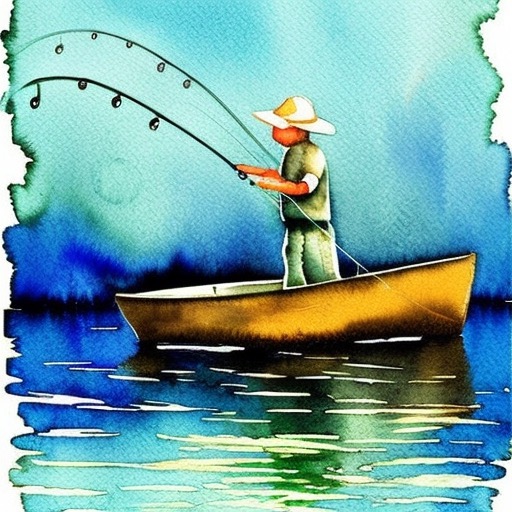}
    \end{minipage}%
    \begin{minipage}[t]{0.13\textwidth}
        \centering
        \includegraphics[width=\textwidth]{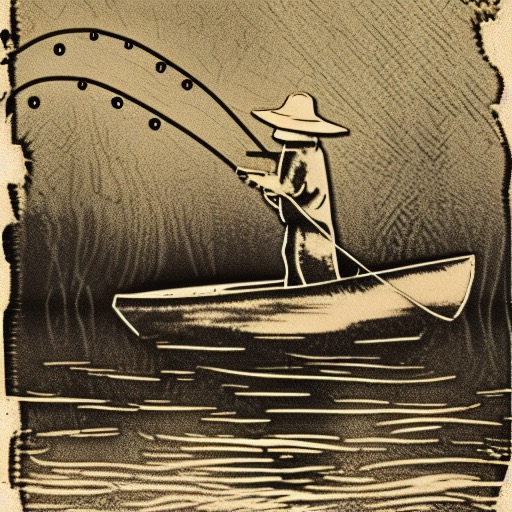}
    \end{minipage}%
    \begin{minipage}[t]{0.13\textwidth}
        \centering
        \includegraphics[width=\textwidth]{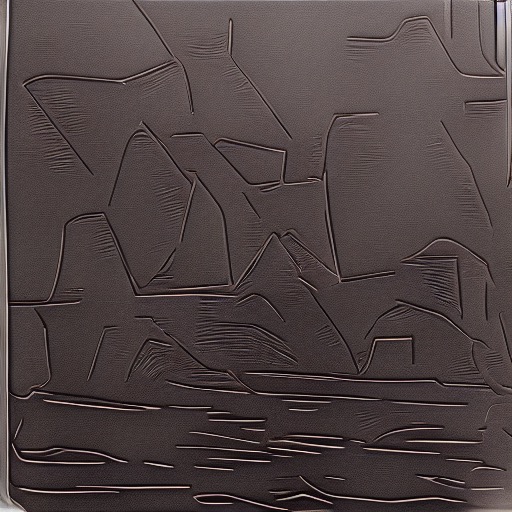}
    \end{minipage}%
    \begin{minipage}[t]{0.13\textwidth}
        \centering
        \includegraphics[width=\textwidth]{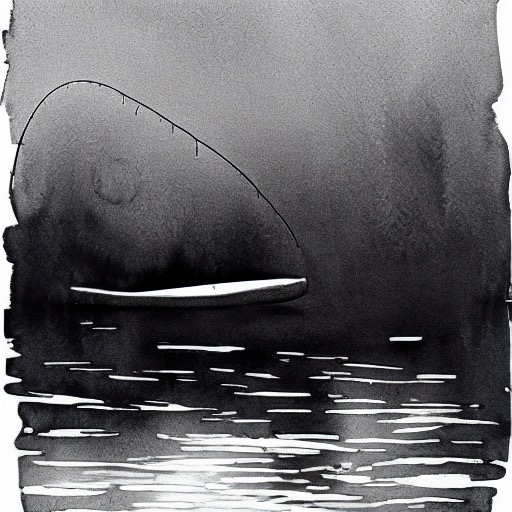}
    \end{minipage}%
    \begin{minipage}[t]{0.13\textwidth}
        \centering
        \includegraphics[width=\textwidth]{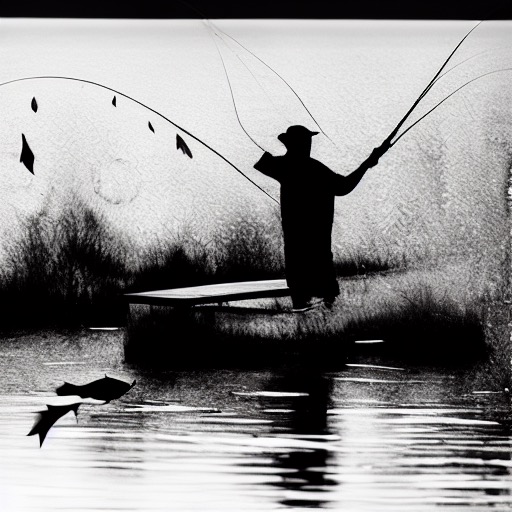}
    \end{minipage}%
    \par

    \begin{minipage}[t]{0.91\textwidth}
        \centering
        \textbf{Addition/Substitution}
    \end{minipage}%
    \par

    \begin{minipage}[t]{0.13\textwidth}
        \centering
        \includegraphics[width=\textwidth]{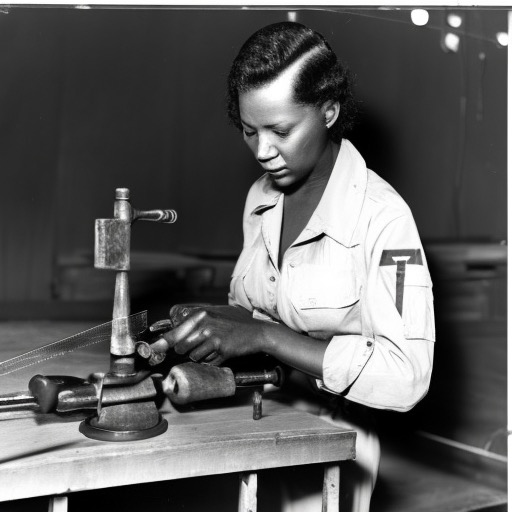}
    \end{minipage}%
    \begin{minipage}[t]{0.13\textwidth}
        \centering
        \includegraphics[width=\textwidth]{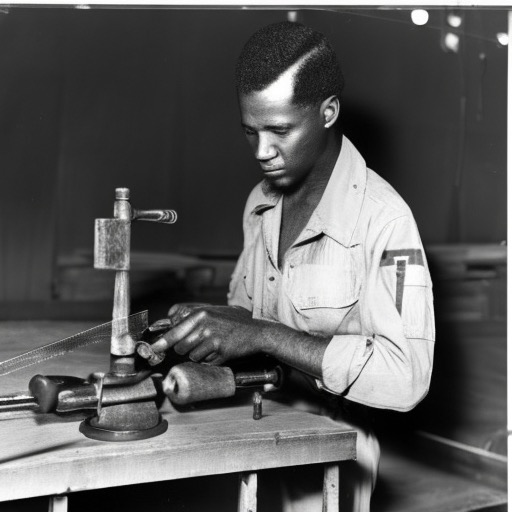}
    \end{minipage}%
    \begin{minipage}[t]{0.13\textwidth}
        \centering
        \includegraphics[width=\textwidth]{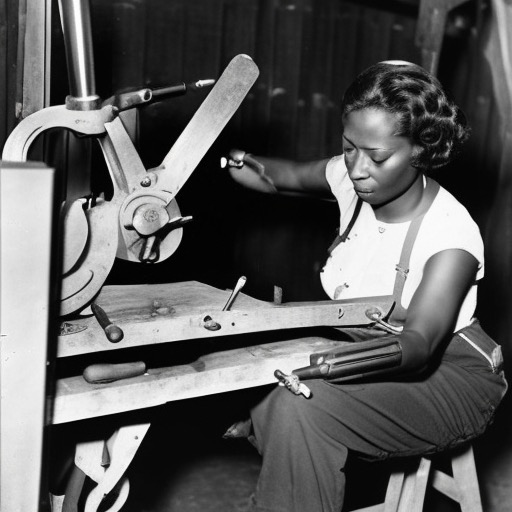}
    \end{minipage}%
    \begin{minipage}[t]{0.13\textwidth}
        \centering
        \includegraphics[width=\textwidth]{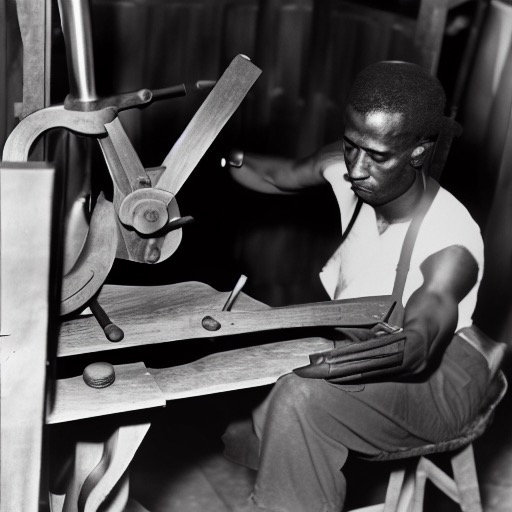}
    \end{minipage}%
    \begin{minipage}[t]{0.13\textwidth}
        \centering
        \includegraphics[width=\textwidth]{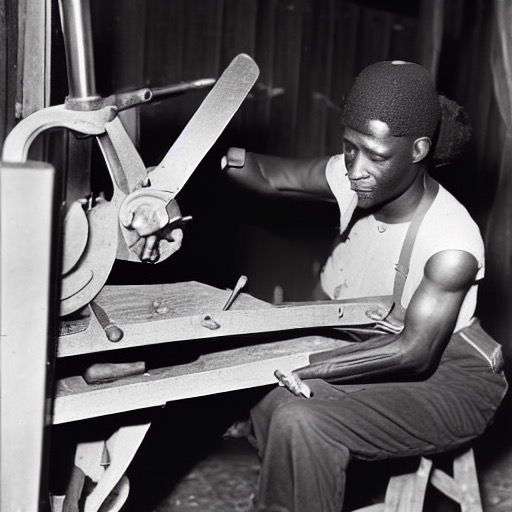}
    \end{minipage}%
    \begin{minipage}[t]{0.13\textwidth}
        \centering
        \includegraphics[width=\textwidth]{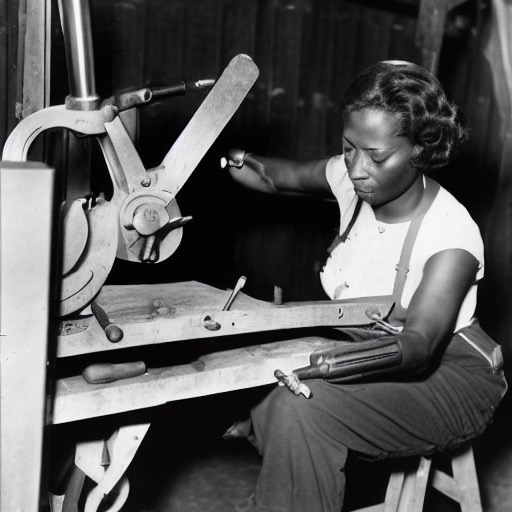}
    \end{minipage}%
    \begin{minipage}[t]{0.13\textwidth}
        \centering
        \includegraphics[width=\textwidth]{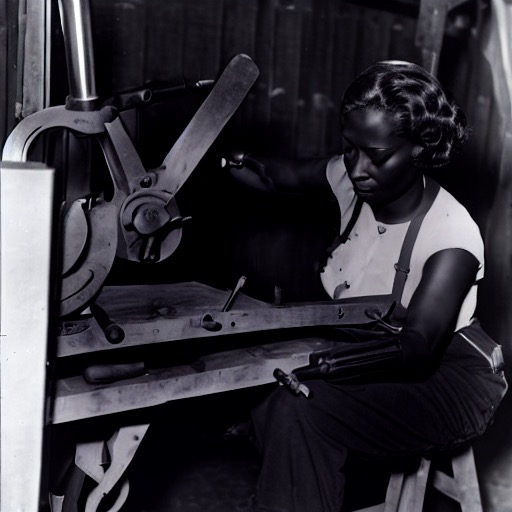}
    \end{minipage}%
    \par

    \begin{minipage}[t]{0.13\textwidth}
        \centering
        \includegraphics[width=\textwidth]{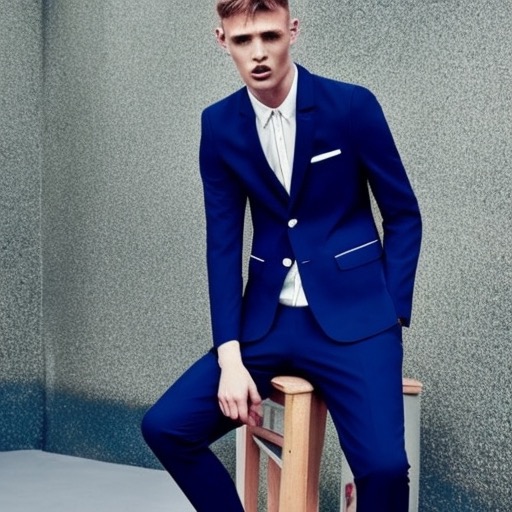}
    \end{minipage}%
    \begin{minipage}[t]{0.13\textwidth}
        \centering
        \includegraphics[width=\textwidth]{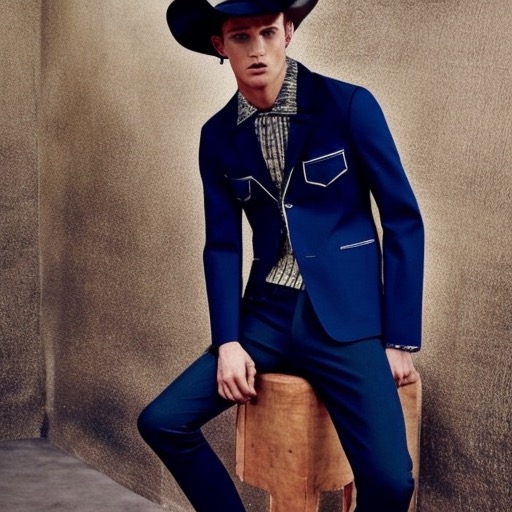}
    \end{minipage}%
    \begin{minipage}[t]{0.13\textwidth}
        \centering
        \includegraphics[width=\textwidth]{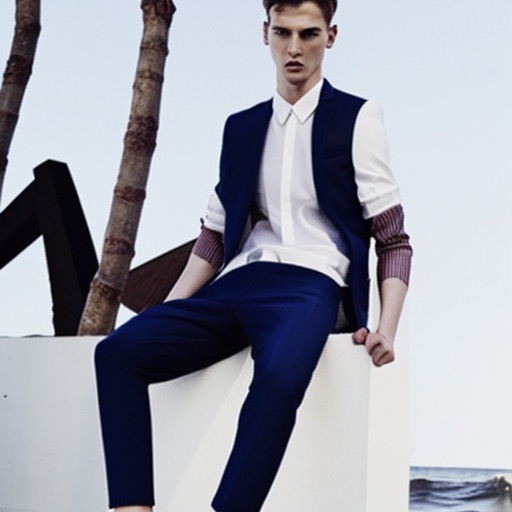}
    \end{minipage}%
    \begin{minipage}[t]{0.13\textwidth}
        \centering
        \includegraphics[width=\textwidth]{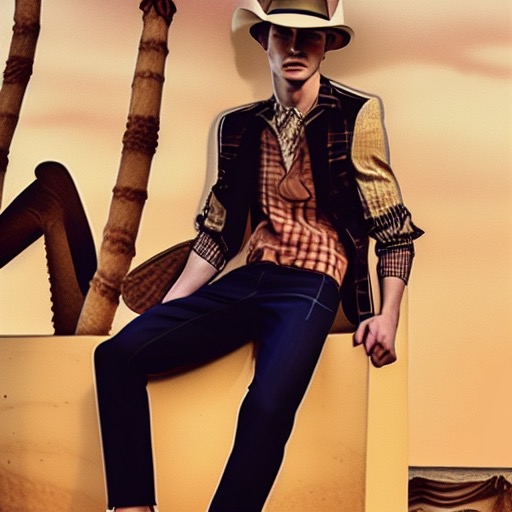}
    \end{minipage}%
    \begin{minipage}[t]{0.13\textwidth}
        \centering
        \includegraphics[width=\textwidth]{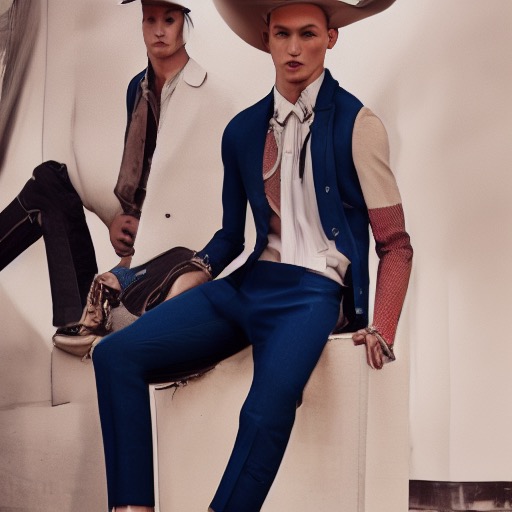}
    \end{minipage}%
    \begin{minipage}[t]{0.13\textwidth}
        \centering
        \includegraphics[width=\textwidth]{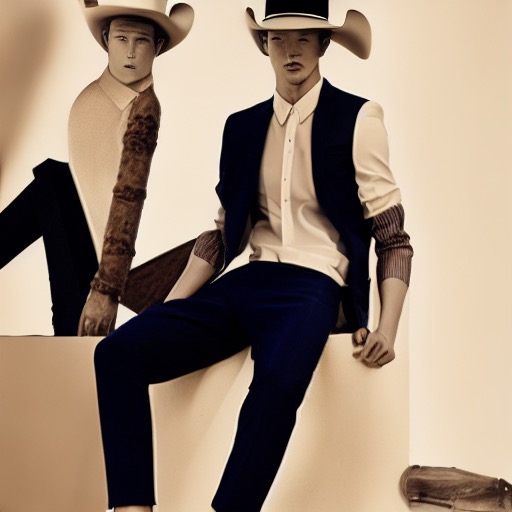}
    \end{minipage}%
    \begin{minipage}[t]{0.13\textwidth}
        \centering
        \includegraphics[width=\textwidth]{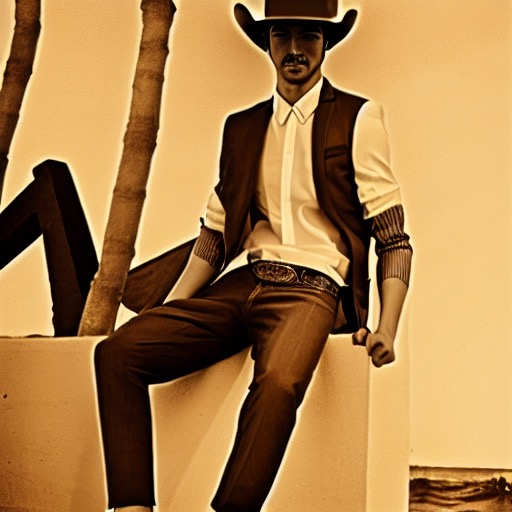}
    \end{minipage}%
    \par

    \begin{minipage}[t]{0.91\textwidth}
        \centering
        \textbf{Background Editing}
    \end{minipage}%
    \par

    \begin{minipage}[t]{0.13\textwidth}
        \centering
        \includegraphics[width=\textwidth]{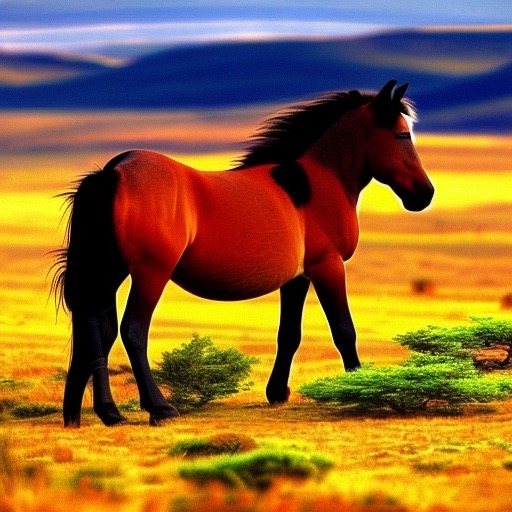}
    \end{minipage}%
    \begin{minipage}[t]{0.13\textwidth}
        \centering
        \includegraphics[width=\textwidth]{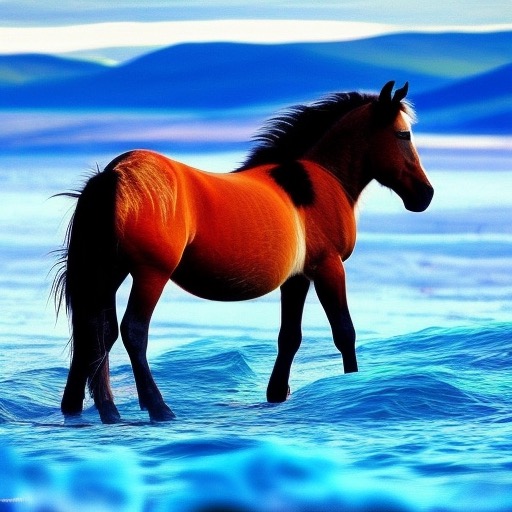}
    \end{minipage}%
    \begin{minipage}[t]{0.13\textwidth}
        \centering
        \includegraphics[width=\textwidth]{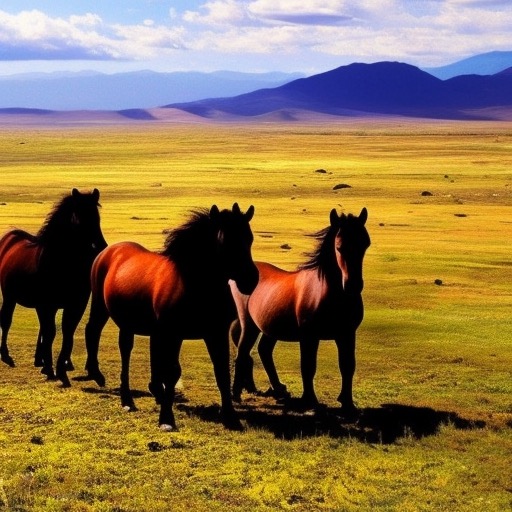}
    \end{minipage}%
    \begin{minipage}[t]{0.13\textwidth}
        \centering
        \includegraphics[width=\textwidth]{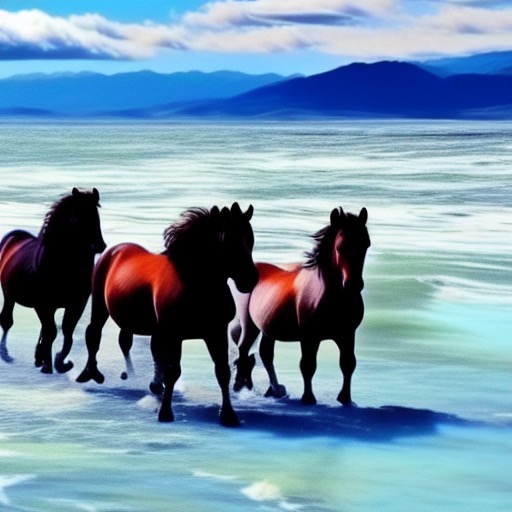}
    \end{minipage}%
    \begin{minipage}[t]{0.13\textwidth}
        \centering
        \includegraphics[width=\textwidth]{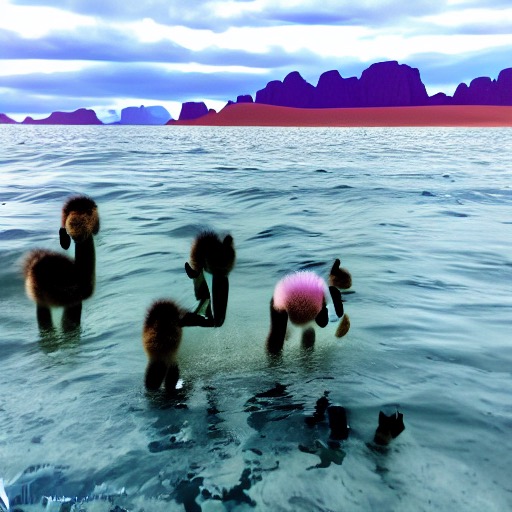}
    \end{minipage}%
    \begin{minipage}[t]{0.13\textwidth}
        \centering
        \includegraphics[width=\textwidth]{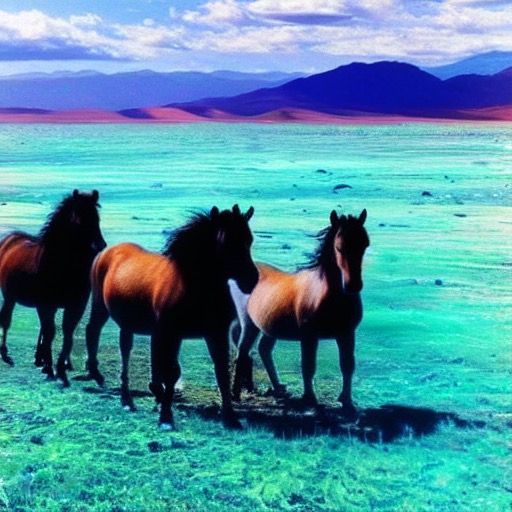}
    \end{minipage}%
    \begin{minipage}[t]{0.13\textwidth}
        \centering
        \includegraphics[width=\textwidth]{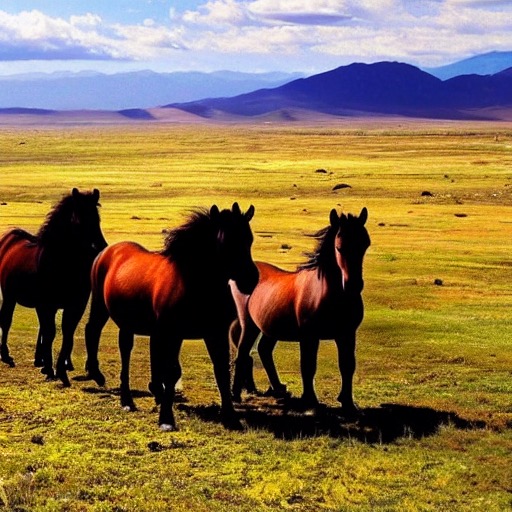}
    \end{minipage}%
    \par

    \begin{minipage}[t]{0.13\textwidth}
        \centering
        \includegraphics[width=\textwidth]{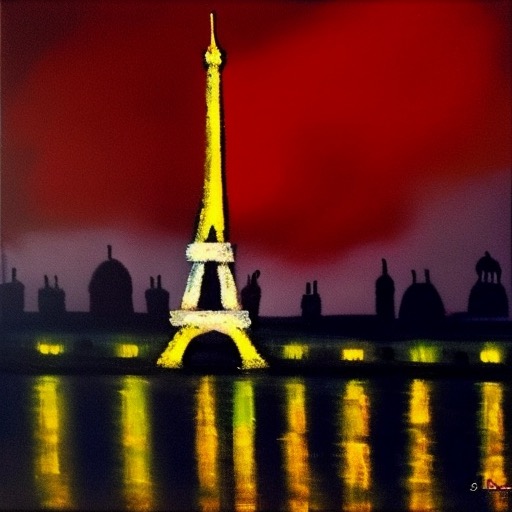}
    \end{minipage}%
    \begin{minipage}[t]{0.13\textwidth}
        \centering
        \includegraphics[width=\textwidth]{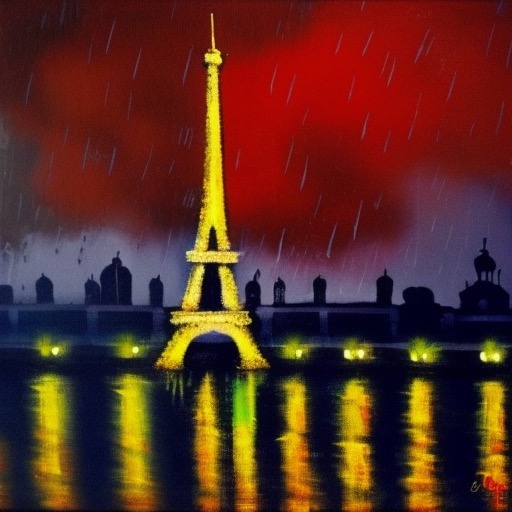}
    \end{minipage}%
    \begin{minipage}[t]{0.13\textwidth}
        \centering
        \includegraphics[width=\textwidth]{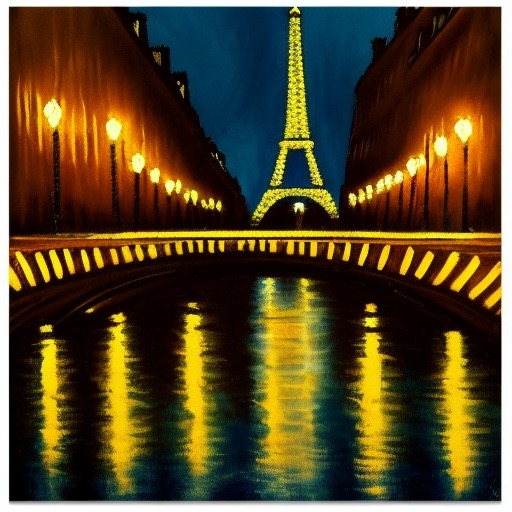}
    \end{minipage}%
    \begin{minipage}[t]{0.13\textwidth}
        \centering
        \includegraphics[width=\textwidth]{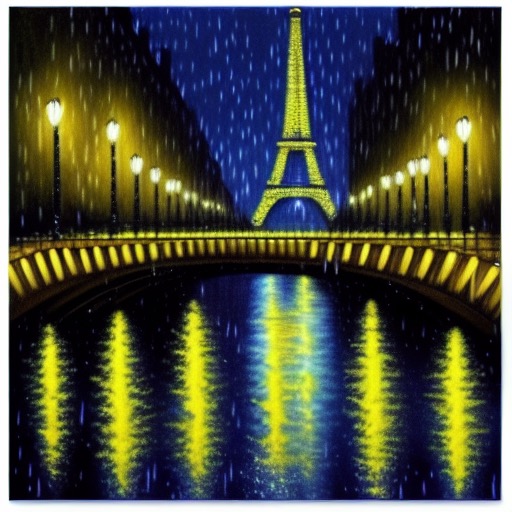}
    \end{minipage}%
    \begin{minipage}[t]{0.13\textwidth}
        \centering
        \includegraphics[width=\textwidth]{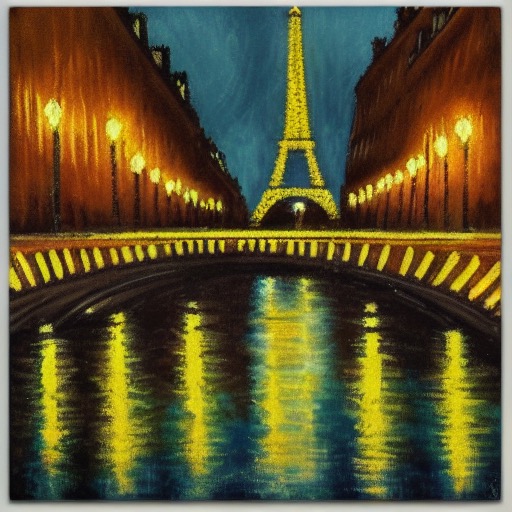}
    \end{minipage}%
    \begin{minipage}[t]{0.13\textwidth}
        \centering
        \includegraphics[width=\textwidth]{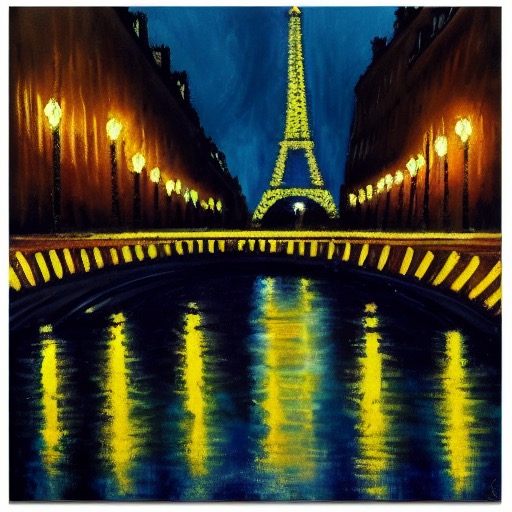}
    \end{minipage}%
    \begin{minipage}[t]{0.13\textwidth}
        \centering
        \includegraphics[width=\textwidth]{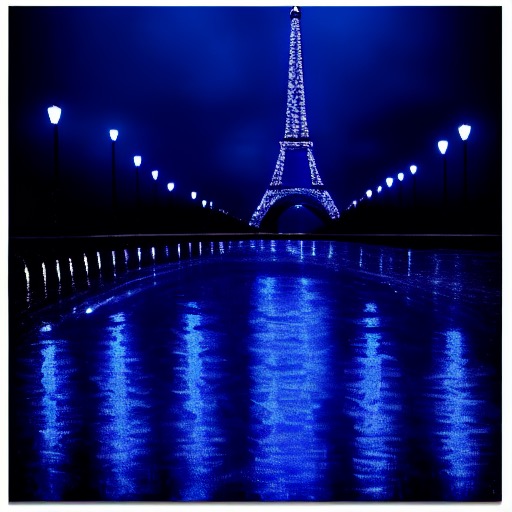}
    \end{minipage}%
    \par

    \begin{minipage}[t]{0.91\textwidth}
        \centering
        \textbf{Localized Edits}
    \end{minipage}%
    \par

    \begin{minipage}[t]{0.13\textwidth}
        \centering
        \includegraphics[width=\textwidth]{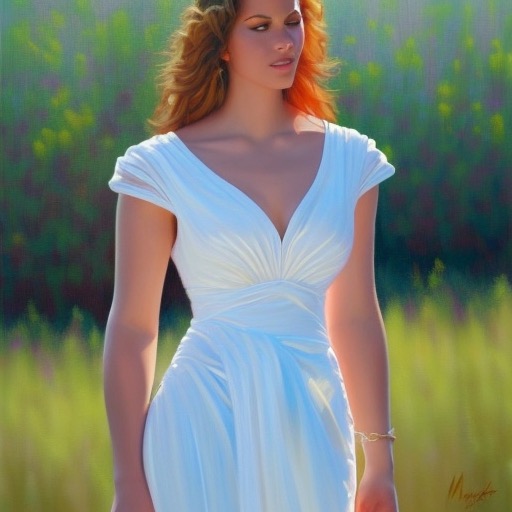}
    \end{minipage}%
    \begin{minipage}[t]{0.13\textwidth}
        \centering
        \includegraphics[width=\textwidth]{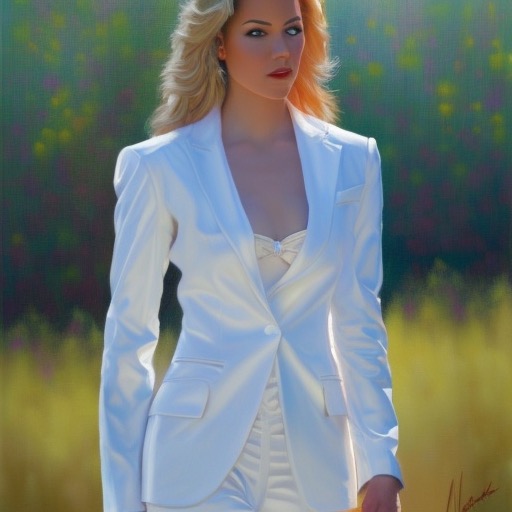}
    \end{minipage}%
    \begin{minipage}[t]{0.13\textwidth}
        \centering
        \includegraphics[width=\textwidth]{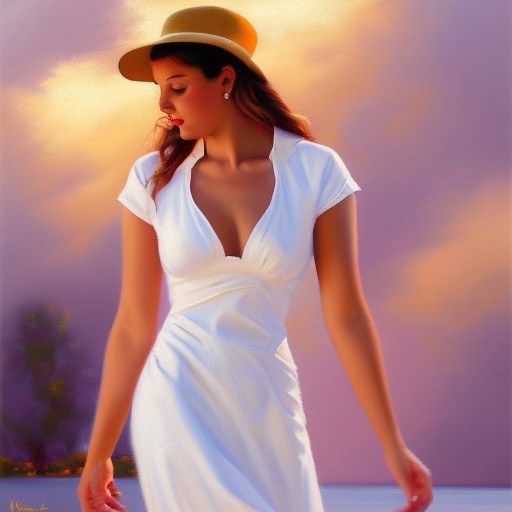}
    \end{minipage}%
    \begin{minipage}[t]{0.13\textwidth}
        \centering
        \includegraphics[width=\textwidth]{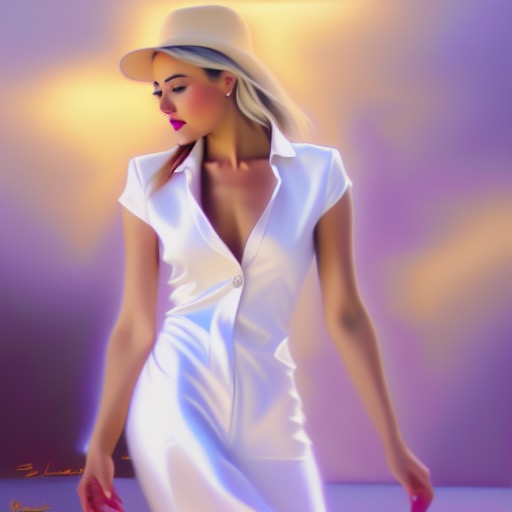}
    \end{minipage}%
    \begin{minipage}[t]{0.13\textwidth}
        \centering
        \includegraphics[width=\textwidth]{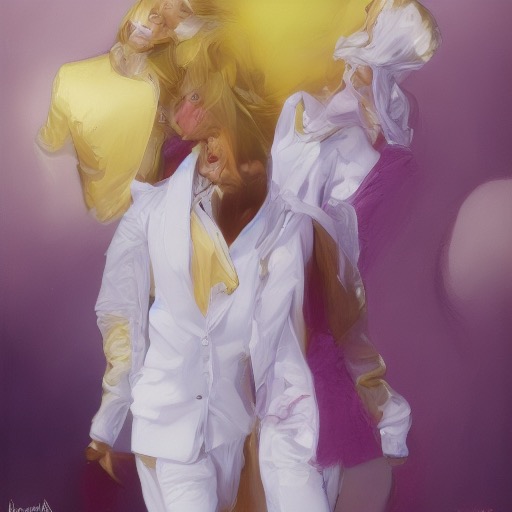}
    \end{minipage}%
    \begin{minipage}[t]{0.13\textwidth}
        \centering
        \includegraphics[width=\textwidth]{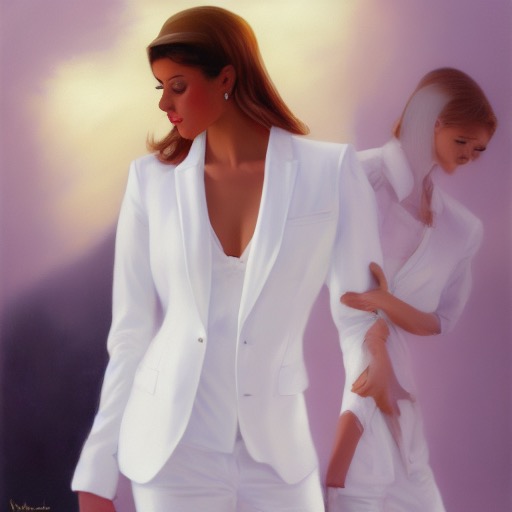}
    \end{minipage}%
    \begin{minipage}[t]{0.13\textwidth}
        \centering
        \includegraphics[width=\textwidth]{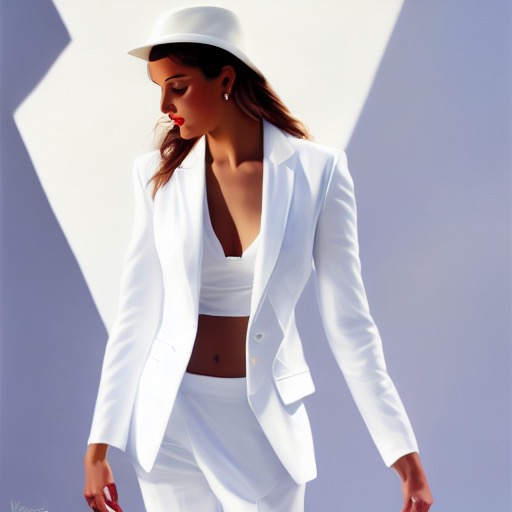}
    \end{minipage}%
    \par

    \begin{minipage}[t]{0.13\textwidth}
        \centering
        \includegraphics[width=\textwidth]{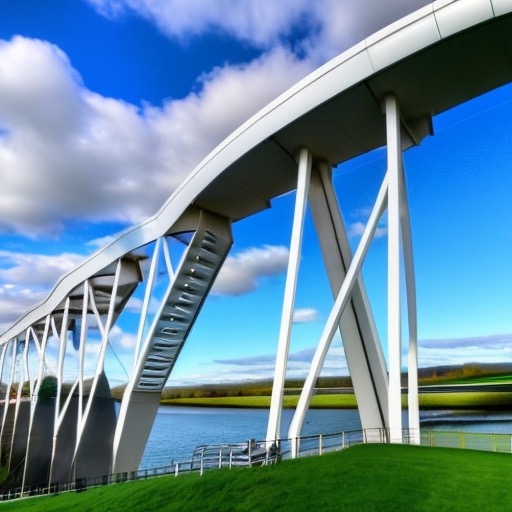}
    \end{minipage}%
    \begin{minipage}[t]{0.13\textwidth}
        \centering
        \includegraphics[width=\textwidth]{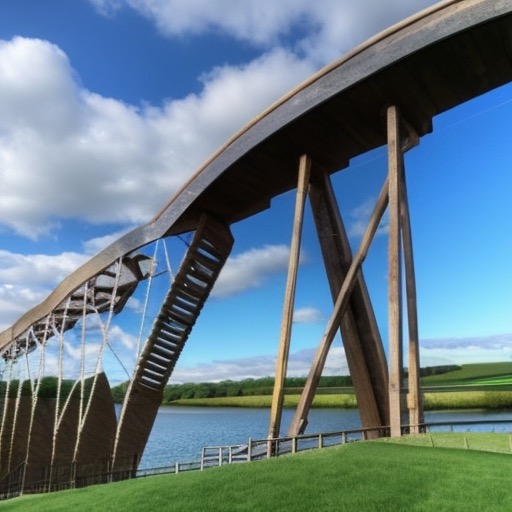}
    \end{minipage}%
    \begin{minipage}[t]{0.13\textwidth}
        \centering
        \includegraphics[width=\textwidth]{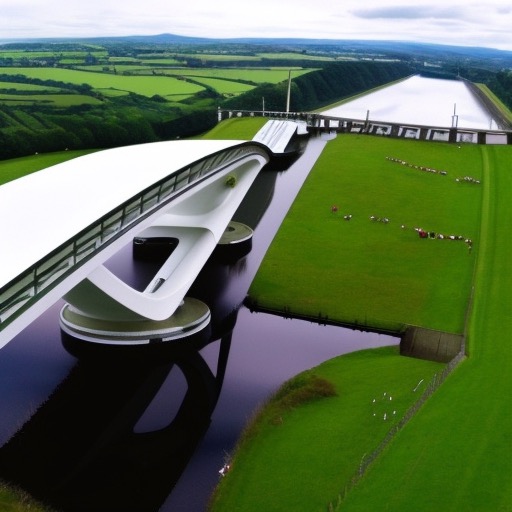}
    \end{minipage}%
    \begin{minipage}[t]{0.13\textwidth}
        \centering
        \includegraphics[width=\textwidth]{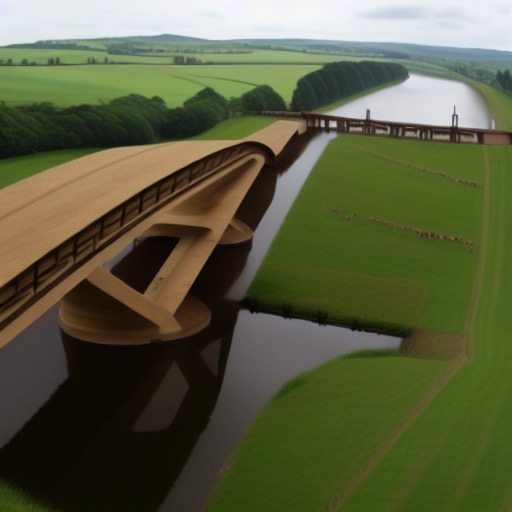}
    \end{minipage}%
    \begin{minipage}[t]{0.13\textwidth}
        \centering
        \includegraphics[width=\textwidth]{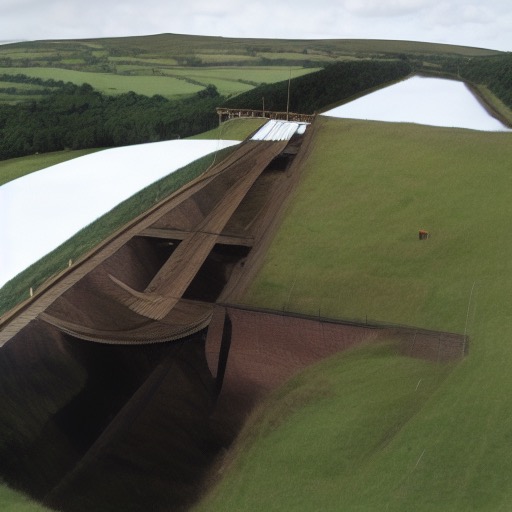}
    \end{minipage}%
    \begin{minipage}[t]{0.13\textwidth}
        \centering
        \includegraphics[width=\textwidth]{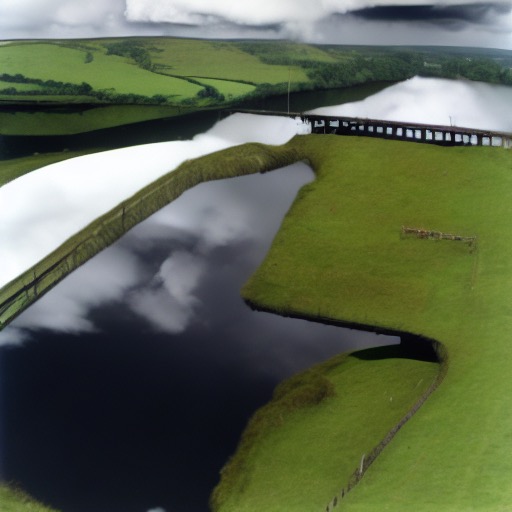}
    \end{minipage}%
    \begin{minipage}[t]{0.13\textwidth}
        \centering
        \includegraphics[width=\textwidth]{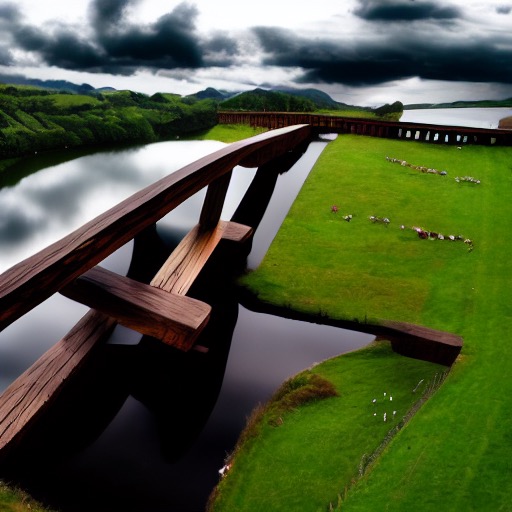}
    \end{minipage}%
    \par

    \caption{Qualitative comparison of our framework \method with strong baselines (VISII, InstructPix2Pix) for exemplar-based image editing. \method consistently produces images with higher edit accuracy and better consistency in non-edited regions compared to the baselines. Zoom in for better view. Additional results presented in Appendix~\ref{sec:addn-qual}\label{fig:qual}}
\end{figure*}

In this section, we first provide a detailed description of the implementation which includes the hyperparameter choices of both \method and baselines. Next, equipped with our curated dataset, we evaluate the performance of \method. The key feature of our dataset is the presence of a ground truth edited image, denoted by $y_{\text{edit}}$ which enables us to use several standard image quality evaluation metrics. As a result, we compute five quantitative metrics across the full dataset. measuring structural and perceptual similarity between $(\hat{y}_{\text{edit}}, y_{\text{edit}})$ (LPIPS~\cite{zhang2018unreasonable}, SSIM~\cite{wang2004image}), faithfulness of the edit with the exemplar pair (CLIP score~\cite{hessel2021clipscore}, Dir. Similarity~\cite{gal2021stylegan}, and S-Visual~\cite{nguyen2024visual}). For each method in the comparison, the quantitative scores are reported by selecting the hyperparameter which yields the best average performance across all metrics. Though it is noteworthy that different edit types may require different hyperparamters to yield the best quality results, so in our qualitative analysis we choose the best per-sample hyperparameter for each method.

We present our quantitative results in Table~\ref{tab:quant}, show several qualitative examples in Fig~\ref{fig:qual}, and show the running times of all the methods in Table~\ref{tab:timing}. Further, we also illustrate additional qualitative examples in Appendix~\ref{sec:addn-qual}. A detailed discussion of our results can be found in Sec.~\ref{sec:discussion}. Appendix~\ref{sec:metrics} includes further details on the usage and implementation of the various metrics.

\subsection{Implementation details}
\label{sec:impl}
 We do an extensive comparison of our framework \method with existing approaches. The main work we compare against is VISII~\cite{nguyen2024visual}, another inference-time exemplar based editing method. To ensure a fair comparison, we further augment VISII with LLaVA generated instructions as an additional baseline. Finally, we also compare against a text-based editing approach - InstructPix2Pix, once again using the LLaVA generated edit instruction as input. 
 All baselines are evaluated on a single A100 GPU with 80GB memory, and all images are resized to $512\times512$ pixels. We now further describe the exact setup and hyperparameters for \method and the various baselines when generating results for qualitative and quantitative analysis. We employ LLaVA-1.6~\cite{liu2024llava} for obtaining automated captions and edit instructions in all methods. 
 
\xhdr{\method}
We use SD1.5 with IP-Adapter~\cite{ye2023ip-adapter} as the base model. To compute the image prompt embedding CLIP ViT-L/14 is used, which is then pooled to 4 tokens. 
3 images $x, x_{edit}$ and $y$ are used to generate the image prompt embedding $\Delta_\text{img}$, and the previously describe pipeline to generate the text caption using LLaVA. The input to a standard text-to-image diffusion model (here, Stable Diffusion v1.5) is typically a sequence of $77$ tokens (sequence length of CLIP text encoder), however after adding the image prompt embedding this becomes $81$ tokens. The last 4 tokens are are separately processed in IP-Adapter's cross attention modules.

We perform DDIM inversion of the test image $y$ for $1000$ steps to generate $y_{\text{noise}}$ for feature and self attention guidance. The features and self-attention ($f, Q, K$) from the vanilla denoising of $y_{\text{noise}}$ are injected at each of the $4-11^{th}$ layers of the upsampling blocks respectively. An important parameter is the classifier free guidance (CFG~\cite{ho2022classifier}) weight, which we fix to $10$ in all our experiments. We do not experiment with different values of CFG, and instead we only vary the edit weight ($\lambda$)  in our experiments as the sole hyperparameter in \method. 




\noindent{\xhdr{a. VISII}} \cite{nguyen2024visual} optimizes an edit instruction $c_T$ in the latent space of the CLIP text encoder of InstructPix2Pix to learn the edit from the exemplar pair $(x, x_\text{edit}$. This learnt instruction $c_T$ is used as input along with the test image $y$ to obtain the desired edit. We adopt their original setp, and optimize for $T=1000$ steps using AdamW~\cite{loshchilov2017decoupled} with learning rate 1e-4, and the respective weights for the loss term in VISII as $\lambda_{\text{mse}} = 4$ and $\lambda_{\text{clip}} = 0.1$. Following the original setup, for each inference, we perform 8 independent optimizations with different random seeds, and choose the $c_T$ that minimizes the overall loss. We experiment with multiple values for text guidance from ${8, 10, 12}$ and use the default image guidance of $1.5$ as the hyperparameters. 

\noindent{\xhdr{b. VISII with text}} We introduce an important augmentation to VISII, concatenating an additional textual instruction as suggested in the original work. This helps guide the model using both natural language and image differences. We generate the edit text similar to the approach in Sec.~\ref{sec:method}. First, we pass a grid of exemplar pairs ($x, x_{\text{edit}}$) and a detailed prompt $p1$ to LLaVA to generate a detailed edit text $g_{\text{edit}}$. Next, instead of generating $g_{\text{caption}}$ from $(g_{\text{edit}}, y, p2)$ we instead instruct LLaVA to generate a short summary of the \emph{edit instruction} (say $g_{\text{edit-inst}}$) using $g_{\text{edit}}, y, p3$ where $p3$ is a simple modification of $p2$ instructing LLaVA to generate an edit instruction for InstructPix2Pix. Refer to Appendix~\ref{sec:addn-llava} for all prompts. The hyperparameters and optimization follow ~\textit{a. VISII}

\noindent{\xhdr{c. InstructPix2Pix}} We directly use a supervised pre-trained text-based image editing model, Instruct-Pix2Pix~\cite{brooks2023instructpix2pix}. Thought InstructPix2Pix was intended to be used with custom text instructions, the goal in this setting is to transfer edits without explicit supervision. Hence, we utilize LLaVA edit instructions $g_{\text{edit-inst}}$ as described in \textit{b. VISII with text}. We experiment over the same set of hyperparameter values as the previous two baselines.

\begin{table*}[]
\centering
\small{
\begin{tabular}{@{}lcccc@{}}
\toprule
\textbf{Metric} & \begin{tabular}[c]{@{}c@{}}IP2P w/ LLaVA edit text~\cite{brooks2023instructpix2pix}\end{tabular} & \begin{tabular}[c]{@{}c@{}}VISII w/ LLaVA edit text\end{tabular} & \begin{tabular}[c]{@{}c@{}}VISII~\cite{nguyen2024visual}\end{tabular} & \textbf{\method} (Ours) \\ \midrule

LPIPS~\cite{zhang2018unreasonable} ($\downarrow$) & 0.33 & \underline{0.27} & 0.29 & \textbf{0.26} \\

SSIM~\cite{wang2004image} ($\uparrow$) & 0.46 & \textbf{0.51} & \underline{0.48} & \textbf{0.51} \\

CLIP Score~\cite{hessel2021clipscore} ($\uparrow$) & 29.77  & \underline{31.62}  & \textbf{31.75} & 31.38\\

Dir. Similarity & 0.03 & \underline{0.04} & \underline{0.04} & \textbf{0.05} \\

S-Visual~\cite{nguyen2024visual} ($\uparrow$) & 0.22 & \underline{0.32} & \textbf{0.39} & \textbf{0.39} \\ \bottomrule
\end{tabular}}
\caption{Quantitative comparison of our framework \method against strong baselines -- VISII (and its modifications), and Instruct-pix2pix (IP2P). Reported are the mean of of five different metrics on our dataset. (best scores in \textbf{bold}; second best \underline{underlined})} \label{tab:quant}
\end{table*}





\subsection{Discussion of Results}
\label{sec:discussion}
We report the results with the best hyperparameter values for each baseline, as shown in Table~\ref{tab:quant}. \method outperforms strong baselines in SSIM, LIPIS, Dir. Similarity, and S-Visual, and is competitive in CLIP Score. VISII optimizes the latent instruction $c_T$ with eight random seeds per sample, and selects the $c_T$ that minimizes the loss term. In contrast, \method does not use repeated generations but still performs well on average, highlighting its computational efficiency, which is crucial for practical editing applications. These findings are further supported by qualitative examples in Fig.~\ref{fig:qual}, where \method demonstrates superior performance across various edit types. Next, we discuss our key observations and results from Fig~\ref{fig:qual}.

\xhdr{Style Transfer} In Rows 1-2, \method successfully captures the stylistic edit from the exemplar pair and applies it to the target image, while completely maintaining the original structure. In both cases, VISII captures the desired style from the exemplar pair, but is unable to maintain the structure of the $y$ image while applying the edit. InstructPix2Pix on the other hand does not sufficiently capture the stylistic information, showing the inability of text alone to sufficiently capture the edit. 

\xhdr{Subject/Background Editing} require addition or replacement in the subject or background of the image, while leaving other elements unchanged. \method is able to capture and apply these edits while causing less visual disruption compared to the baselines. In Row 5, only \method is successful at changing the background to an ocean while keeping the horses intact. \method is able to add the subtle raindrops from the exemplar pair in Row 6. Further, only \method is able to capture \textit{all} aspects of the desired edit in Row 4, while other baselines only edit the hat, and not the man's shirt. 

\xhdr{Localized Editing} Rows 7-8 show the ability of \method to capture and apply fine-grained edits, such as changing the woman's dress to a suit, or altering her appearance without changing the dressing, while keeping all other elements largely intact. Here, VISII, and VISII w/ Text introduce noise and artifacts into the test image, while InstructPix2Pix + LLaVA text is unable to maintain the background. In Row 8, the subtle change to a wooden structure is perfectly captured and applied to \textit{only the required regions} by \method, while other methods completely fail. Additional qualitative results are presented in Appendix~\ref{sec:addn-qual}. 

\begin{table}[]
\centering
\small{
    \begin{tabular}{lc}
        \toprule
        Method & Average Inference Time (s) \\
        \midrule
        \method & 120s \\
        VISII & 540s \\
        VISII w/ Text & 550s \\
        IP2P w/ Text & 40s\\
        \bottomrule
    \end{tabular}}
    \caption{Average running time for different methods. Includes all steps in the respective pipelines. \method is more than $\sim$4 times faster than the most performant baseline - VISII w/ Text. As shown in Sec.~\ref{sec:experiments}, IP2P w/text is not performant in this setting.}
    \label{tab:timing}
\end{table}
\section{Ablation Analysis}
\label{sec:additional-ablations}

\begin{figure*}[h!]
    \centering
    \begin{minipage}[t]{0.13\textwidth}
        \centering
        \subcaption[]{$x$}{}
        \includegraphics[width=\textwidth]{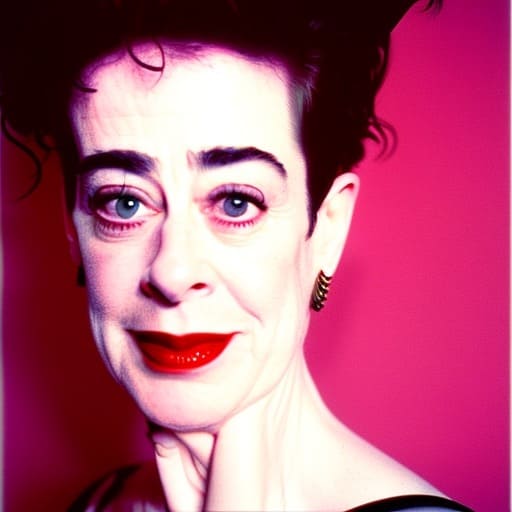}
    \end{minipage}%
    \begin{minipage}[t]{0.13\textwidth}
        \centering
        \subcaption[]{$x_{\text{edit}}$}
        \includegraphics[width=\textwidth]{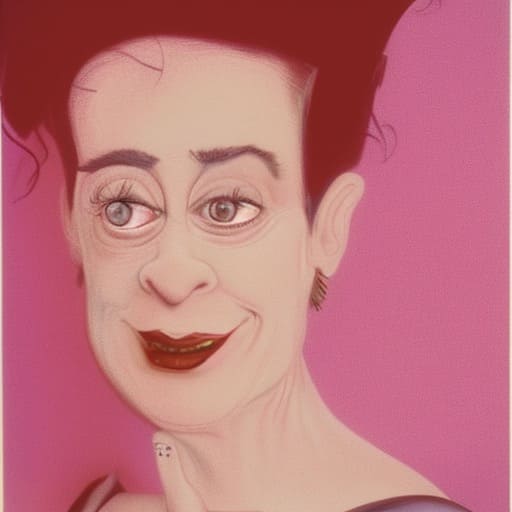}
    \end{minipage}%
    \begin{minipage}[t]{0.13\textwidth}
        \centering
        \subcaption[]{$y$}{}
        \includegraphics[width=\textwidth]{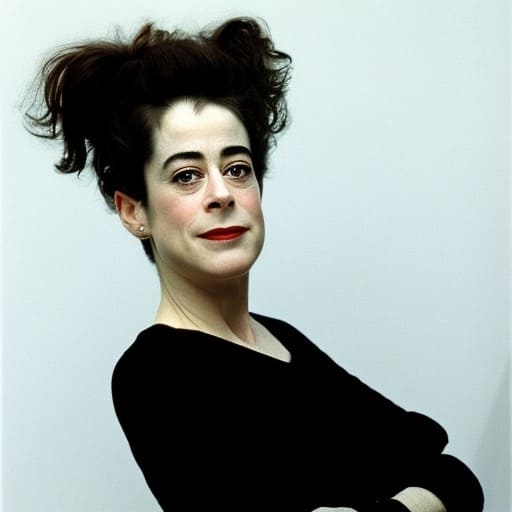}
    \end{minipage}%
    \begin{minipage}[t]{0.13\textwidth}
        \centering
        \subcaption[]{\method}{}
        \includegraphics[width=\textwidth]{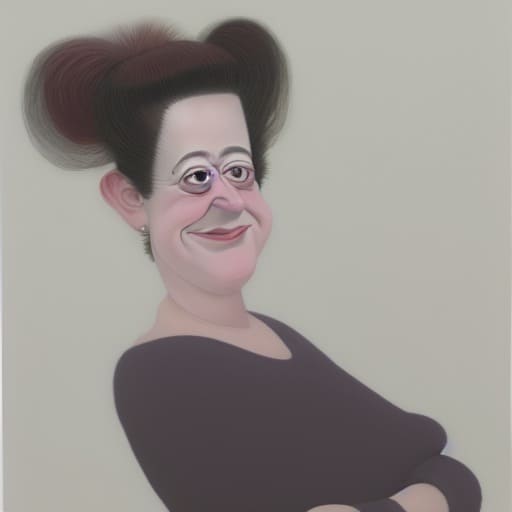}
    \end{minipage}%
    \begin{minipage}[t]{0.13\textwidth}
        \centering
        \subcaption[]{~$-f, Q, K$}{}
        \includegraphics[width=\textwidth]{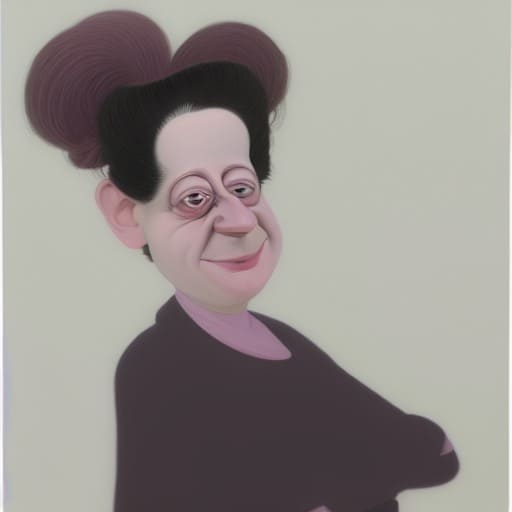}
    \end{minipage}%
    \begin{minipage}[t]{0.13\textwidth}
        \centering
        \subcaption[]{~$-g_{\text{caption}}$}{}
        \includegraphics[width=\textwidth]{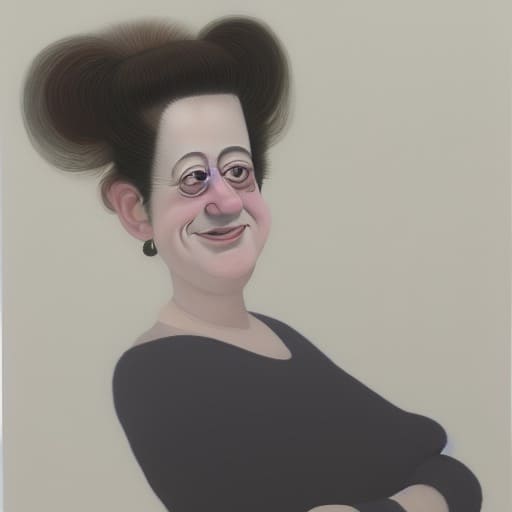}
    \end{minipage}%
    \begin{minipage}[t]{0.13\textwidth}
        \centering
        \subcaption[]{~$-\Delta_{\text{img}}$}{}
        \includegraphics[width=\textwidth]{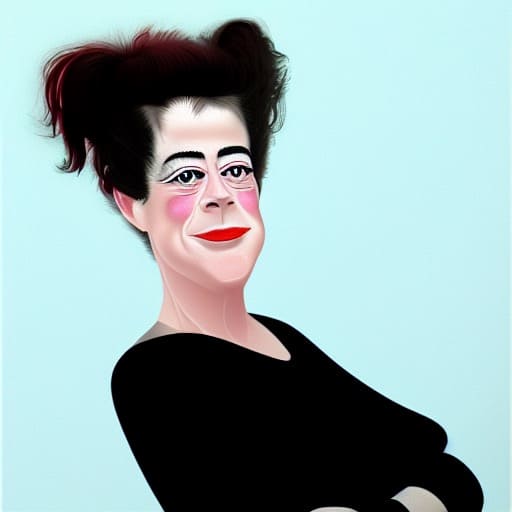}
    \end{minipage}%
    \par

    \begin{minipage}[t]{0.13\textwidth}
        \centering
        \includegraphics[width=\textwidth]{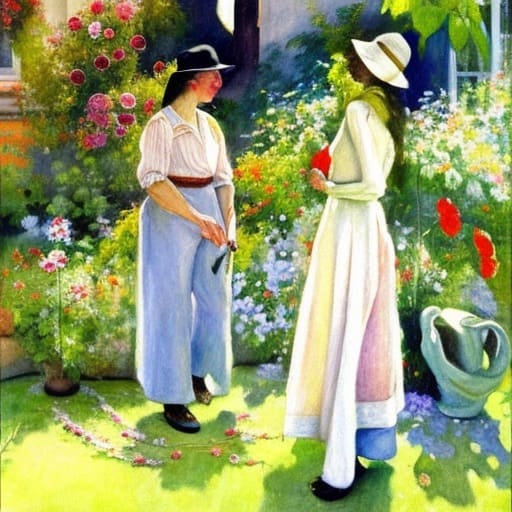}
    \end{minipage}%
    \begin{minipage}[t]{0.13\textwidth}
        \centering
        \includegraphics[width=\textwidth]{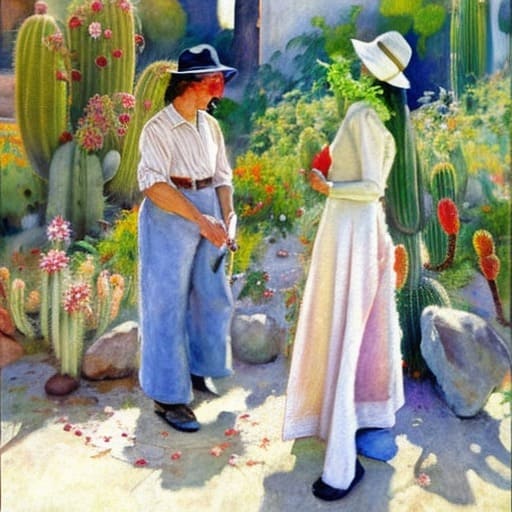}
    \end{minipage}%
    \begin{minipage}[t]{0.13\textwidth}
        \centering
        \includegraphics[width=\textwidth]{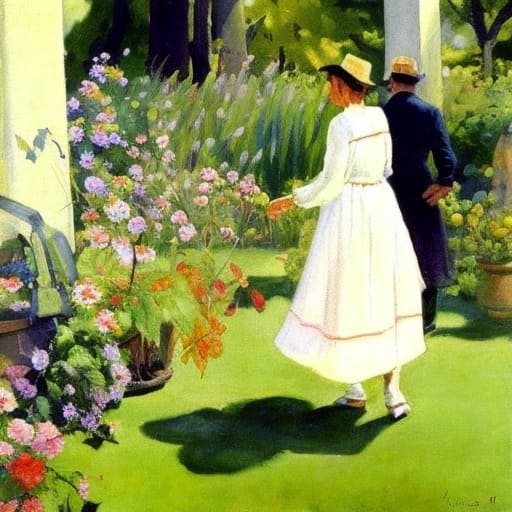}
    \end{minipage}%
    \begin{minipage}[t]{0.13\textwidth}
        \centering
        \includegraphics[width=\textwidth]{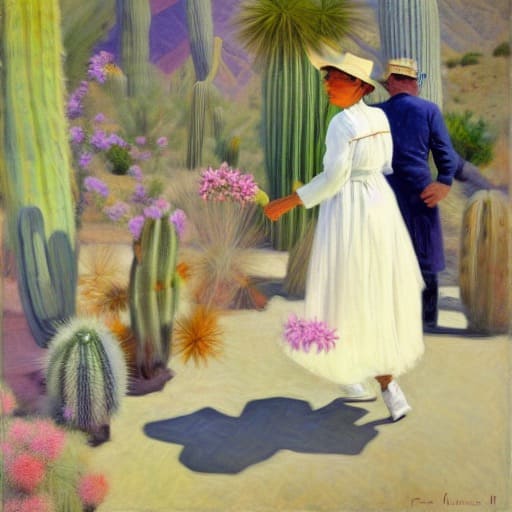}
    \end{minipage}%
    \begin{minipage}[t]{0.13\textwidth}
        \centering
        \includegraphics[width=\textwidth]{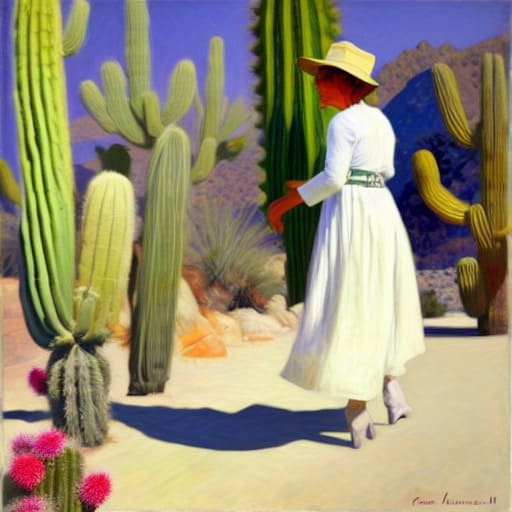}
    \end{minipage}%
    \begin{minipage}[t]{0.13\textwidth}
        \centering
        \includegraphics[width=\textwidth]{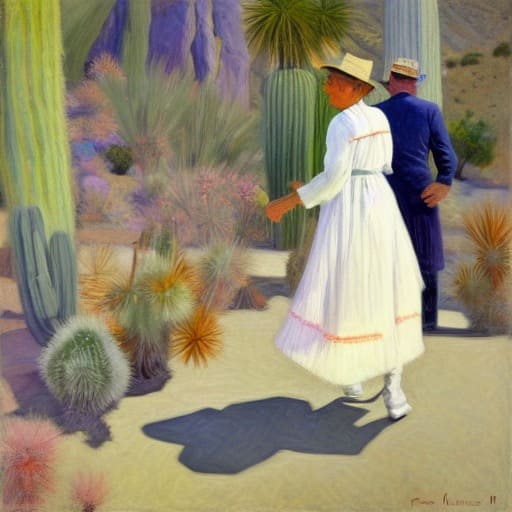}
    \end{minipage}%
    \begin{minipage}[t]{0.13\textwidth}
        \centering
        \includegraphics[width=\textwidth]{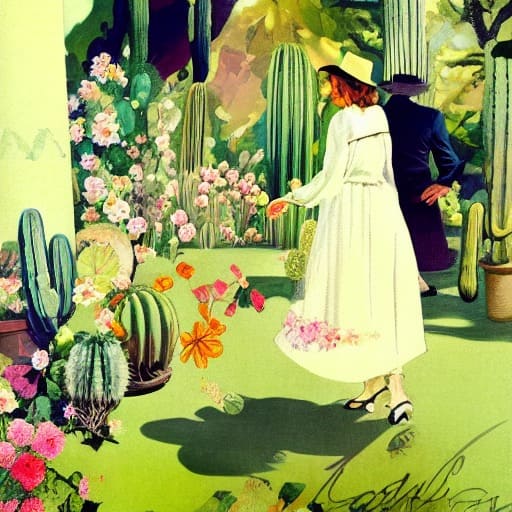}
    \end{minipage}%
    \par

    \begin{minipage}[t]{0.13\textwidth}
        \centering
        \includegraphics[width=\textwidth]{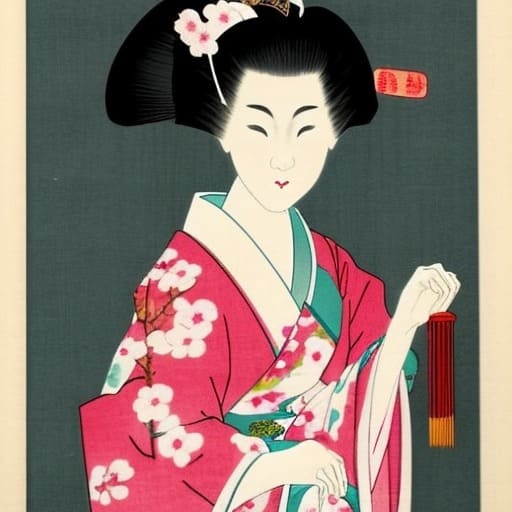}
    \end{minipage}%
    \begin{minipage}[t]{0.13\textwidth}
        \centering
        \includegraphics[width=\textwidth]{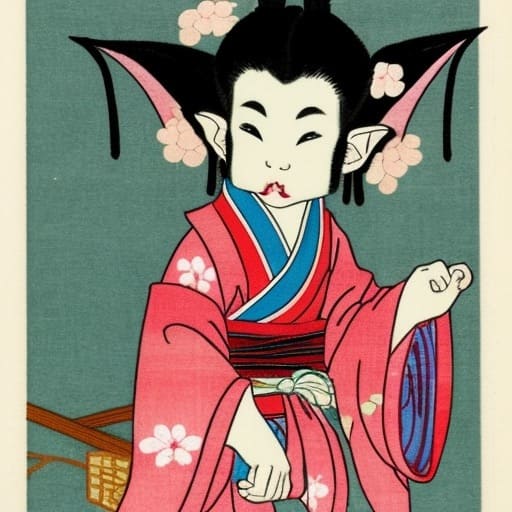}
    \end{minipage}%
    \begin{minipage}[t]{0.13\textwidth}
        \centering
        \includegraphics[width=\textwidth]{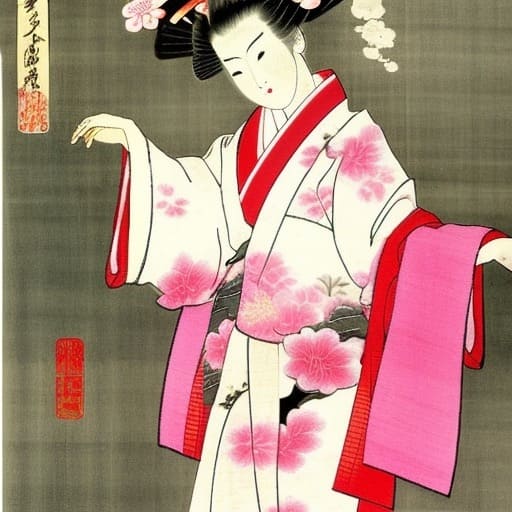}
    \end{minipage}%
    \begin{minipage}[t]{0.13\textwidth}
        \centering
        \includegraphics[width=\textwidth]{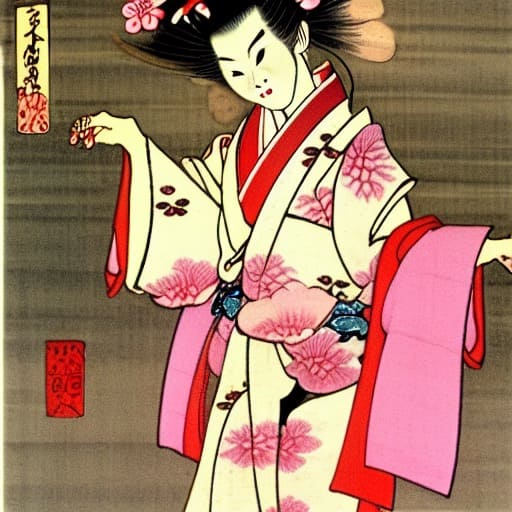}
    \end{minipage}%
    \begin{minipage}[t]{0.13\textwidth}
        \centering
        \includegraphics[width=\textwidth]{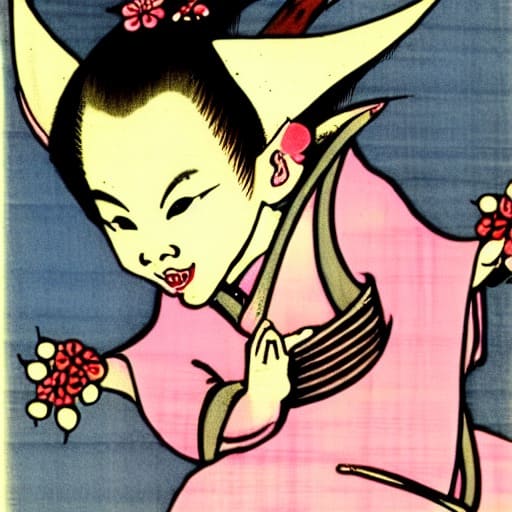}
    \end{minipage}%
    \begin{minipage}[t]{0.13\textwidth}
        \centering
        \includegraphics[width=\textwidth]{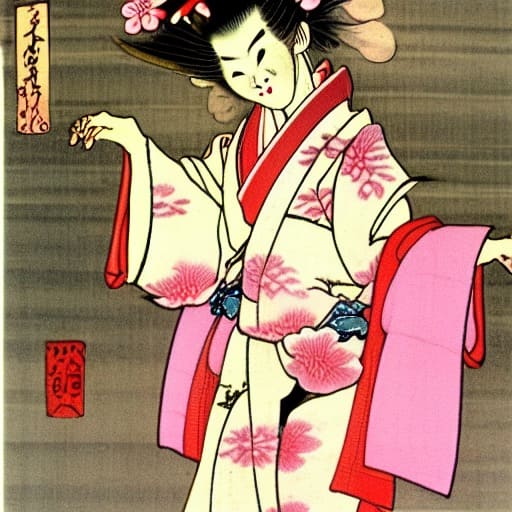}
    \end{minipage}%
    \begin{minipage}[t]{0.13\textwidth}
        \centering
        \includegraphics[width=\textwidth]{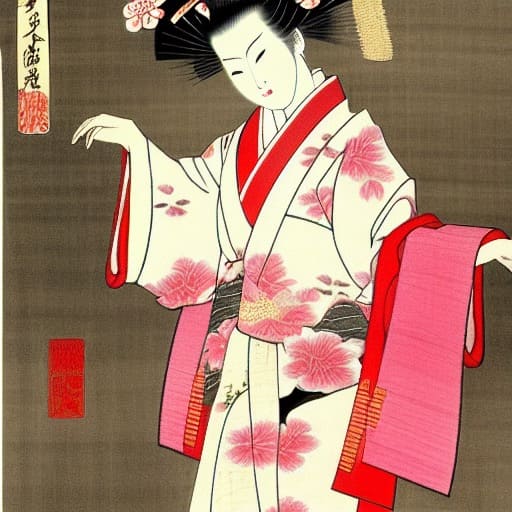}
    \end{minipage}%
    \par

    \begin{minipage}[t]{0.13\textwidth}
        \centering
        \includegraphics[width=\textwidth]{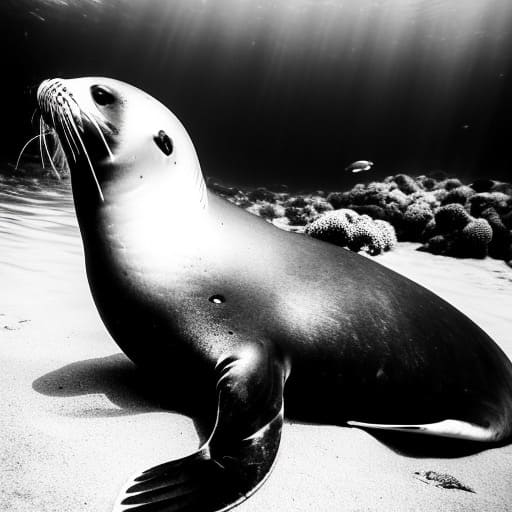}
    \end{minipage}%
    \begin{minipage}[t]{0.13\textwidth}
        \centering
        \includegraphics[width=\textwidth]{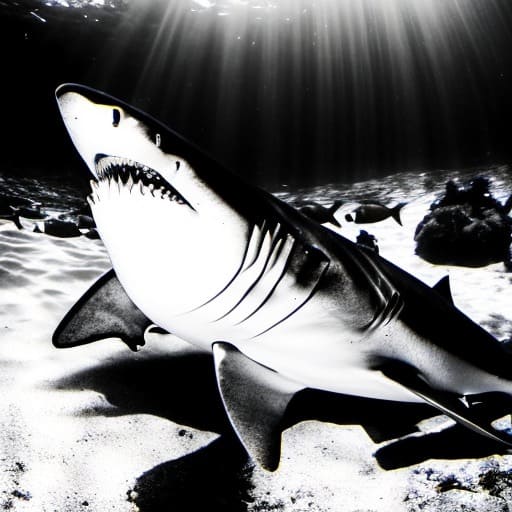}
    \end{minipage}%
    \begin{minipage}[t]{0.13\textwidth}
        \centering
        \includegraphics[width=\textwidth]{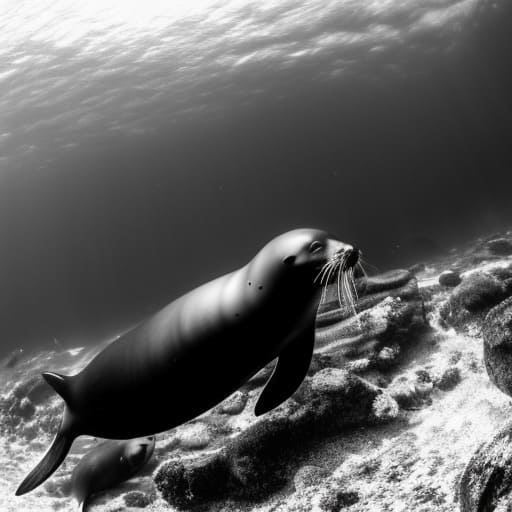}
    \end{minipage}%
    \begin{minipage}[t]{0.13\textwidth}
        \centering
        \includegraphics[width=\textwidth]{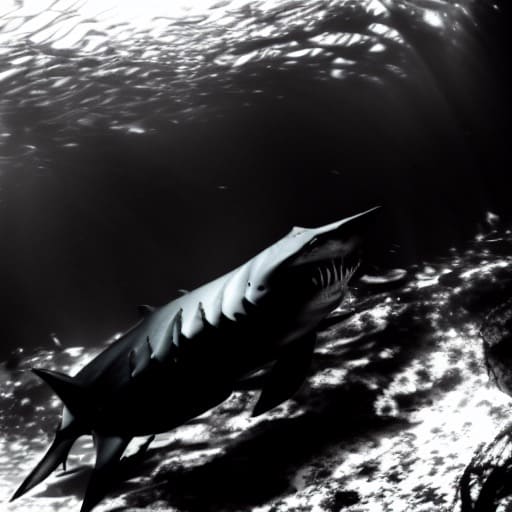}
    \end{minipage}%
    \begin{minipage}[t]{0.13\textwidth}
        \centering
        \includegraphics[width=\textwidth]{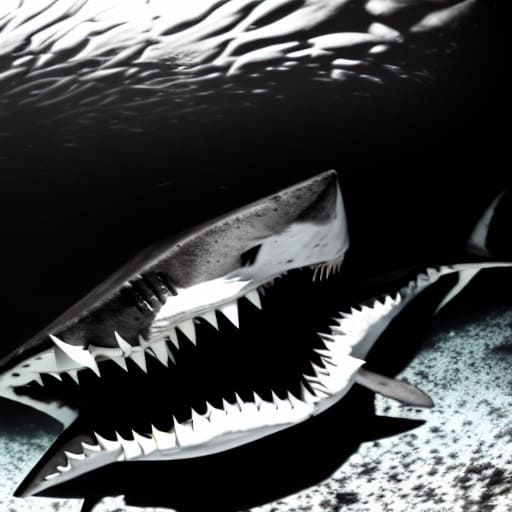}
    \end{minipage}%
    \begin{minipage}[t]{0.13\textwidth}
        \centering
        \includegraphics[width=\textwidth]{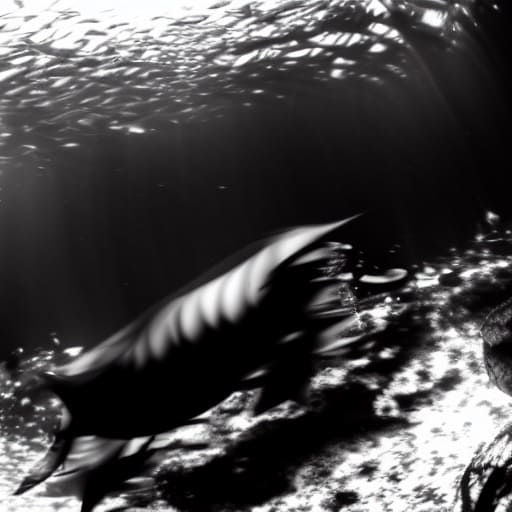}
    \end{minipage}%
    \begin{minipage}[t]{0.13\textwidth}
        \centering
        \includegraphics[width=\textwidth]{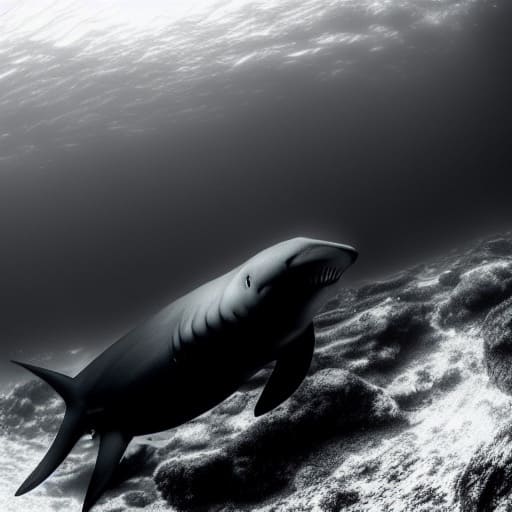}
    \end{minipage}%
    \par

    \begin{minipage}[t]{0.13\textwidth}
        \centering
        \includegraphics[width=\textwidth]{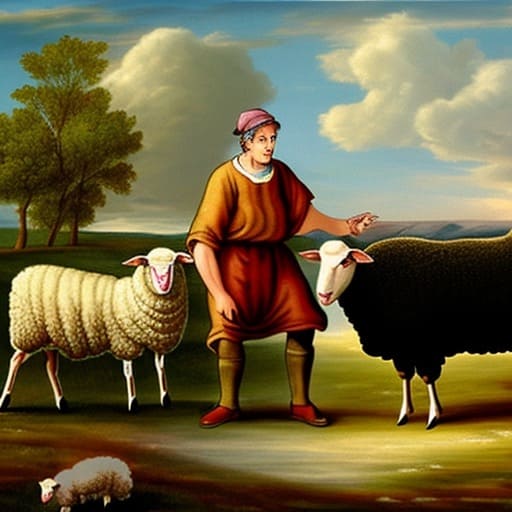}
    \end{minipage}%
    \begin{minipage}[t]{0.13\textwidth}
        \centering
        \includegraphics[width=\textwidth]{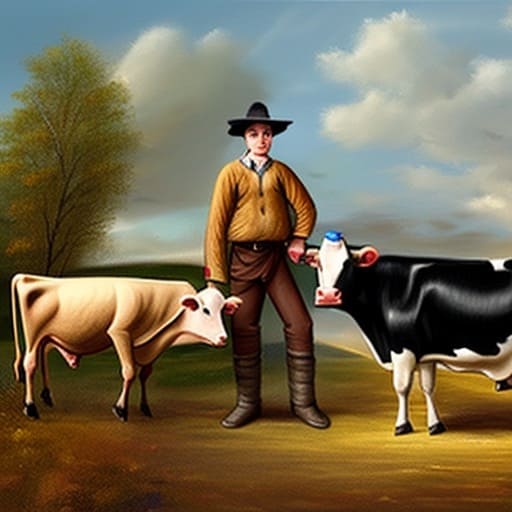}
    \end{minipage}%
    \begin{minipage}[t]{0.13\textwidth}
        \centering
        \includegraphics[width=\textwidth]{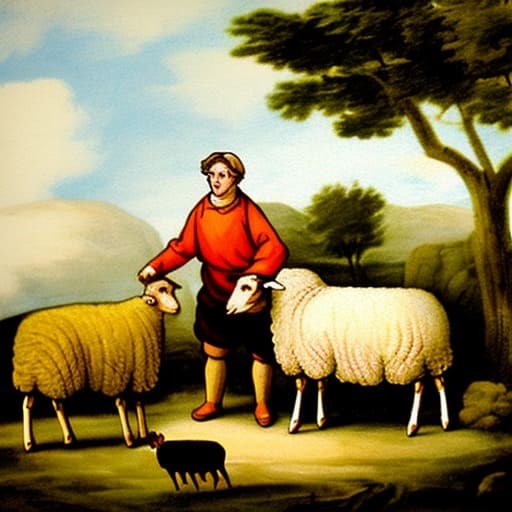}
    \end{minipage}%
    \begin{minipage}[t]{0.13\textwidth}
        \centering
        \includegraphics[width=\textwidth]{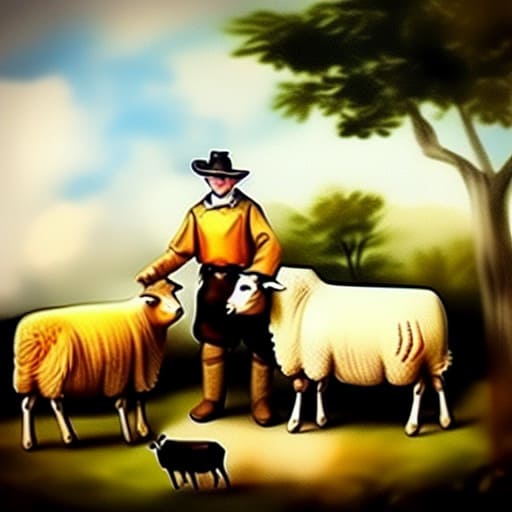}
    \end{minipage}%
    \begin{minipage}[t]{0.13\textwidth}
        \centering
        \includegraphics[width=\textwidth]{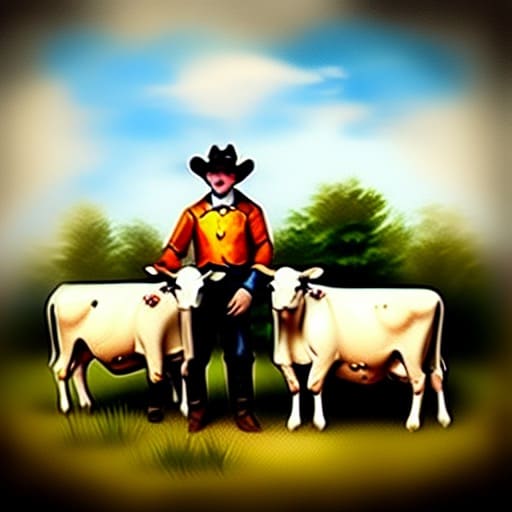}
    \end{minipage}%
    \begin{minipage}[t]{0.13\textwidth}
        \centering
        \includegraphics[width=\textwidth]{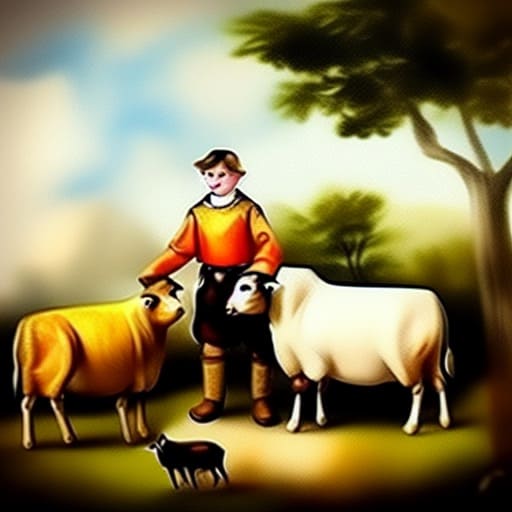}
    \end{minipage}%
    \begin{minipage}[t]{0.13\textwidth}
        \centering
        \includegraphics[width=\textwidth]{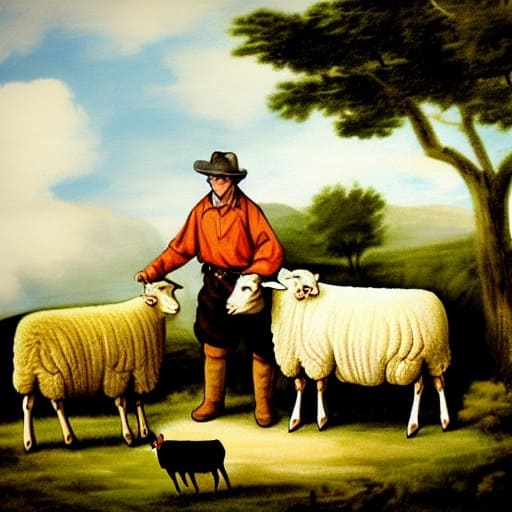}
    \end{minipage}%
    \par

    \begin{minipage}[t]{0.13\textwidth}
        \centering
        \includegraphics[width=\textwidth]{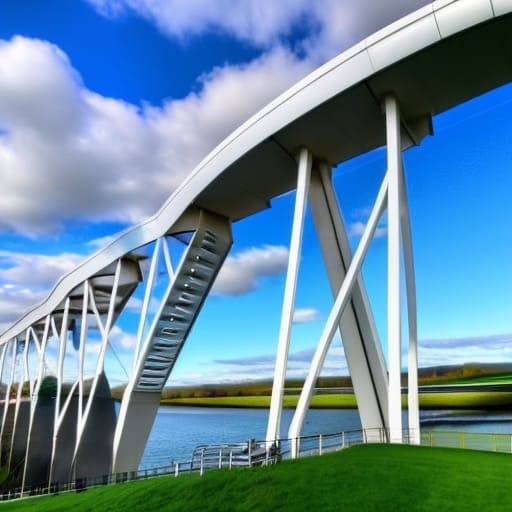}
    \end{minipage}%
    \begin{minipage}[t]{0.13\textwidth}
        \centering
        \includegraphics[width=\textwidth]{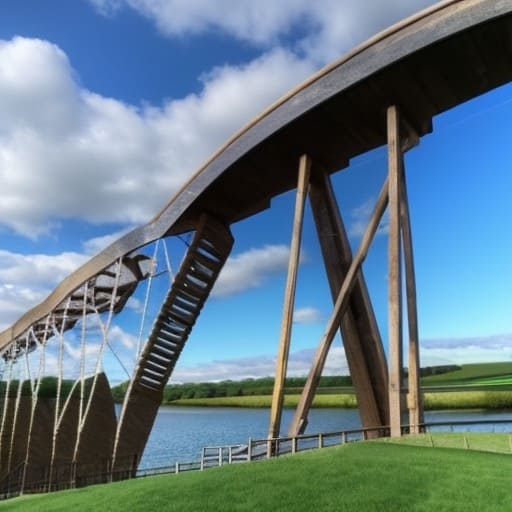}
    \end{minipage}%
    \begin{minipage}[t]{0.13\textwidth}
        \centering
        \includegraphics[width=\textwidth]{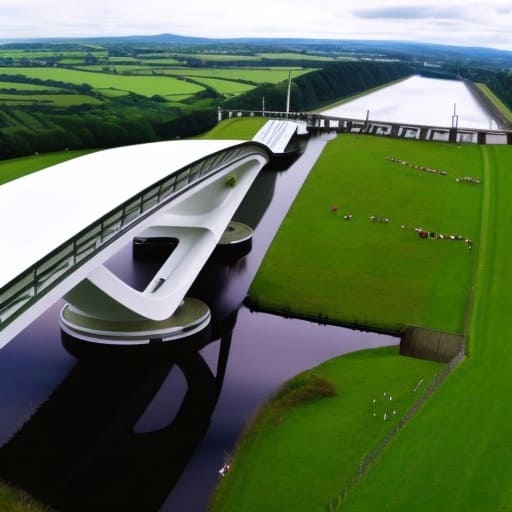}
    \end{minipage}%
    \begin{minipage}[t]{0.13\textwidth}
        \centering
        \includegraphics[width=\textwidth]{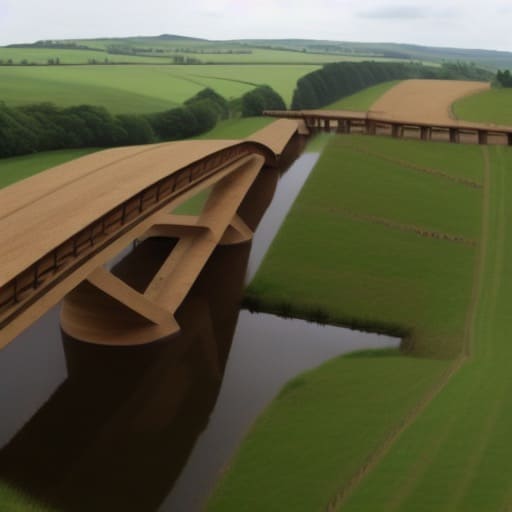}
    \end{minipage}%
    \begin{minipage}[t]{0.13\textwidth}
        \centering
        \includegraphics[width=\textwidth]{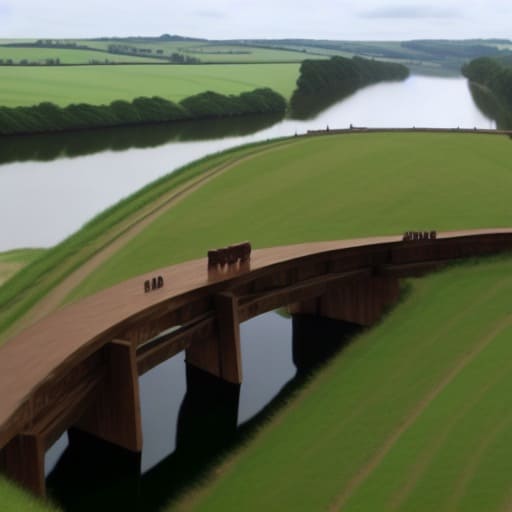}
    \end{minipage}%
    \begin{minipage}[t]{0.13\textwidth}
        \centering
        \includegraphics[width=\textwidth]{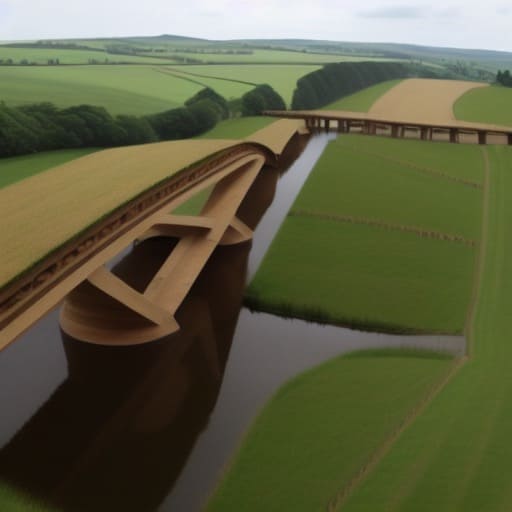}
    \end{minipage}%
    \begin{minipage}[t]{0.13\textwidth}
        \centering
        \includegraphics[width=\textwidth]{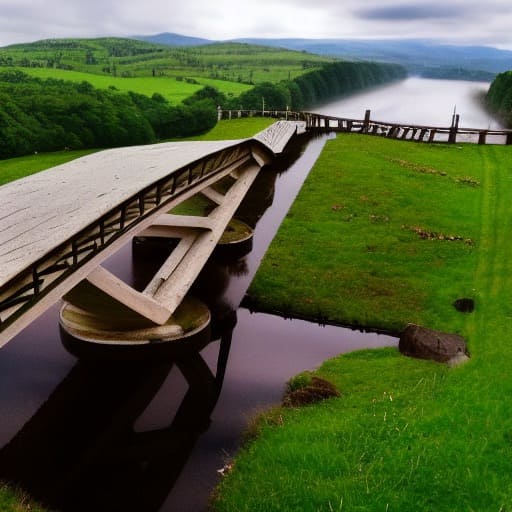}
    \end{minipage}%
    \par

    \caption{Qualitative results of \method with and without its key components. Clearly, \method outperforms all other variations in terms of adhering faithfully to the edit illustrated in the exemplar without distorting the test image unnecessarily. Use high levels of magnifications to observe subtle edits. \label{fig:ablations}}
\end{figure*}
\begin{figure*}[h!]
    \centering
    \begin{minipage}[t]{0.13\textwidth}
        \centering
        \subcaption[]{$x$}{}
        \includegraphics[width=\textwidth]{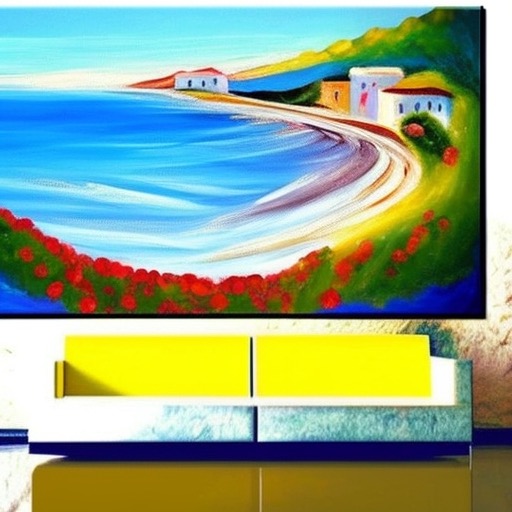}
    \end{minipage}%
    \begin{minipage}[t]{0.13\textwidth}
        \centering
        \subcaption[]{$x_{\text{edit}}$}
        \includegraphics[width=\textwidth]{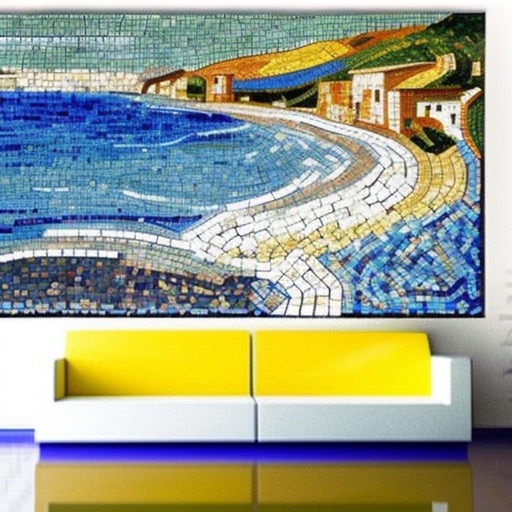}
    \end{minipage}%
    \begin{minipage}[t]{0.13\textwidth}
        \centering
        \subcaption[]{$y$}{}
        \includegraphics[width=\textwidth]{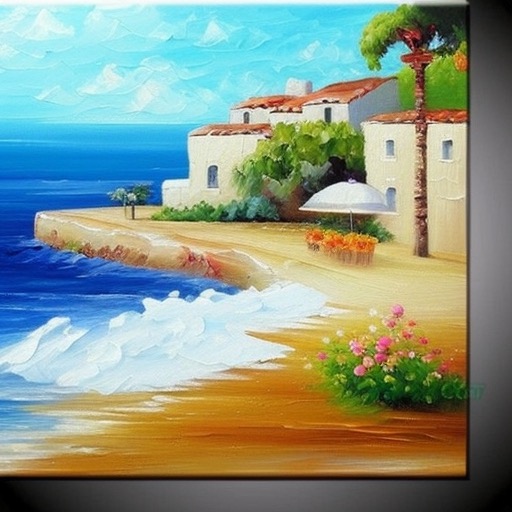}
    \end{minipage}%
    \begin{minipage}[t]{0.13\textwidth}
        \centering
        \subcaption[]{$\lambda=0$}{}
        \includegraphics[width=\textwidth]{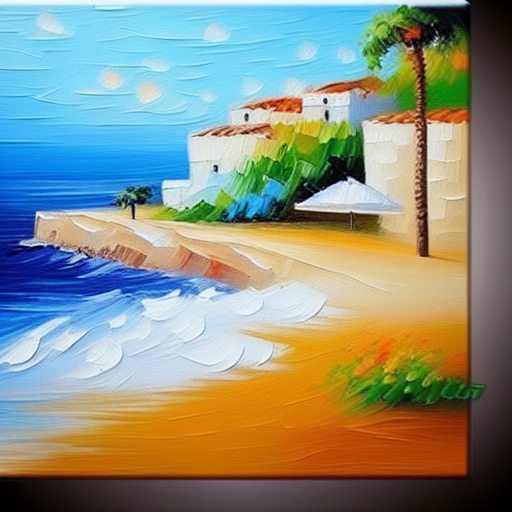}
    \end{minipage}%
   \begin{minipage}[t]{0.13\textwidth}
        \centering
        \subcaption[]{$\lambda=0.6$}{}
        \includegraphics[width=\textwidth]{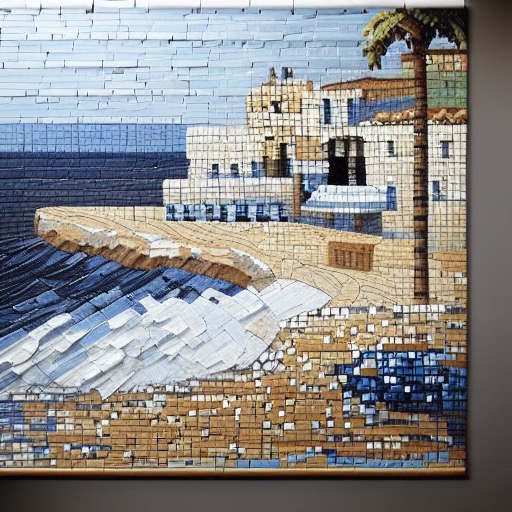}
    \end{minipage}%
       \begin{minipage}[t]{0.13\textwidth}
        \centering
        \subcaption[]{$\lambda=0.7$}{}
        \includegraphics[width=\textwidth]{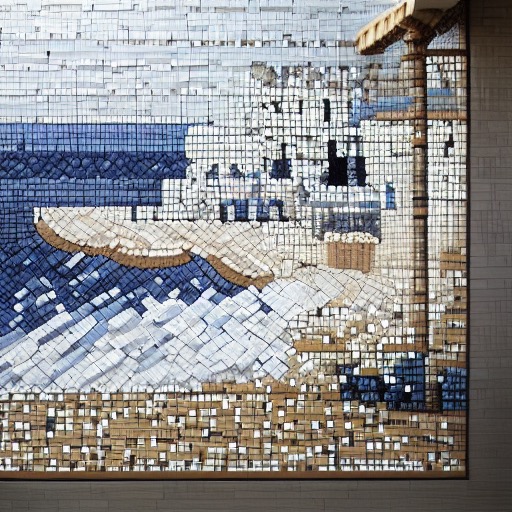}
    \end{minipage}%
       \begin{minipage}[t]{0.13\textwidth}
        \centering
        \subcaption[]{$\lambda=0.8$}{}
        \includegraphics[width=\textwidth]{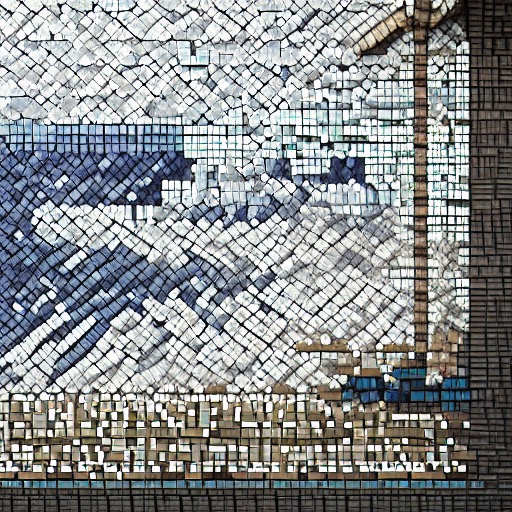}
    \end{minipage}%

    \begin{minipage}[t]{0.13\textwidth}
        \centering
        \includegraphics[width=\textwidth]{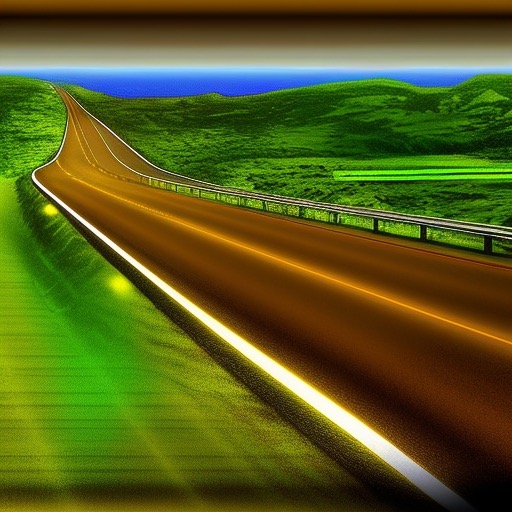}
    \end{minipage}%
    \begin{minipage}[t]{0.13\textwidth}
        \centering
        \includegraphics[width=\textwidth]{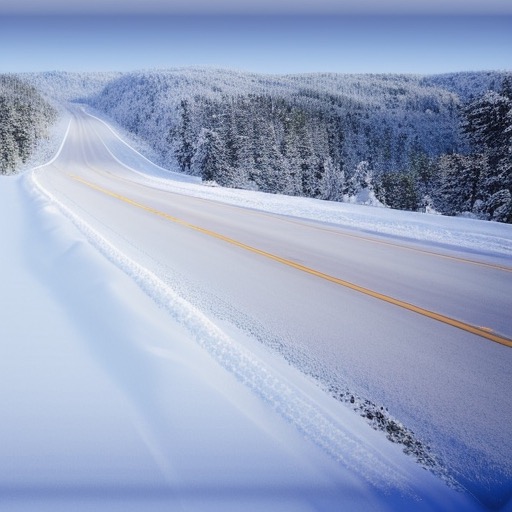}
    \end{minipage}%
    \begin{minipage}[t]{0.13\textwidth}
        \centering
        \includegraphics[width=\textwidth]{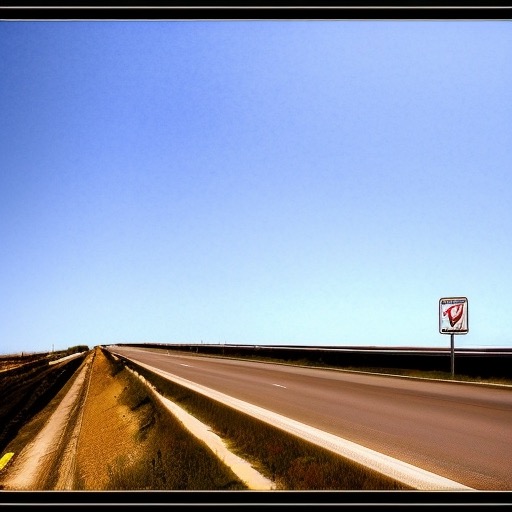}
    \end{minipage}%
    \begin{minipage}[t]{0.13\textwidth}
        \centering
        \includegraphics[width=\textwidth]{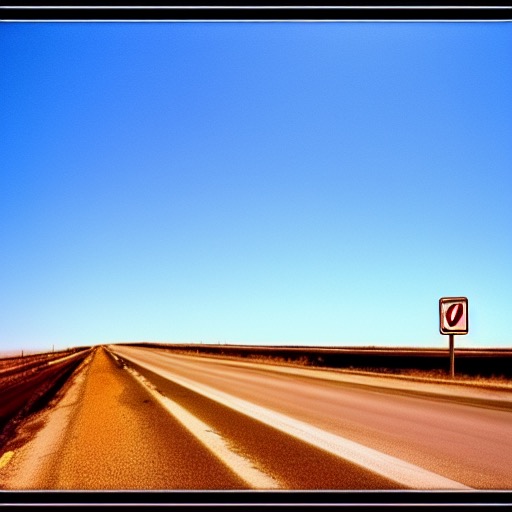}
    \end{minipage}%
   \begin{minipage}[t]{0.13\textwidth}
        \centering
        \includegraphics[width=\textwidth]{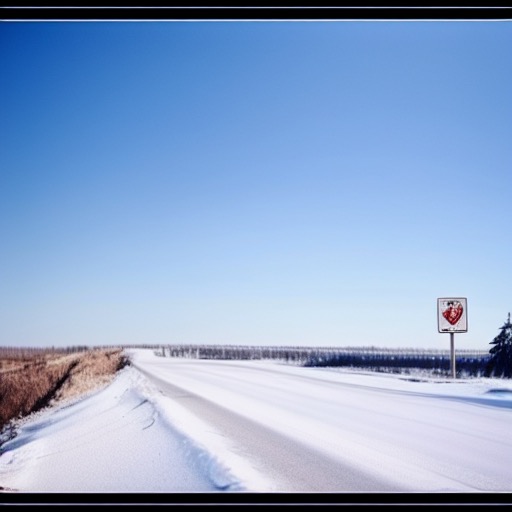}
    \end{minipage}%
       \begin{minipage}[t]{0.13\textwidth}
        \centering
        \includegraphics[width=\textwidth]{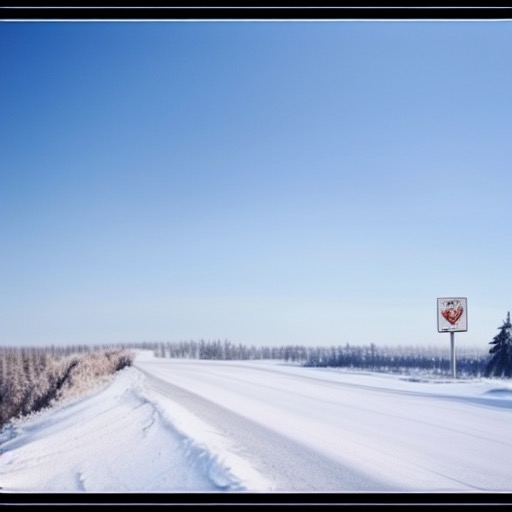}
    \end{minipage}%
       \begin{minipage}[t]{0.13\textwidth}
        \centering
        \includegraphics[width=\textwidth]{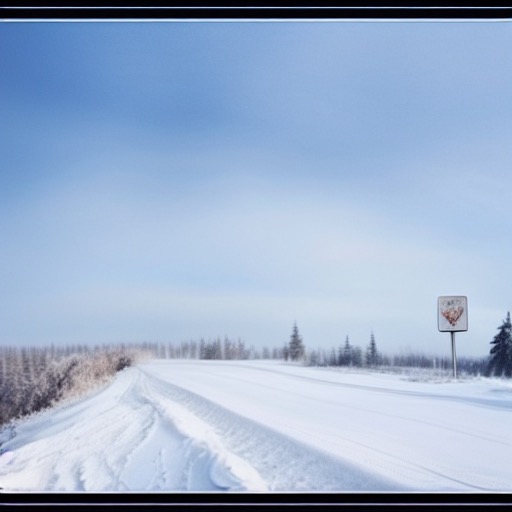}
    \end{minipage}%

    \par
    \caption{Depiction of the effect of edit weight $\lambda$ on the edited image. higher values correspond to higher influence from the desired edit.}
    \label{fig:slider}
\end{figure*}

This section comprises of detailed experiments and ablations of our framework, \method. We ablate on the key aspects in our approach - \textbf{a.} \textbf{LLaVa Text} ($g_{\text{caption}}$), \textbf{b.} \textbf\\{Guidance} ($f, Q, K$ injection), and \textbf{c.} \textbf{Clip Difference} ($\Delta_{\text{img}}$). The results are presented in Fig.~\ref{fig:ablations}. Additionally, we show the effect of varying our the edit weight ($\lambda$) in Fig.~\ref{fig:slider}. We discuss both sets of results in detail below

\noindent{\xhdr{Impact of LLaVA Text ($g_{\text{caption}}$)}}
Our approach leverages both image and text guidance to sufficiently capture all aspects of the edit from the exemplar pair $(x, x_\text{edit})$. There is a synergestic relation between both guidances, filling in the gaps left by the other. This is clearly seen in Row 2, where removing small details like the flowers atop the cactii are missed in the absence of $g_\text{caption}$, and in Row 4, where the LLaVA edit text instructs \method to include the sharp teeth of the shark, a subtle cue not captured in its absence.

\noindent{\xhdr{Impact of Guidance ($f, Q, K$)}} 
Although $\Delta_{\text{img}}$ and DDIM inversion both provide cues to maintain the structure of the original image, this is enforced much more robustly through feature injection and self attention injection. In the absense of this component, the edited images fail to maintain the structure from the original image, even though stylistic cues are well captured. This is evident from Row 3, where the person is no longer the same, and Rows 4-8, where the structure of the edited object has been completely destroyed. 

\noindent{\xhdr{Impact of Image Clip Difference ($\Delta_{\text{img}}$)}} 
This key component of \method captures all types of nuanced details about the edit from the exemplar pair $(x, x_\text{edit})$. As we posit in Sec.~\ref{sec:method}, certain aspects cannot be sufficiently explained through text, which is where this component because increasingly important. The effect is clearly visible across multiple examples. In Row 1, removing $\Delta_\text{img}$ causes the edited image to miss the required style, and instead simply mimics a generic 'caricature'. In Row 2, it is key in capturing the exact required style. The subtle change to a wooden structure in Row 6 is also perfectly capture only in the presence of $\Delta_\text{img}$. It is easy for the LLaVA generated edit text to miss this detail, while it is easily picked up when analyzing the edit in the latent image space.

\noindent{\xhdr{Impact of edit weight ($\lambda$)}}
The only hyperparameter to be set when performing inference time edits using \method is $\lambda$, the edit weight. This acts as a mixing ratio, weighing the contributions of the exemplar pair $(x, x_\text{edit})$ and input image $y$ when generating the final output. The effect of this is portrayed in Fig.~\ref{fig:slider}. As we vary the value of $\lambda$, we observe a smooth increase in the infuence of the desired edit on the target image. This is a convenient way for a practitioner to exercise control over the editing process. In practice, we find that a value of $0.65$ works best across varied types of edits, and this is also the value that yields the best quantitative results.

\section{Conclusion}
\label{sec:conclusion}
In this paper, we introduce \method, an efficient, optimization-free framework for exemplar-based image editing. We motivate that precise edits cannot be captured by textual modality alone, and propose a novel strategy that leverages the reasoning capabilities of VLMs, and edits in image space to capture the desired user intent using exemplar pairs. Our results demonstrate \method's practical applicability because of its speed and ease of use compared to strong baselines. Our results also position our method as the state of the art both quantitatively and qualitatively. We hope our findings motivate further research in this area.  


{\small
\bibliographystyle{ieee_fullname}
\bibliography{main}
}

\clearpage
\appendix
\section{Appendix}

The appendix is structured as follows - In Appendix~\ref{sec:addn-llava} we include additional details of leveraging LLaVA to generate optimal edit instructions. We provide further descriptions of the metrics used for our quantitative analysis in Appendix~\ref{sec:metrics}. Appendix~\ref{sec:addn-qual} contains additional qualitative comparisons between \method and the various baselines. Finally, Appendix~\ref{sec:ip2p-failure} contains additional examples showing poor and ambiguous samples from the InstructPix2Pix dataset, and Appendix~\ref{sec:limit} discusses the current limitations of \method. 

\begin{figure*}[htp]
    \centering
    \includegraphics[width=\textwidth]{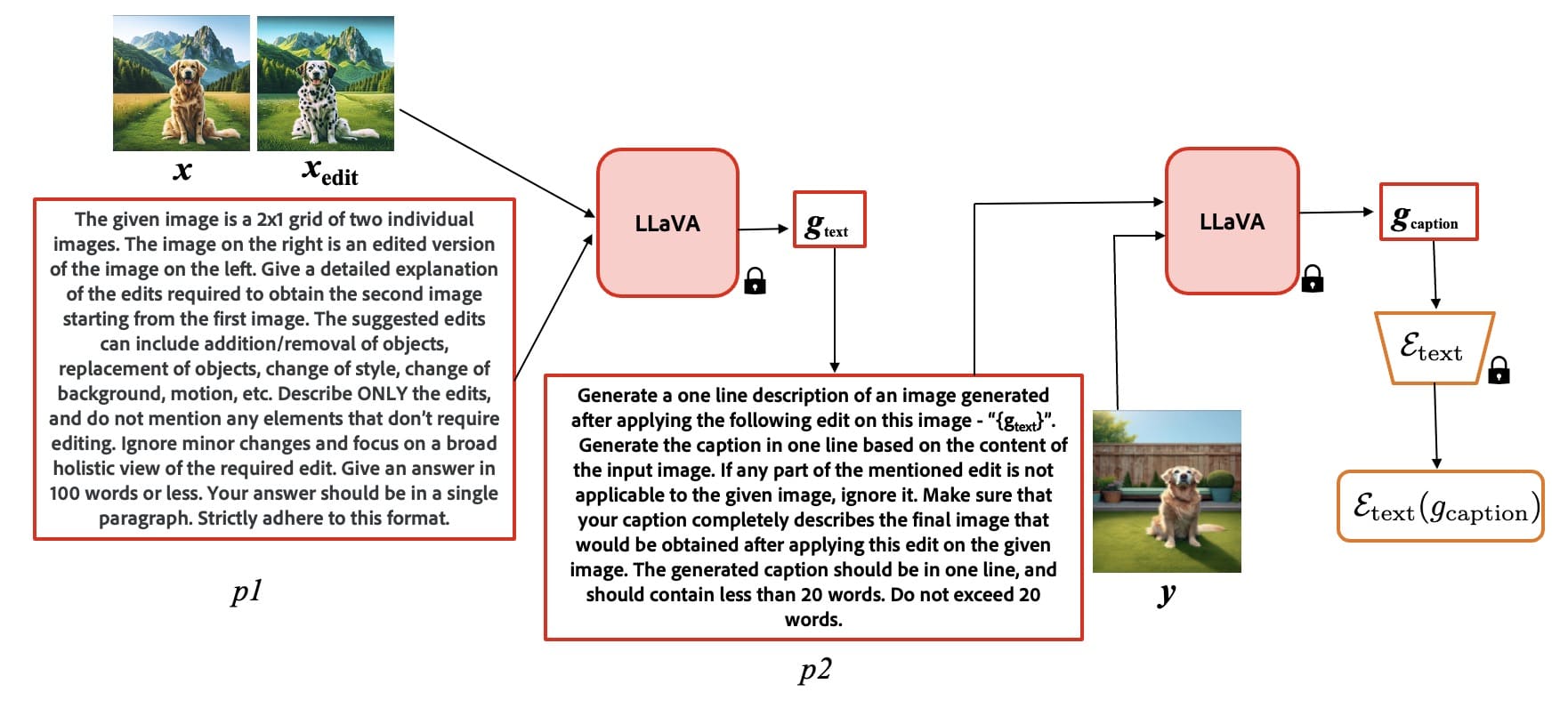}
    \caption{Overview of generating text-based edits using multimodal VLMS. \textbf{a.} In the first step, we input a detailed prompt $p1$, and a grid of exemplar pairs. The output $g_{\text{text}}$ is then curated in the form of another prompt $p2$ which is passed as input to LLaVA with image $y$ to generate $g_{\text{caption}}$. Note that all models are frozen and are used in inference mode.}
    \label{fig:llava}
\end{figure*}

\subsection{Additional details of LLaVA-based edits}
\label{sec:addn-llava}
Below are the prompts $p1$, $p2$, and $p3$ that are used as input to LLaVA in various stages of our method and baselines. Fig.~\ref{fig:llava} summarizes the pipeline for generating captions and edit instruction using LLaVA.
\begin{enumerate}
    \item \textbf{p1} \textit{The given image is a 2x1 grid of two individual images. The image on the right is an edited version of the image on the left. Give a detailed explanation of the edits required to obtain the second image starting from the first image. The suggested edits can include addition/removal of objects, replacement of objects, change of style, change of background, motion, etc. Describe ONLY the edits, and do not mention any elements that don’t require editing. Ignore minor changes and focus on a broad holistic view of the required edit. Give an answer in 100 words or less. Your answer should be in a single paragraph. Strictly adhere to this format.}
    \item \textbf{p2} \textit{Generate a one line description of an image generated after applying the following edit on this image - “$<$Response from LLaVA using p1$>$”.
    Generate the caption in one line based on the content of the input image. If any part of the mentioned edit is not applicable to the given image, ignore it. Make sure that your caption completely describes the final image that would be obtained after applying this edit on the given image. The generated caption should be in one line, and should contain less than 20 words. Do not exceed 20 words.}
    \item \textbf{p3} \textit{Generate a one line edit instruction to edit the given image. The edit should follow the instruction in this longer edit - “$<$Response from LLaVA using p1$>$”
    Generate the edit instruction in a single line based on the content of the input image. If any part of the mentioned image is not applicable to the given image, ignore it. Make sure that your instruction is sufficient to replicate the describe edit. The generated instruction should be in one line, and should contain less than 20 words. Do not exceed 20 words.}
\end{enumerate}

\subsection{Details about Metrics} 
\label{sec:metrics}
In this work, we use several image quality assessment metrics. Each metric provides a measure of a different aspect of the generation, refer to Table~\ref{tab:quant} for the average performance of \method on our entire dataset of ~1500 images. $\downarrow, \uparrow$ denote that a lower value of the metric is better and a higher value of the metric is better respectively. 


\noindent{\xhdr{a. LPIPS ($\downarrow$)}} The Learned Perceptual Image Patch Similarity~\cite{zhang2018unreasonable} calculates perceptual similarity between two images (here, $\hat{y}_{\text{edit}}, y_{\text{edit}}$) by comparing the deep features of two images. Traditionally, VGG~\cite{simonyan2014very} has been used to compute these features. This makes LPIPS more aligned with human visual perception, capturing subtle differences that traditional metrics like PSNR and SSIM might miss. Lower LPIPS values indicate higher similarity between images.

\noindent{\xhdr{b. SSIM ($\uparrow$)}}~\cite{wang2004image} is a measure of Structural Similarity between two images. A higher SSIM score generally indicates higher structural similarity. In Table~\ref{tab:quant}, we report the structural similarity (SSIM) between $\hat{y}_{\text{edit}}, y_{\text{edit}}$. A higher value of SSIM indicates that the edit has been performed correctly on $y$.

\noindent{\xhdr{c. CLIP Score ($\uparrow$)}} This score~\cite{hessel2021clipscore} is a reference-free metric that measures the alignment between images and textual descriptions. Specifically, in our paper, it corresponds to the cosine similarity (normalized dot product) of $\hat{y}_{\text{edit}}, \mathcal{E}_{\text{text}}(g_{\text{caption}})$ where $\mathcal{E}_{\text{text}}(g_{\text{caption}})$ is the clip text embedding of the generated caption and the generated image. 


\noindent{\xhdr{d. Directional Similarity ($\uparrow$)}} StyleGAN-Nada~\cite{gal2021stylegan} proposed a directional CLIP similarity measure that measures the cosine similarity between the difference of edited and un-edited image ($\hat{y}_{\text{edit}} - y$), and the caption ($\mathcal{E}_{\text{text}}(g_{\text{caption}})$). A higher similarity indicates that the edit performed is in the direction of the text.

\noindent{\xhdr{e. S-Visual ($\uparrow$)}} Metric proposed in the baseline VISII~\cite{nguyen2024visual} which computes the cosine similarity between the difference between the clip embeddings of the exemplar pair, and the difference between the clip embeddings of test image $y$ and the generated image $\hat{y}_{\text{edit}}$. It is noteworthy that VISII optimizes the same function they use as a metric.



\subsection{Additional Qualitative Results}
\label{sec:addn-qual}
Figs.~\ref{fig:addn-qual1}~\ref{fig:addn-qual2} provides additional qualitative comparisons, highlighting the efficacy of \method in exemplar-based image editing. Specifically, \method \emph{outperforms strong baselines across various types of edits}, including \textbf{a.} global style transfer, \textbf{b.} local style transfer, \textbf{c.} object replacement, and \textbf{d.} object addition.


\subsection{Examples of poor samples in IP2P dataset}
\label{sec:ip2p-failure}
\begin{figure*}[h]
\centering
\begin{minipage}{.13\textwidth}
\centering
\( \mathbf{x} \)
\end{minipage}%
\hspace{5pt}
\begin{minipage}{.13\textwidth}
\centering
\textbf{Edit instruction}
\end{minipage}%
\hspace{5pt}
\begin{minipage}{.13\textwidth}
\centering
\(\mathbf{x_{\textbf{edit}}}\)
\end{minipage}%
\hspace{40pt}
\hspace{5pt}
\begin{minipage}{.13\textwidth}
\centering
\( \mathbf{y} \)
\end{minipage}%
\hspace{5pt}
\begin{minipage}{.13\textwidth}
\centering
\textbf{Edit instruction}
\end{minipage}%

\vspace{5pt}
\centering
\begin{minipage}{.13\textwidth}
\includegraphics[width=\textwidth]{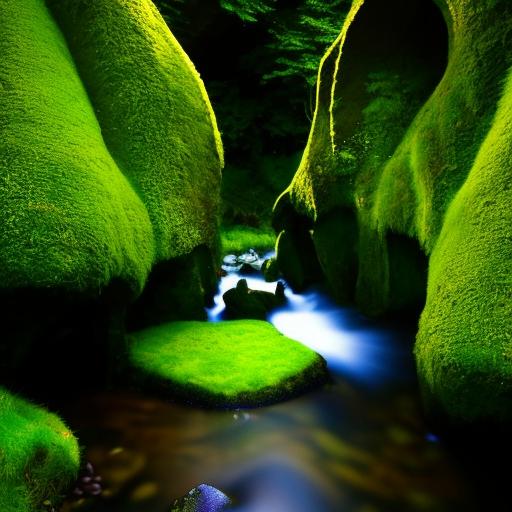}
\end{minipage}%
\hspace{5pt} \begin{minipage}{.13\textwidth}
\centering
\textit{Add a horse}
\end{minipage}%
\hspace{5pt} \begin{minipage}{.13\textwidth}
\includegraphics[width=\textwidth]{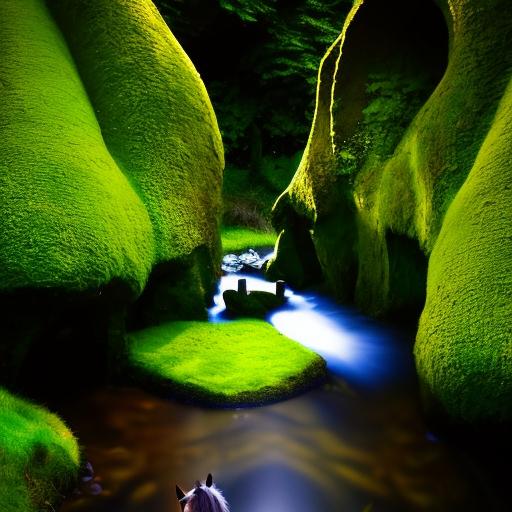}
\end{minipage}%
\hspace{40pt}
\hspace{5pt} \begin{minipage}{.13\textwidth}
\includegraphics[width=\textwidth]{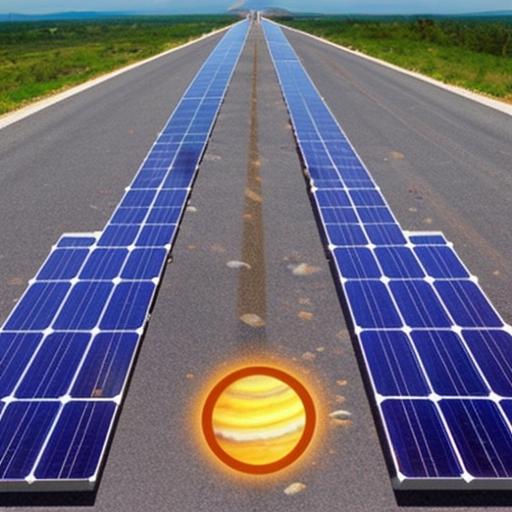}
\end{minipage}%
\hspace{5pt} \begin{minipage}{.13\textwidth}
\centering
\textit{Turn it into a shopping mall}
\end{minipage}%

\vspace{5pt}
\centering
\begin{minipage}{.13\textwidth}
\includegraphics[width=\textwidth]{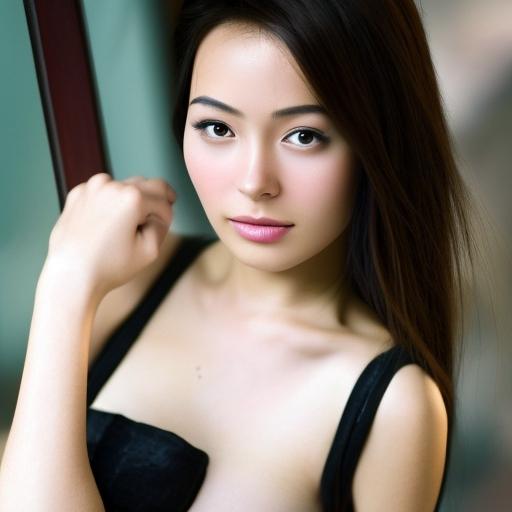}
\end{minipage}%
\hspace{5pt} \begin{minipage}{.13\textwidth}
\centering
\textit{Make her a panda}
\end{minipage}%
\hspace{5pt} \begin{minipage}{.13\textwidth}
\includegraphics[width=\textwidth]{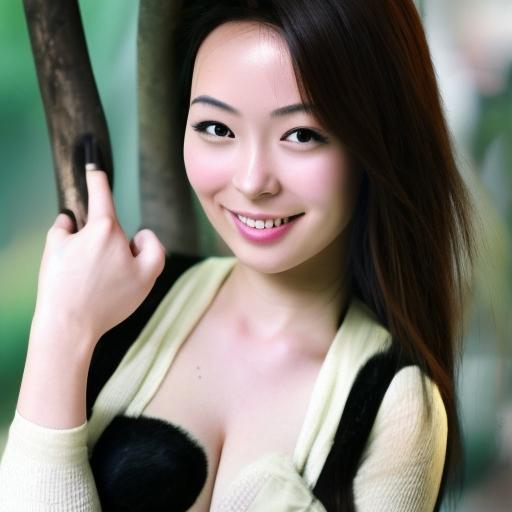}
\end{minipage}%
\hspace{40pt}
\hspace{5pt} \begin{minipage}{.13\textwidth}
\includegraphics[width=\textwidth]{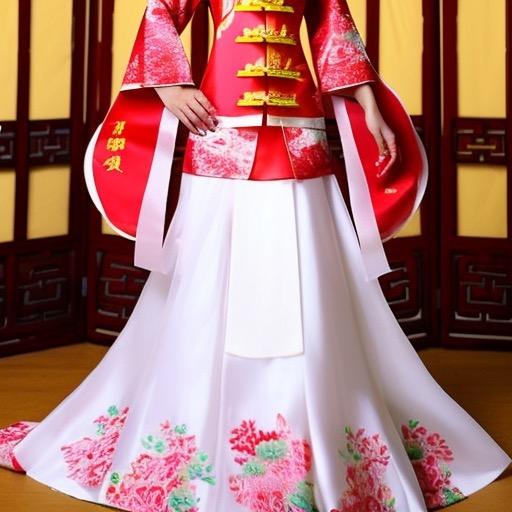}
\end{minipage}%
\hspace{5pt} \begin{minipage}{.13\textwidth}
\centering
\textit{Make her wear a crown}
\end{minipage}%

\vspace{5pt}
\centering
\begin{minipage}{.13\textwidth}
\includegraphics[width=\textwidth]{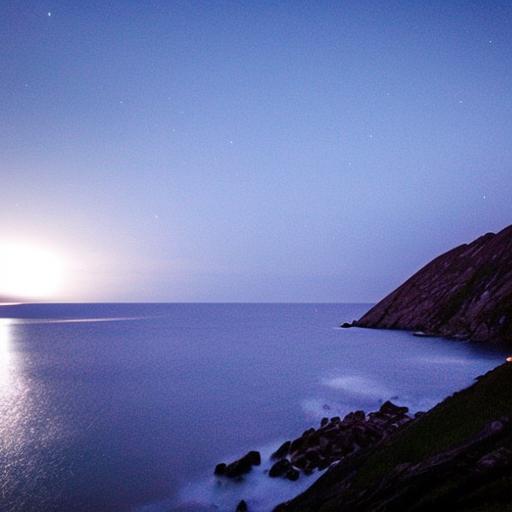}
\end{minipage}%
\hspace{5pt} \begin{minipage}{.13\textwidth}
\centering
\textit{Make it a stormy night}
\end{minipage}%
\hspace{5pt} \begin{minipage}{.13\textwidth}
\includegraphics[width=\textwidth]{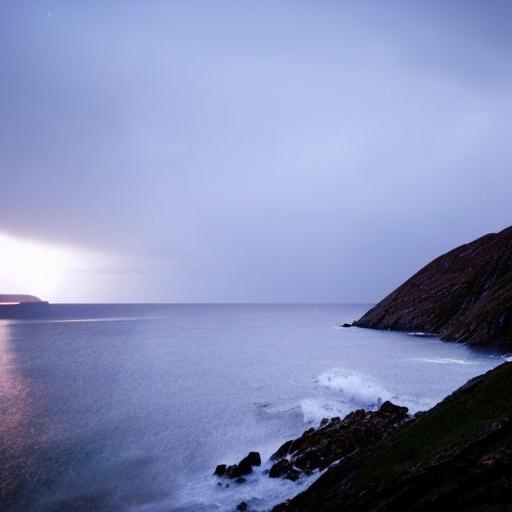}
\end{minipage}%
\hspace{40pt}
\hspace{5pt} \begin{minipage}{.13\textwidth}
\includegraphics[width=\textwidth]{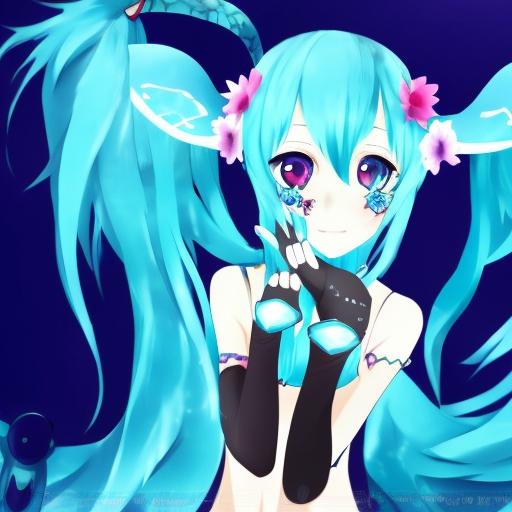}
\end{minipage}%
\hspace{5pt} \begin{minipage}{.13\textwidth}
\centering
\textit{Turn the animal into a dragon}
\end{minipage}%

\vspace{5pt}
\centering
\begin{minipage}{.13\textwidth}
\includegraphics[width=\textwidth]{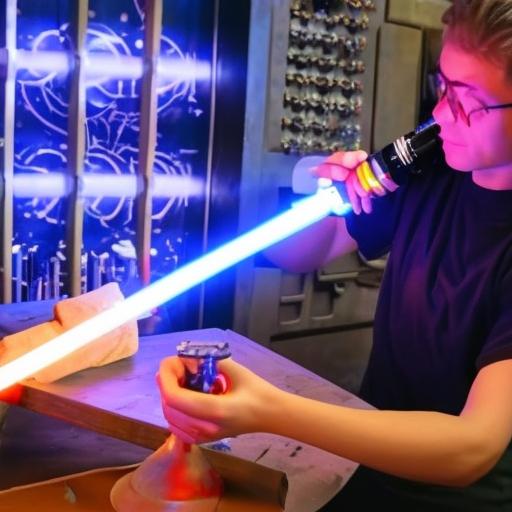}
\end{minipage}%
\hspace{5pt} \begin{minipage}{.13\textwidth}
\centering
\textit{Have it made of wood}
\end{minipage}%
\hspace{5pt} \begin{minipage}{.13\textwidth}
\includegraphics[width=\textwidth]{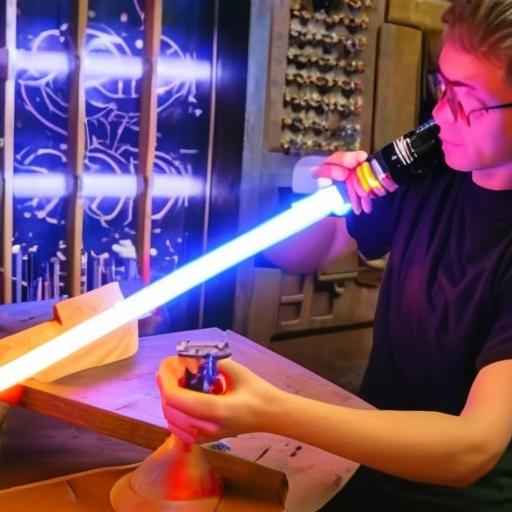}
\end{minipage}%
\hspace{40pt}
\hspace{5pt} \begin{minipage}{.13\textwidth}
\includegraphics[width=\textwidth]{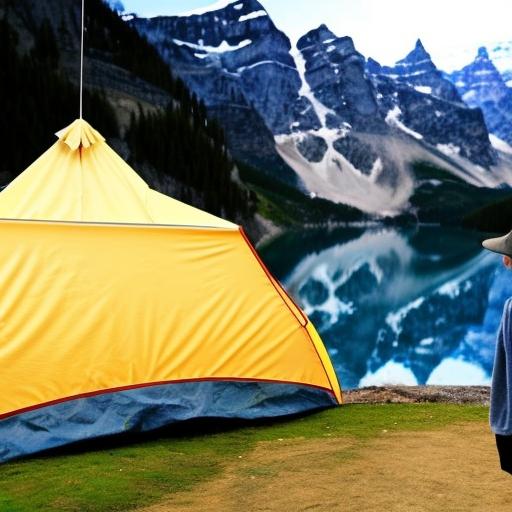}
\end{minipage}%
\hspace{5pt} \begin{minipage}{.13\textwidth}
\centering
\textit{make him a biologist}
\end{minipage}%

\vspace{5pt}
\centering
\hspace{5pt} \begin{minipage}{.39\textwidth}
\centering
\textbf{(a)}
\end{minipage}%
\hspace{40pt}
\hspace{5pt} \begin{minipage}{.26\textwidth}
\centering
\textbf{(b)}
\end{minipage}%

\caption {Images illustrating failure of automated dataset generation. \textbf{a}: Cases where exemplar pair $x, x_{\text{edit}}$ does not represent expected edit. \textbf{b}: Cases where test image $y$ does not conform with edit}
\label{fig:p2p_failure_appendix}
\end{figure*}

We present additional examples of poor and ambiguous samples from the InstructPix2Pix dataset in Fig.~\ref{fig:p2p_failure_appendix}. We noticed a number of these samples, necessitating the manual curation of our evaluation dataset, as described in Sec.~\ref{sec:dataset}.

\begin{figure*}
    \centering
    \begin{minipage}[t]{0.13\textwidth}
        \centering
        \subcaption[]{$x$}
        \includegraphics[width=\textwidth]{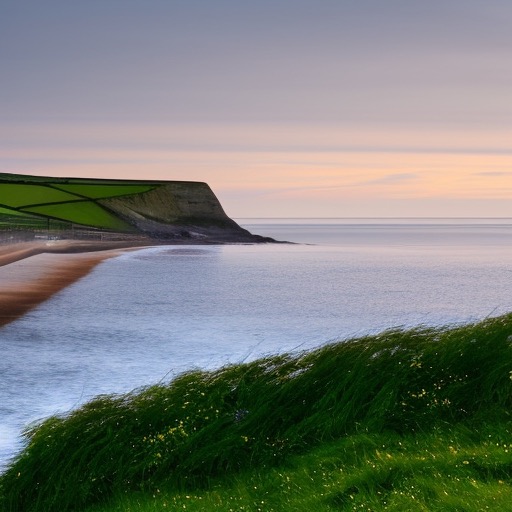}
    \end{minipage}%
    \begin{minipage}[t]{0.13\textwidth}
        \centering
        \subcaption[]{$x_{\text{edit}}$}
        \includegraphics[width=\textwidth]{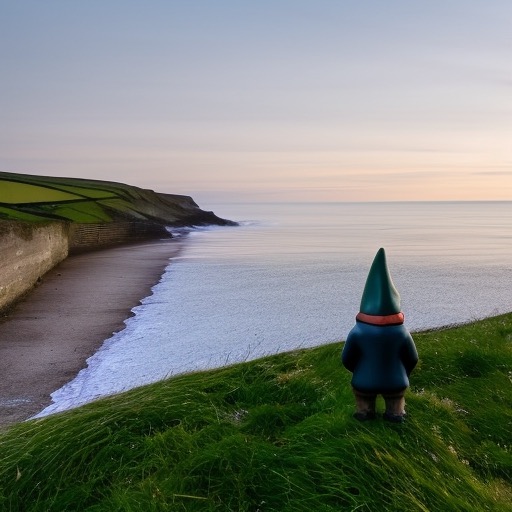}
    \end{minipage}%
    \begin{minipage}[t]{0.13\textwidth}
        \centering
        \subcaption[]{$y$}
        \includegraphics[width=\textwidth]{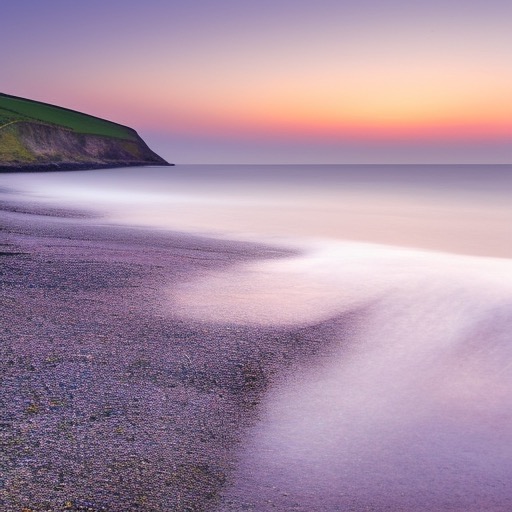}
    \end{minipage}%
    \begin{minipage}[t]{0.13\textwidth}
        \centering
        \subcaption[]{\method}
        \includegraphics[width=\textwidth]{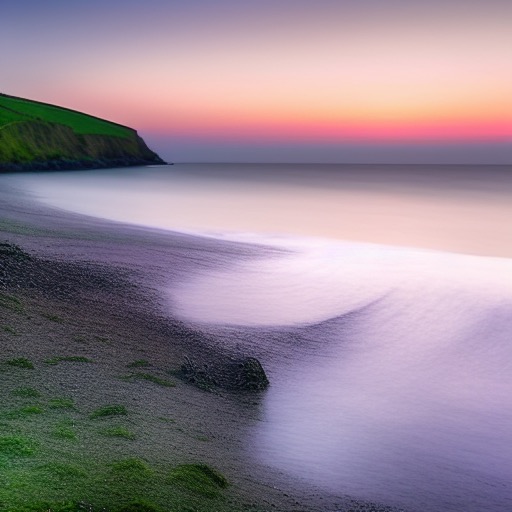}
    \end{minipage}%
    \begin{minipage}[t]{0.13\textwidth}
        \centering
        \subcaption[]{VISII}{}
        \includegraphics[width=\textwidth]{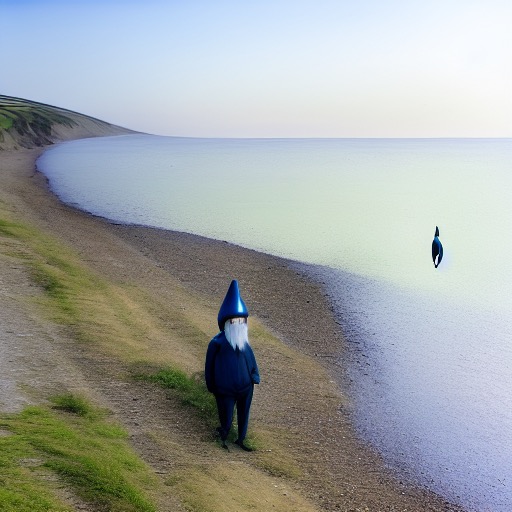}
    \end{minipage}%
    \begin{minipage}[t]{0.13\textwidth}
        \centering
        \subcaption[]{VISII w/ Text}
        \includegraphics[width=\textwidth]{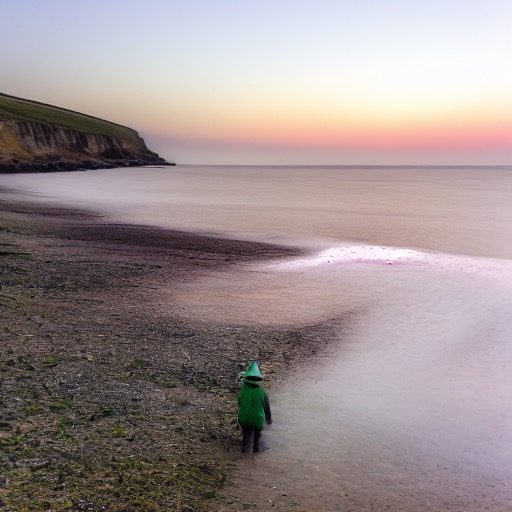}
    \end{minipage}%
    \begin{minipage}[t]{0.13\textwidth}
        \centering
        \subcaption[]{IP2P w/ Text}
        \includegraphics[width=\textwidth]{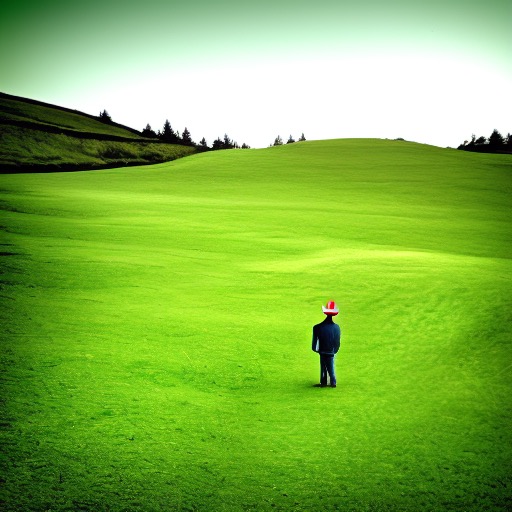}
    \end{minipage}%
    \par

    \begin{minipage}[t]{0.13\textwidth}
        \centering
        \includegraphics[width=\textwidth]{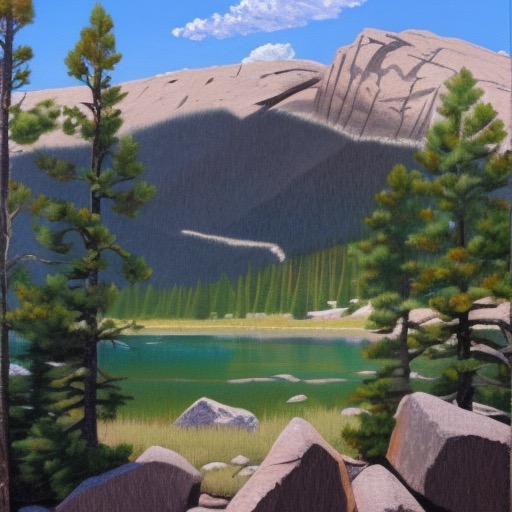}
    \end{minipage}%
    \begin{minipage}[t]{0.13\textwidth}
        \centering
        \includegraphics[width=\textwidth]{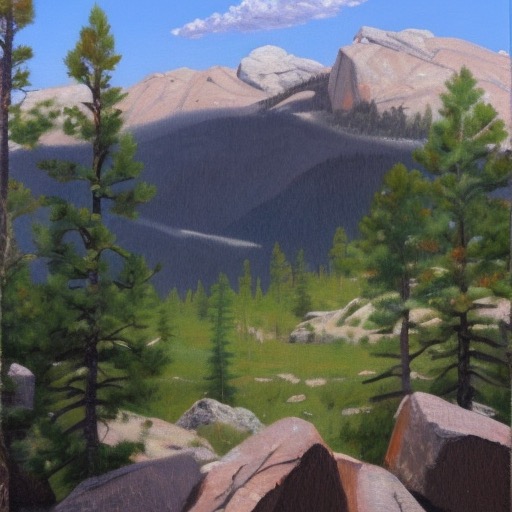}
    \end{minipage}%
    \begin{minipage}[t]{0.13\textwidth}
        \centering
        \includegraphics[width=\textwidth]{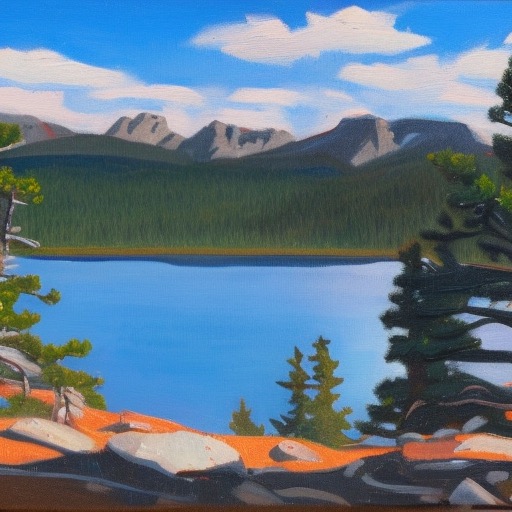}
    \end{minipage}%
    \begin{minipage}[t]{0.13\textwidth}
        \centering
        \includegraphics[width=\textwidth]{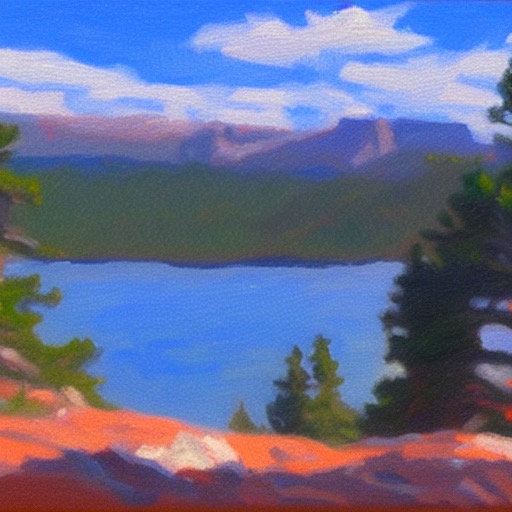}
    \end{minipage}%
    \begin{minipage}[t]{0.13\textwidth}
        \centering
        \includegraphics[width=\textwidth]{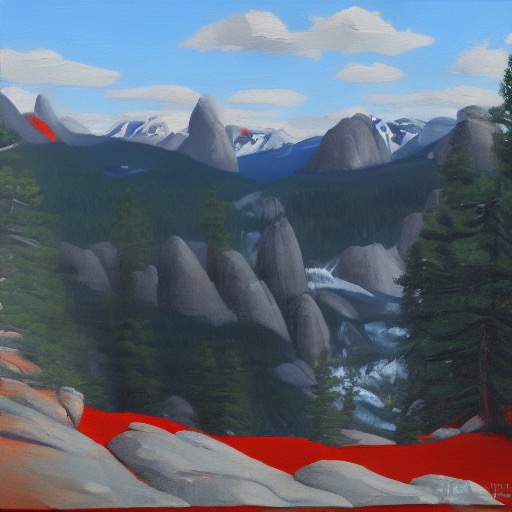}
    \end{minipage}%
    \begin{minipage}[t]{0.13\textwidth}
        \centering
        \includegraphics[width=\textwidth]{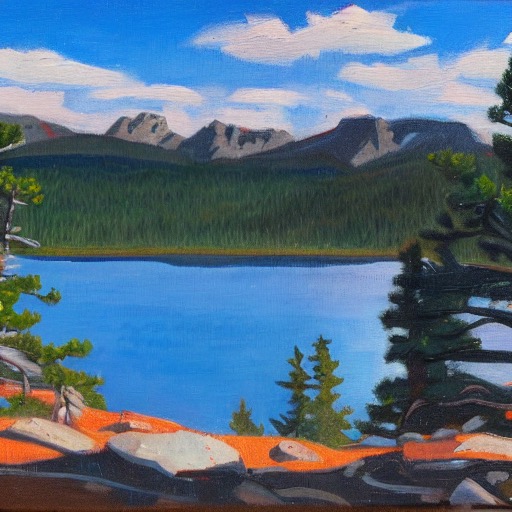}
    \end{minipage}%
    \begin{minipage}[t]{0.13\textwidth}
        \centering
        \includegraphics[width=\textwidth]{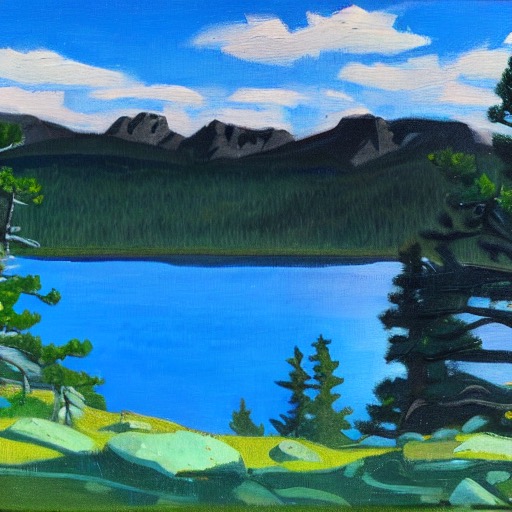}
    \end{minipage}%
    \par

    \begin{minipage}[t]{0.13\textwidth}
        \centering
        \includegraphics[width=\textwidth]{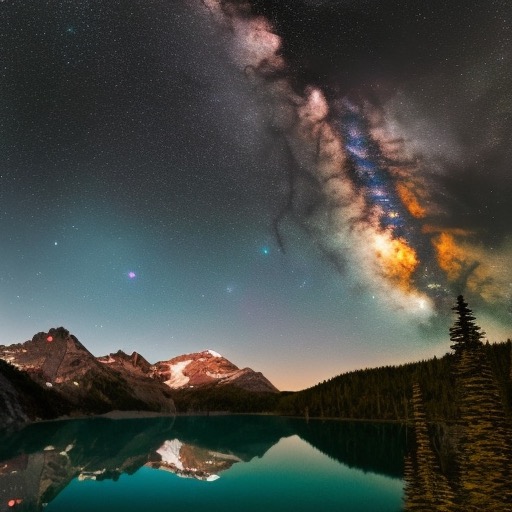}
    \end{minipage}%
    \begin{minipage}[t]{0.13\textwidth}
        \centering
        \includegraphics[width=\textwidth]{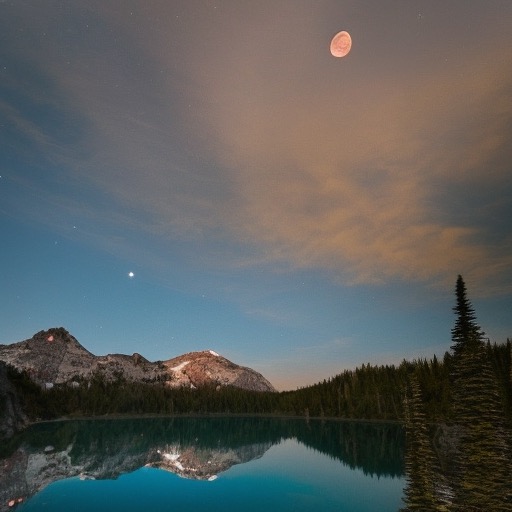}
    \end{minipage}%
    \begin{minipage}[t]{0.13\textwidth}
        \centering
        \includegraphics[width=\textwidth]{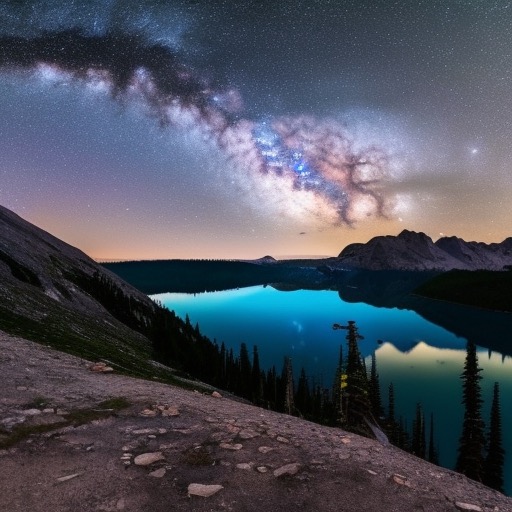}
    \end{minipage}%
    \begin{minipage}[t]{0.13\textwidth}
        \centering
        \includegraphics[width=\textwidth]{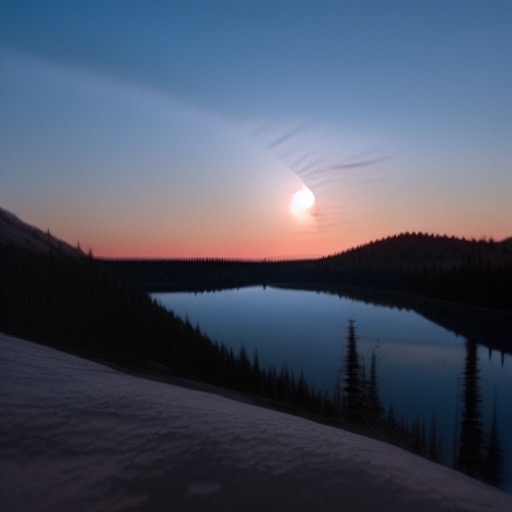}
    \end{minipage}%
    \begin{minipage}[t]{0.13\textwidth}
        \centering
        \includegraphics[width=\textwidth]{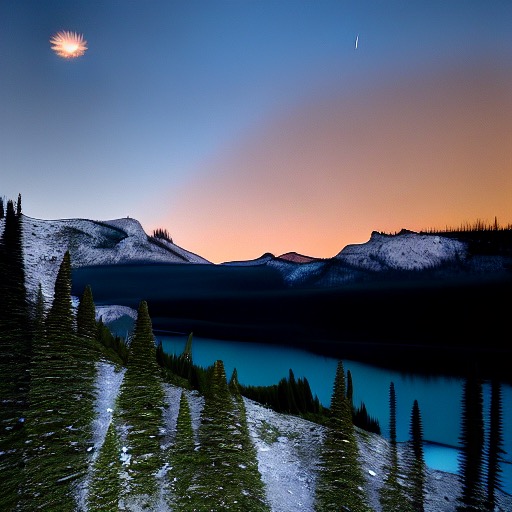}
    \end{minipage}%
    \begin{minipage}[t]{0.13\textwidth}
        \centering
        \includegraphics[width=\textwidth]{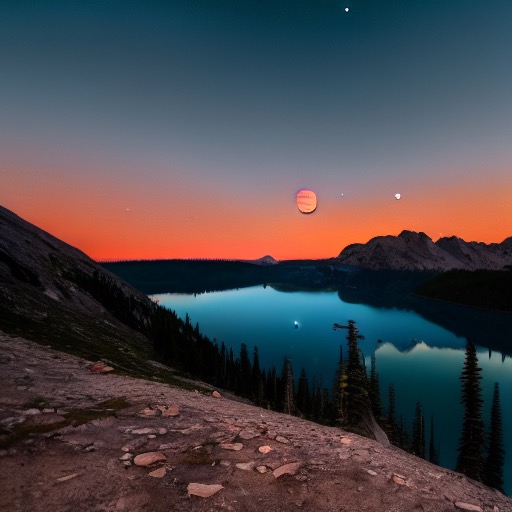}
    \end{minipage}%
    \begin{minipage}[t]{0.13\textwidth}
        \centering
        \includegraphics[width=\textwidth]{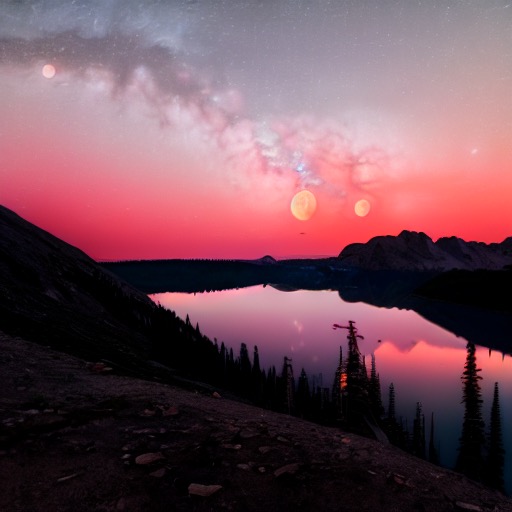}
    \end{minipage}%
    \par
    \caption{Illustration of failure cases of \method. \method struggles most in addition or removal of objects. However, baselines also produce undesirable results in these cases.}
    \label{fig:failure}
\end{figure*}

\subsection{Limitations of \method}
\label{sec:limit}

We present a novel approach for exemplar-based image editing that addresses several limitations of existing methods, such as over-reliance on models like InstructPix2Pix~\cite{brooks2023instructpix2pix} (VISII). Our method produces state-of-the-art results approximately four times faster than strong baselines. However, it has some limitations. We illustrate some of these limitations in Fig.~\ref{fig:failure}. For edits like \emph{object addition}, our method's performance can be poor, especially when the objects are extremely small. Additionally, as seen in Row 2 of the same figure, \method also fails to remove the large lake. However, all the remaining baselines also fail in these cases, producing high levels of distortions to produce the edit. We attribute of \method in these cases due to the over-reliance on the guidance ($f, Q, K$), which prevents large changes in structure. A key area of exploration is \textit{selective guidance} to circumvent this problem, which is part of our future work.

\begin{figure*}[]
    \centering
    \begin{minipage}[t]{0.13\textwidth}
        \centering
        \subcaption[]{$x$}
        \includegraphics[width=\textwidth]{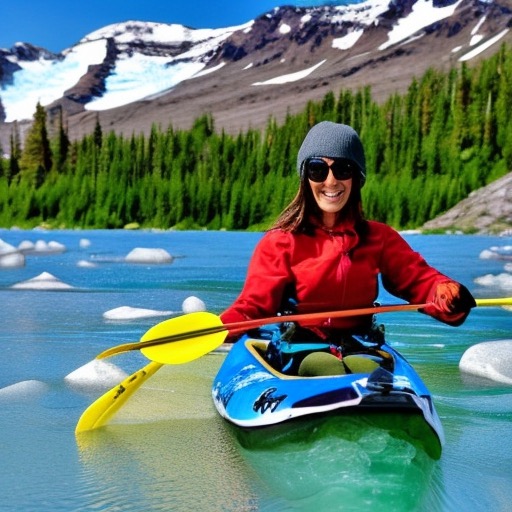}
    \end{minipage}%
    \begin{minipage}[t]{0.13\textwidth}
        \centering
        \subcaption[]{$x_{\text{edit}}$}
        \includegraphics[width=\textwidth]{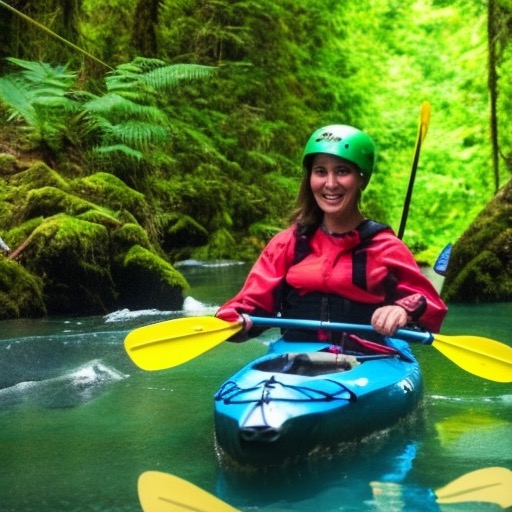}
    \end{minipage}%
    \begin{minipage}[t]{0.13\textwidth}
        \centering
        \subcaption[]{$y$}
        \includegraphics[width=\textwidth]{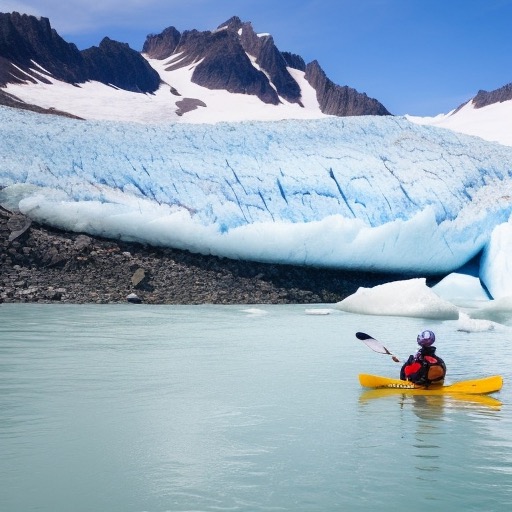}
    \end{minipage}%
    \begin{minipage}[t]{0.13\textwidth}
        \centering
        \subcaption[]{\method}
        \includegraphics[width=\textwidth]{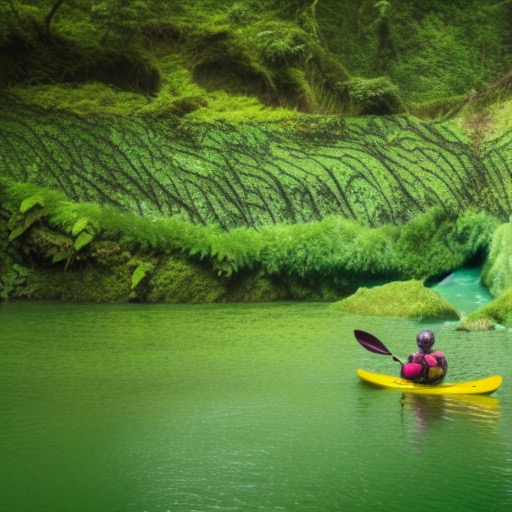}
    \end{minipage}%
    \begin{minipage}[t]{0.13\textwidth}
        \centering
        \subcaption[]{VISII}{}
        \includegraphics[width=\textwidth]{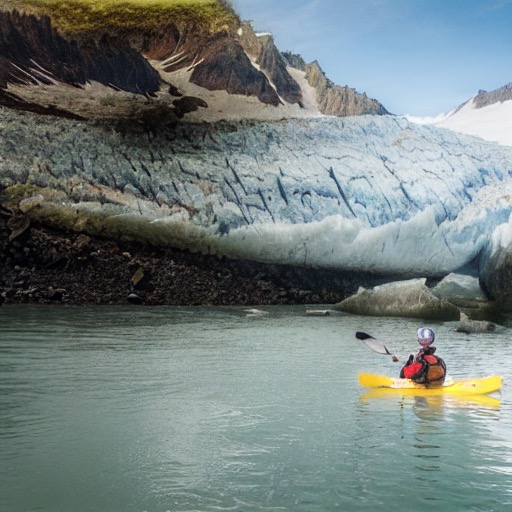}
    \end{minipage}%
    \begin{minipage}[t]{0.13\textwidth}
        \centering
        \subcaption[]{VISII Text}
        \includegraphics[width=\textwidth]{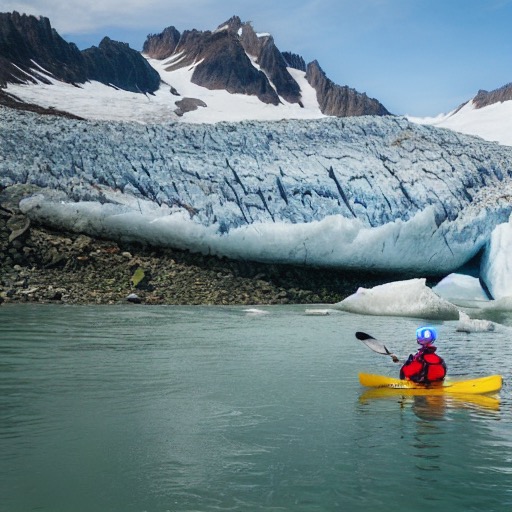}
    \end{minipage}%
    \begin{minipage}[t]{0.13\textwidth}
        \centering
        \subcaption[]{IP2P Text}
        \includegraphics[width=\textwidth]{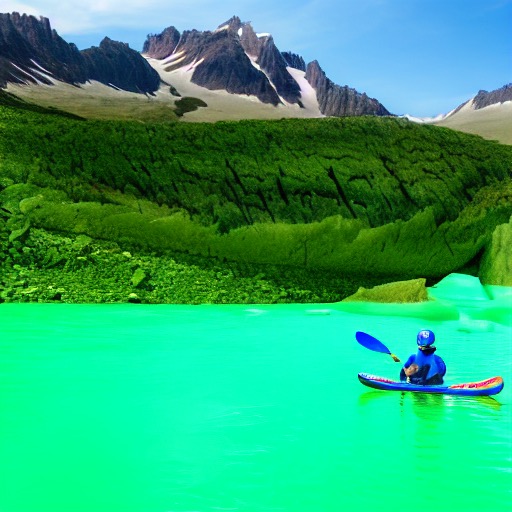}
    \end{minipage}%
    \par
    
    \begin{minipage}[t]{0.13\textwidth}
        \centering
        \includegraphics[width=\textwidth]{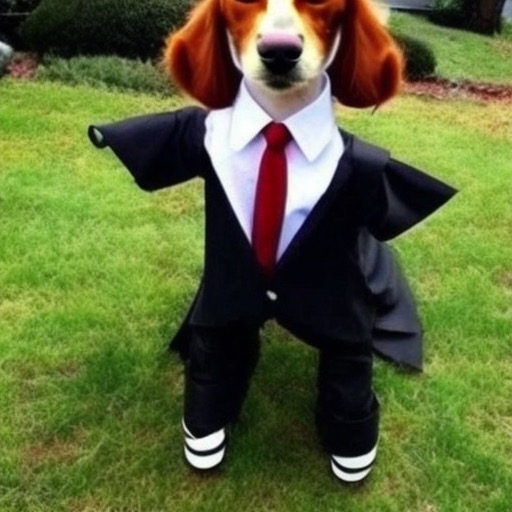}
    \end{minipage}%
    \begin{minipage}[t]{0.13\textwidth}
        \centering
        \includegraphics[width=\textwidth]{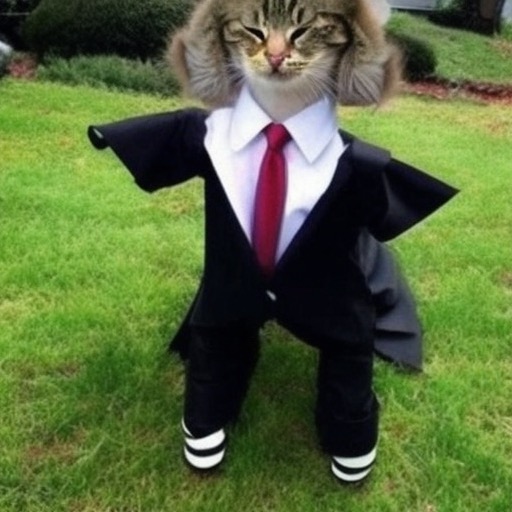}
    \end{minipage}%
    \begin{minipage}[t]{0.13\textwidth}
        \centering
        \includegraphics[width=\textwidth]{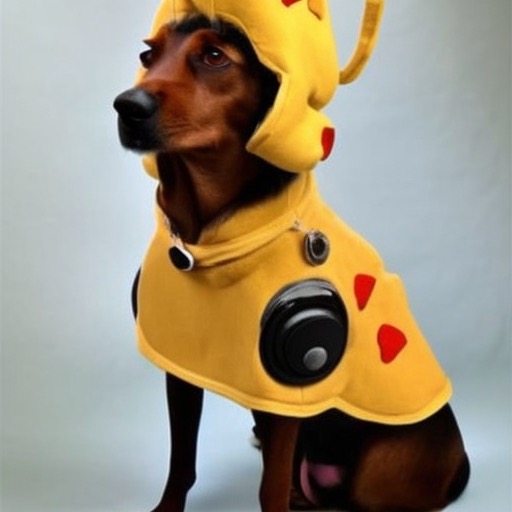}
    \end{minipage}%
    \begin{minipage}[t]{0.13\textwidth}
        \centering
        \includegraphics[width=\textwidth]{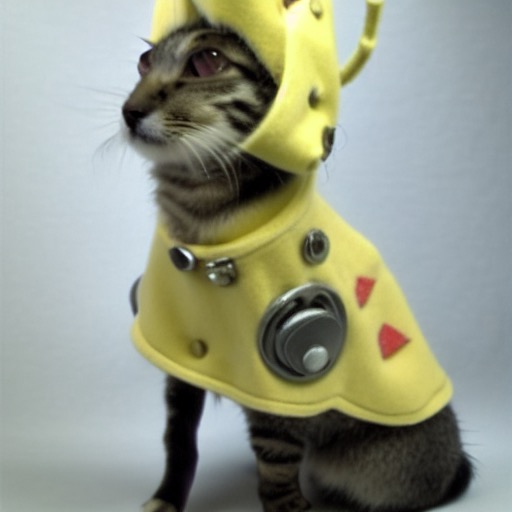}
    \end{minipage}%
    \begin{minipage}[t]{0.13\textwidth}
        \centering
        \includegraphics[width=\textwidth]{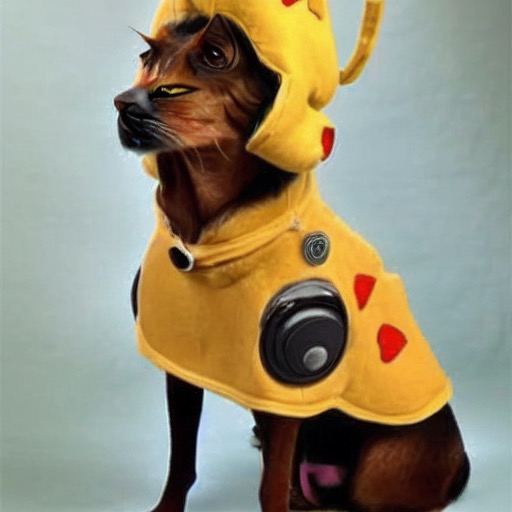}
    \end{minipage}%
    \begin{minipage}[t]{0.13\textwidth}
        \centering
        \includegraphics[width=\textwidth]{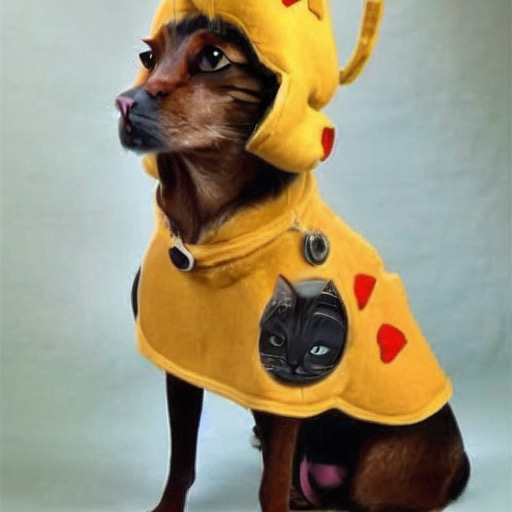}
    \end{minipage}%
    \begin{minipage}[t]{0.13\textwidth}
        \centering
        \includegraphics[width=\textwidth]{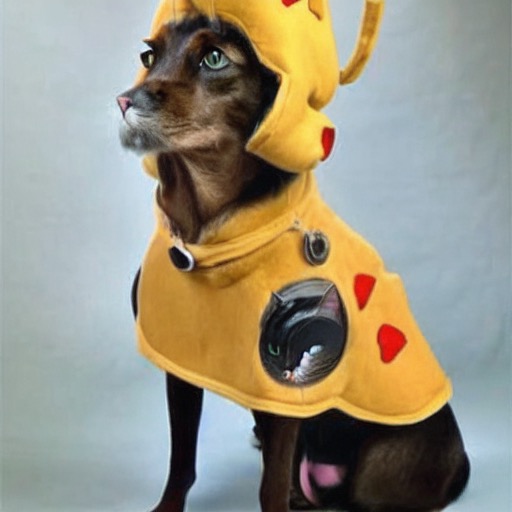}
    \end{minipage}%
    \par
    
    
    \begin{minipage}[t]{0.13\textwidth}
        \centering
        \includegraphics[width=\textwidth]{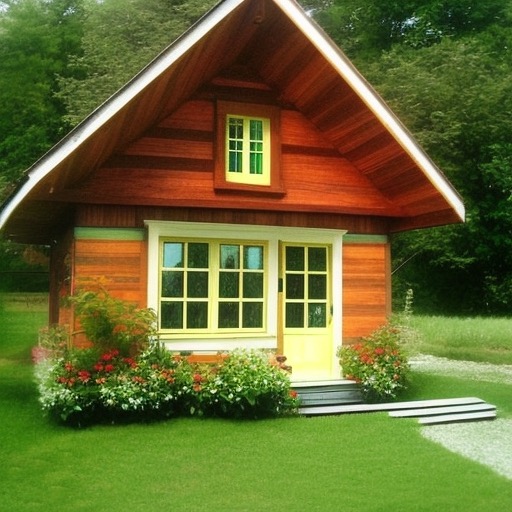}
    \end{minipage}%
    \begin{minipage}[t]{0.13\textwidth}
        \centering
        \includegraphics[width=\textwidth]{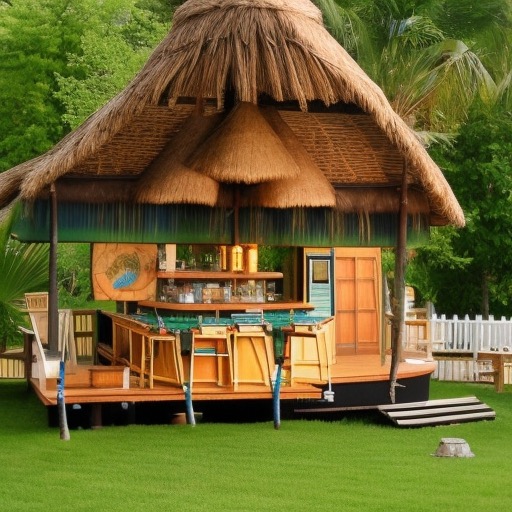}
    \end{minipage}%
    \begin{minipage}[t]{0.13\textwidth}
        \centering
        \includegraphics[width=\textwidth]{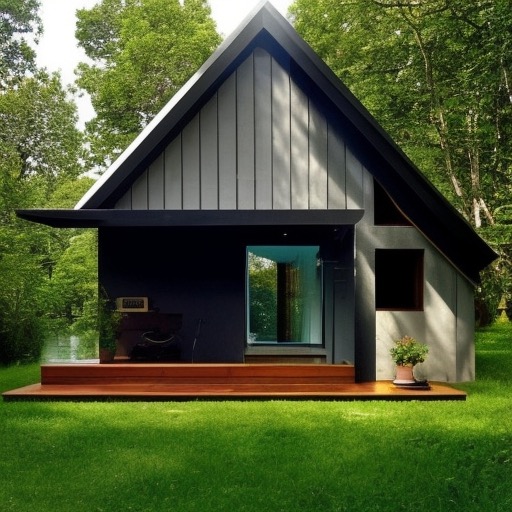}
    \end{minipage}%
    \begin{minipage}[t]{0.13\textwidth}
        \centering
        \includegraphics[width=\textwidth]{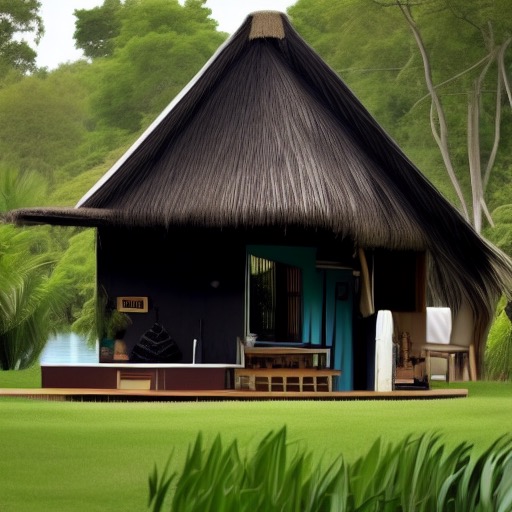}
    \end{minipage}%
    \begin{minipage}[t]{0.13\textwidth}
        \centering
        \includegraphics[width=\textwidth]{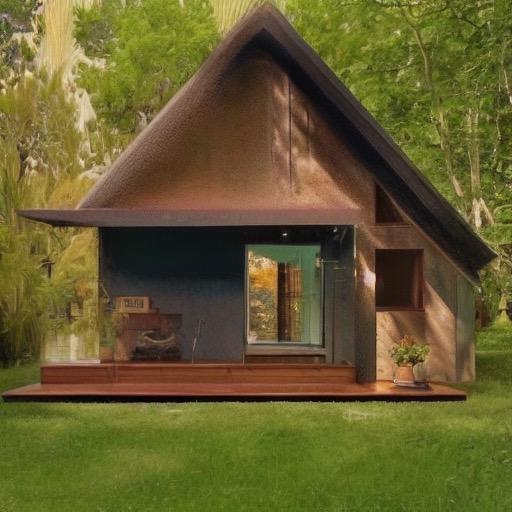}
    \end{minipage}%
    \begin{minipage}[t]{0.13\textwidth}
        \centering
        \includegraphics[width=\textwidth]{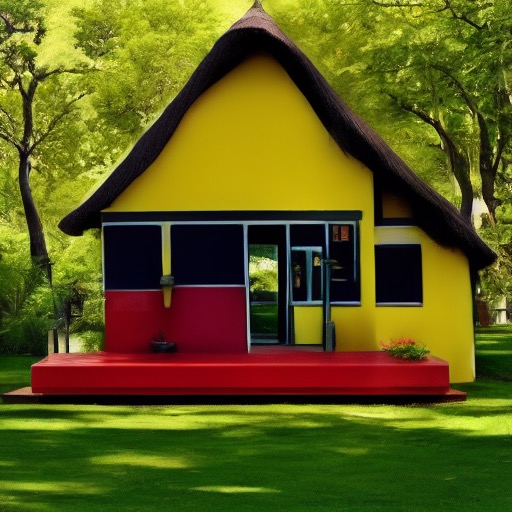}
    \end{minipage}%
    \begin{minipage}[t]{0.13\textwidth}
        \centering
        \includegraphics[width=\textwidth]{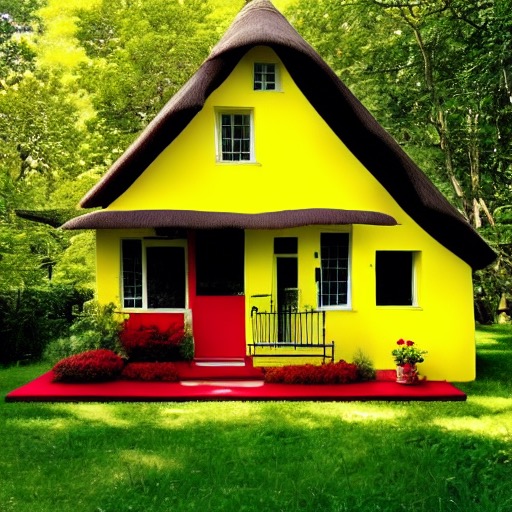}
    \end{minipage}%
    \par

    \begin{minipage}[t]{0.13\textwidth}
        \centering
        \includegraphics[width=\textwidth]{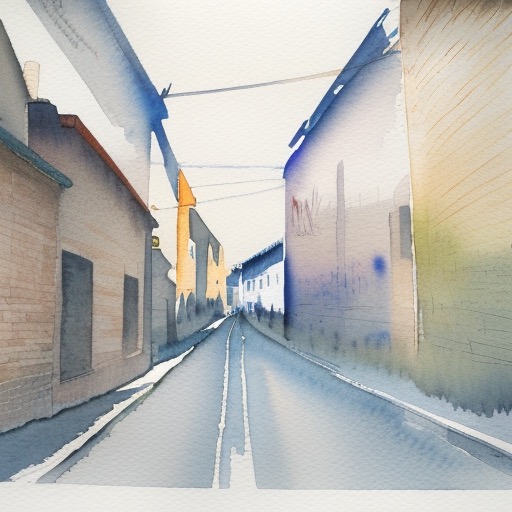}
    \end{minipage}%
    \begin{minipage}[t]{0.13\textwidth}
        \centering
        \includegraphics[width=\textwidth]{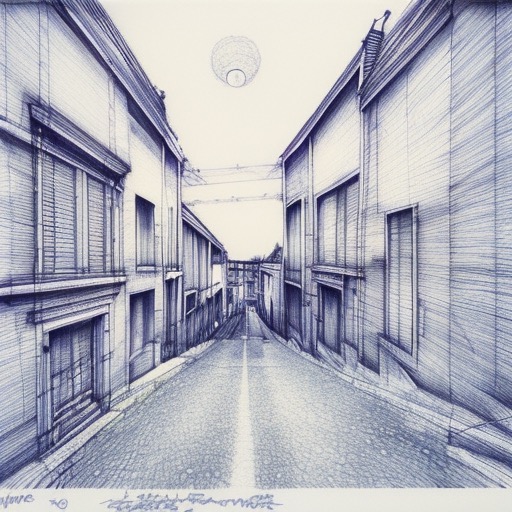}
    \end{minipage}%
    \begin{minipage}[t]{0.13\textwidth}
        \centering
        \includegraphics[width=\textwidth]{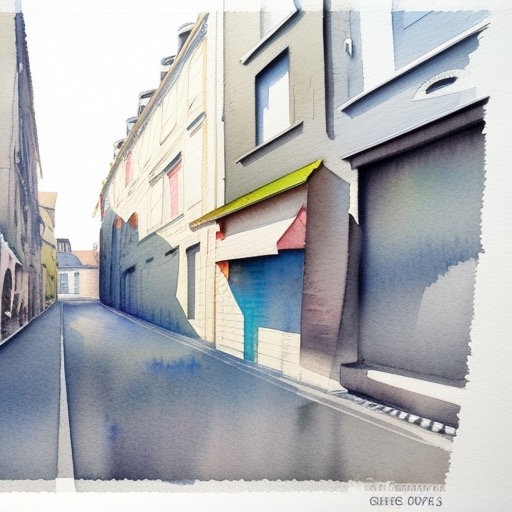}
    \end{minipage}%
    \begin{minipage}[t]{0.13\textwidth}
        \centering
        \includegraphics[width=\textwidth]{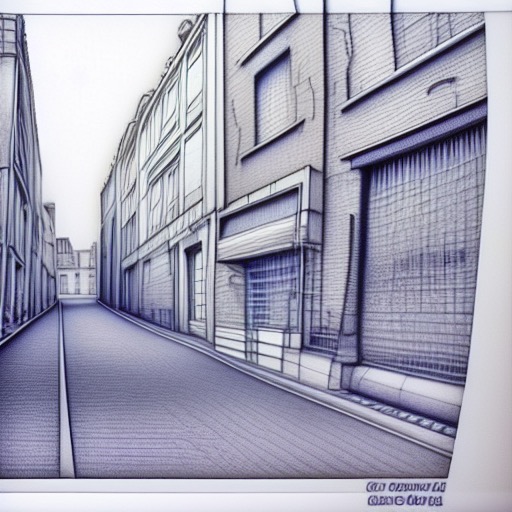}
    \end{minipage}%
    \begin{minipage}[t]{0.13\textwidth}
        \centering
        \includegraphics[width=\textwidth]{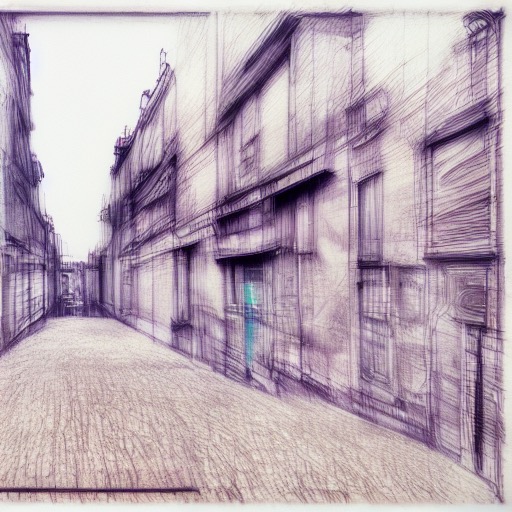}
    \end{minipage}%
    \begin{minipage}[t]{0.13\textwidth}
        \centering
        \includegraphics[width=\textwidth]{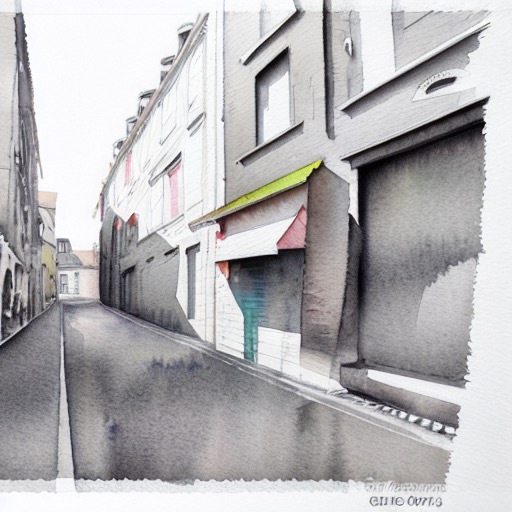}
    \end{minipage}%
    \begin{minipage}[t]{0.13\textwidth}
        \centering
        \includegraphics[width=\textwidth]{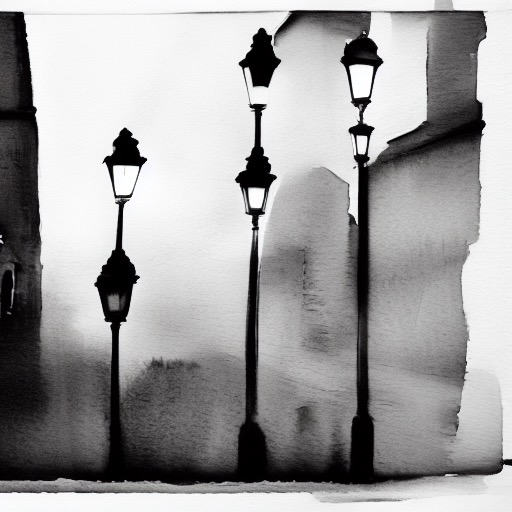}
    \end{minipage}%
    \par

    \begin{minipage}[t]{0.13\textwidth}
        \centering
        \includegraphics[width=\textwidth]{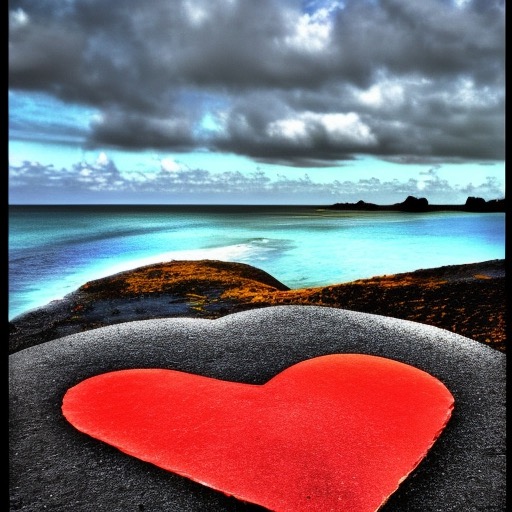}
    \end{minipage}%
    \begin{minipage}[t]{0.13\textwidth}
        \centering
        \includegraphics[width=\textwidth]{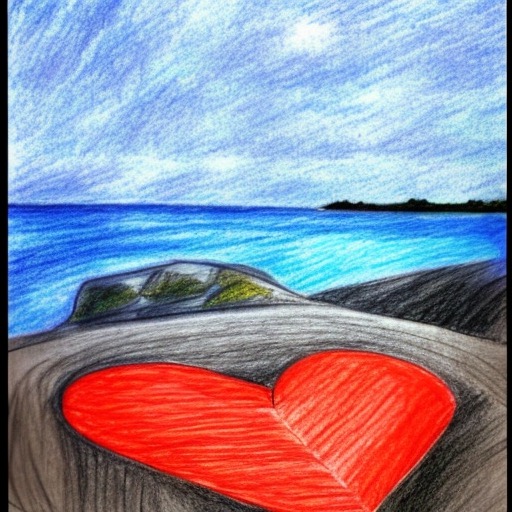}
    \end{minipage}%
    \begin{minipage}[t]{0.13\textwidth}
        \centering
        \includegraphics[width=\textwidth]{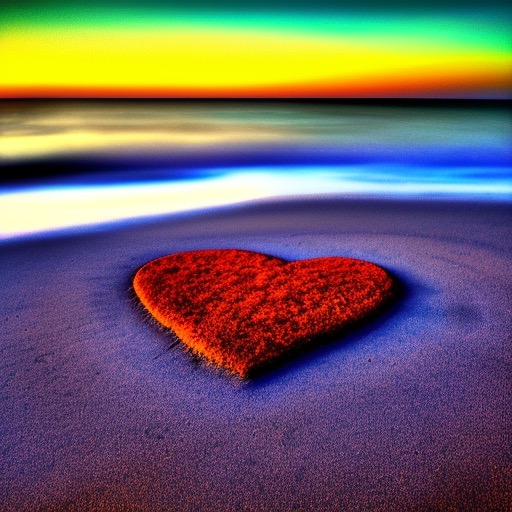}
    \end{minipage}%
    \begin{minipage}[t]{0.13\textwidth}
        \centering
        \includegraphics[width=\textwidth]{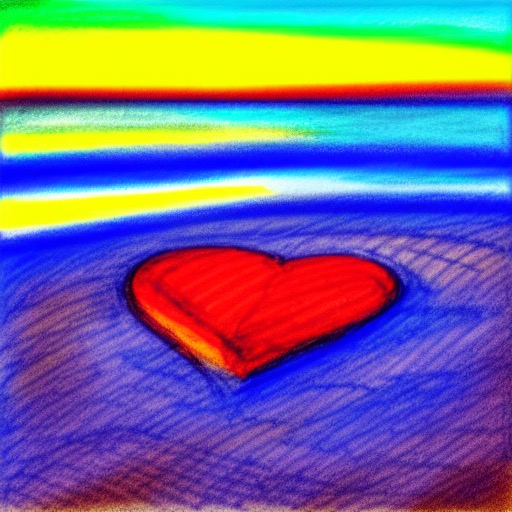}
    \end{minipage}%
    \begin{minipage}[t]{0.13\textwidth}
        \centering
        \includegraphics[width=\textwidth]{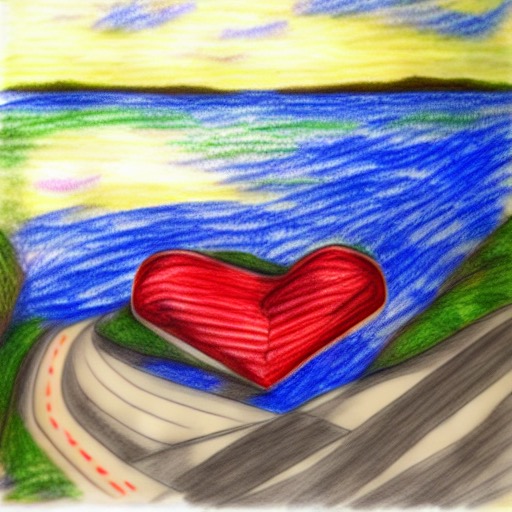}
    \end{minipage}%
    \begin{minipage}[t]{0.13\textwidth}
        \centering
        \includegraphics[width=\textwidth]{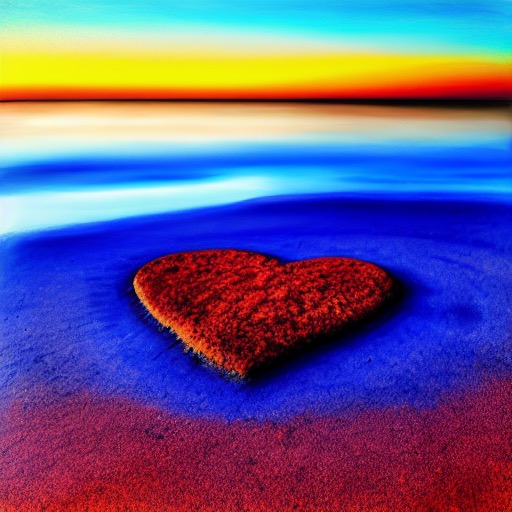}
    \end{minipage}%
    \begin{minipage}[t]{0.13\textwidth}
        \centering
        \includegraphics[width=\textwidth]{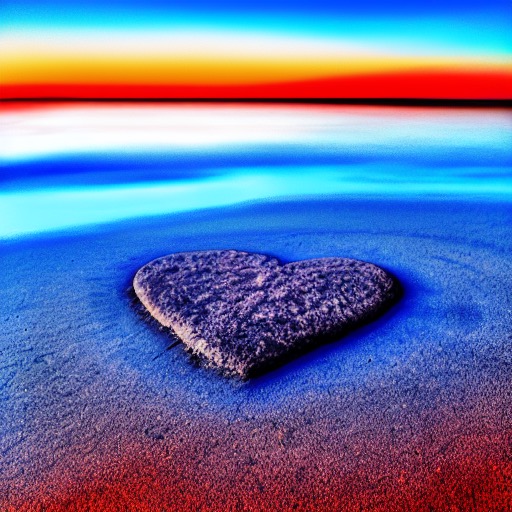}
    \end{minipage}%
    \par

    \begin{minipage}[t]{0.13\textwidth}
        \centering
        \includegraphics[width=\textwidth]{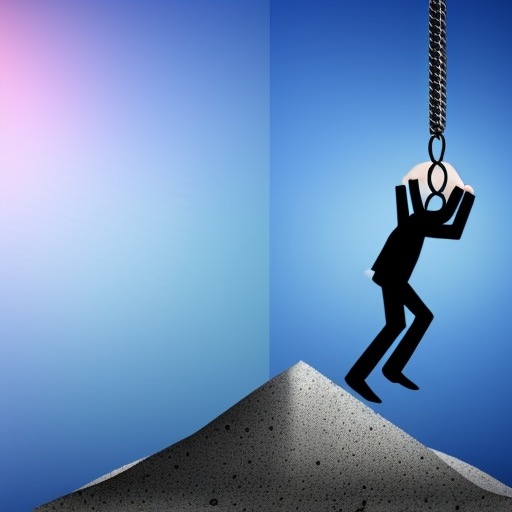}
    \end{minipage}%
    \begin{minipage}[t]{0.13\textwidth}
        \centering
        \includegraphics[width=\textwidth]{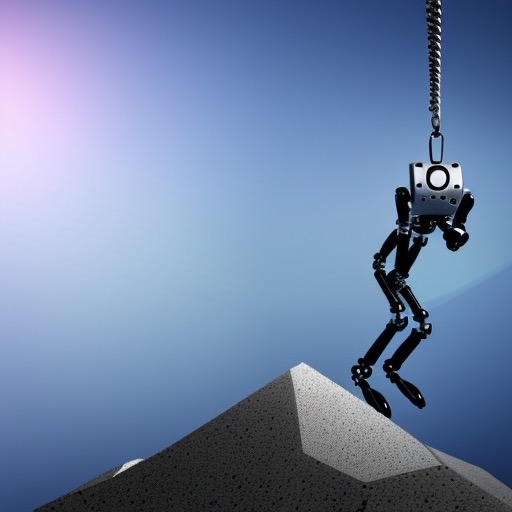}
    \end{minipage}%
    \begin{minipage}[t]{0.13\textwidth}
        \centering
        \includegraphics[width=\textwidth]{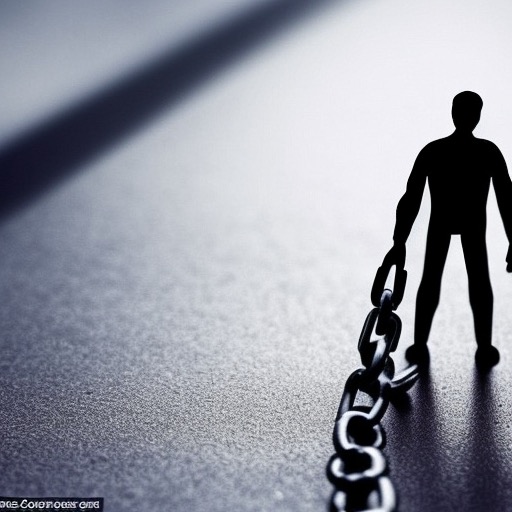}
    \end{minipage}%
    \begin{minipage}[t]{0.13\textwidth}
        \centering
        \includegraphics[width=\textwidth]{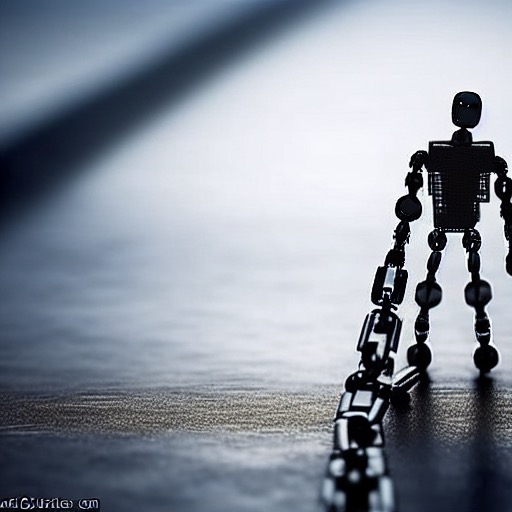}
    \end{minipage}%
    \begin{minipage}[t]{0.13\textwidth}
        \centering
        \includegraphics[width=\textwidth]{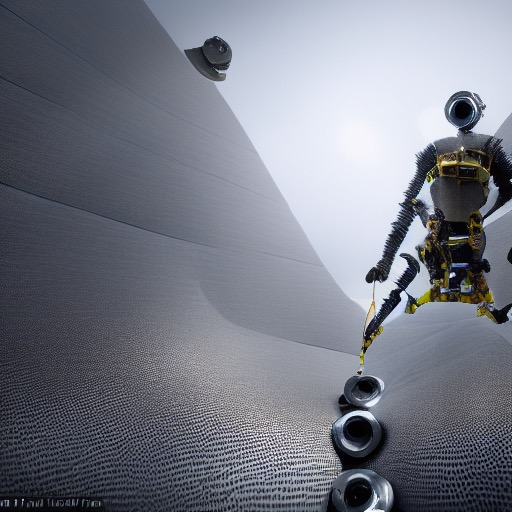}
    \end{minipage}%
    \begin{minipage}[t]{0.13\textwidth}
        \centering
        \includegraphics[width=\textwidth]{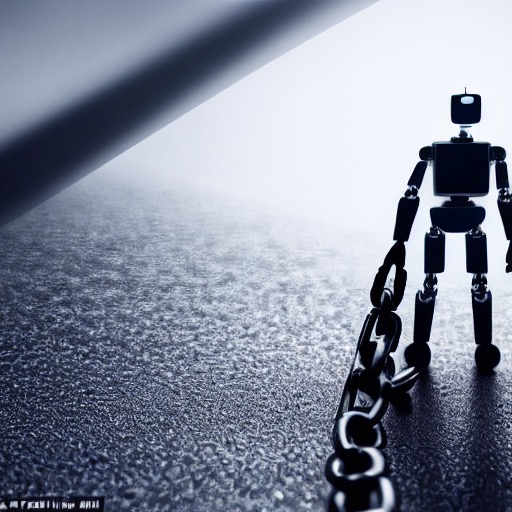}
    \end{minipage}%
    \begin{minipage}[t]{0.13\textwidth}
        \centering
        \includegraphics[width=\textwidth]{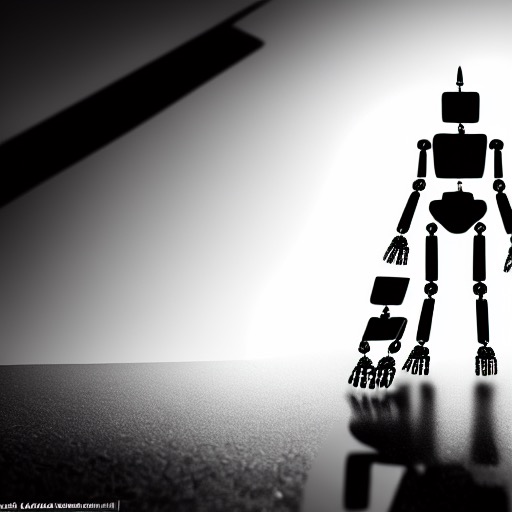}
    \end{minipage}%
    \par

    \begin{minipage}[t]{0.13\textwidth}
        \centering
        \includegraphics[width=\textwidth]{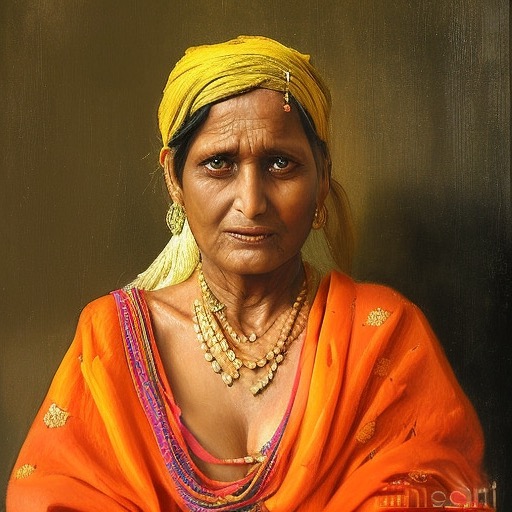}
    \end{minipage}%
    \begin{minipage}[t]{0.13\textwidth}
        \centering
        \includegraphics[width=\textwidth]{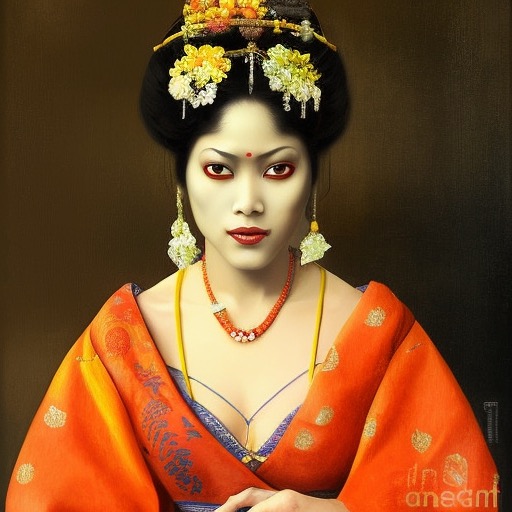}
    \end{minipage}%
    \begin{minipage}[t]{0.13\textwidth}
        \centering
        \includegraphics[width=\textwidth]{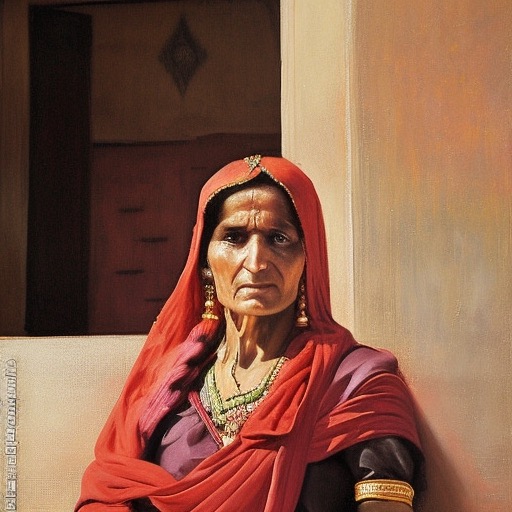}
    \end{minipage}%
    \begin{minipage}[t]{0.13\textwidth}
        \centering
        \includegraphics[width=\textwidth]{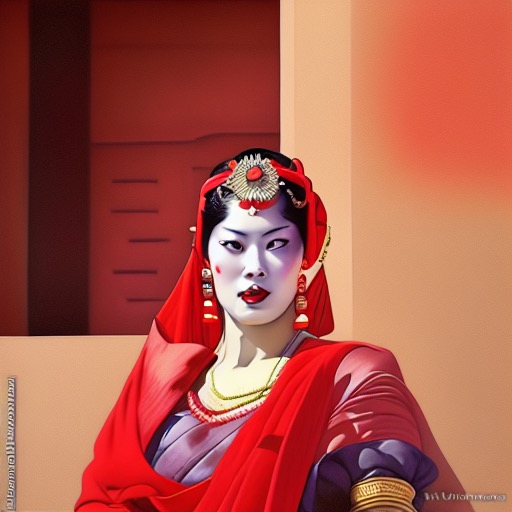}
    \end{minipage}%
    \begin{minipage}[t]{0.13\textwidth}
        \centering
        \includegraphics[width=\textwidth]{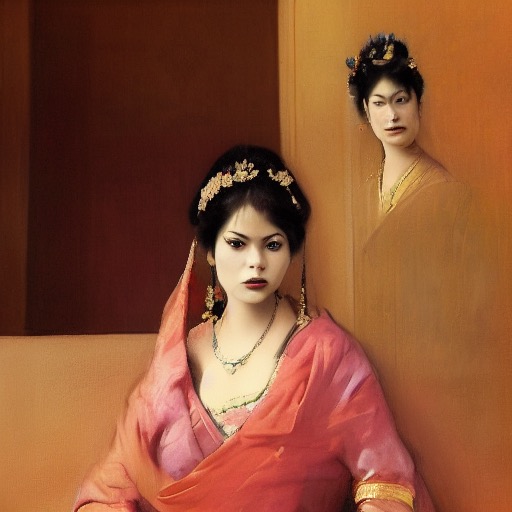}
    \end{minipage}%
    \begin{minipage}[t]{0.13\textwidth}
        \centering
        \includegraphics[width=\textwidth]{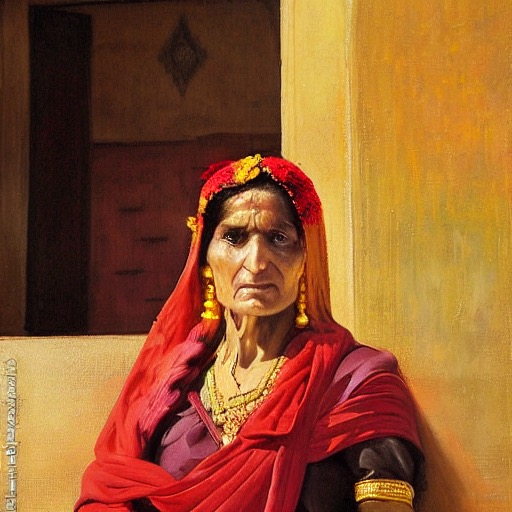}
    \end{minipage}%
    \begin{minipage}[t]{0.13\textwidth}
        \centering
        \includegraphics[width=\textwidth]{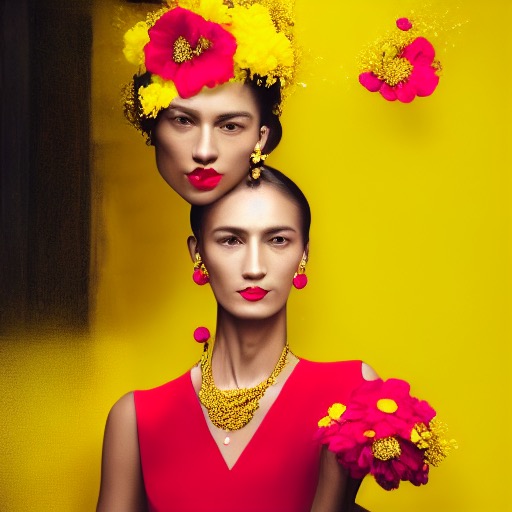}
    \end{minipage}%
    \par

    \caption{Overview of additional qualitative comparisons: We show additional results across different edit types. \method clearly outperforms the baselines consistently, by both maintaining the structure of the test image $y$ and being faithful to the edit illustrated in the exemplar pair.). View at high magnification to observe subtle edits.
    \label{fig:addn-qual1}}
\end{figure*}

\begin{figure*}
\ContinuedFloat
        \begin{minipage}[t]{0.13\textwidth}
        \centering
        \subcaption[]{$x$}
        \includegraphics[width=\textwidth]{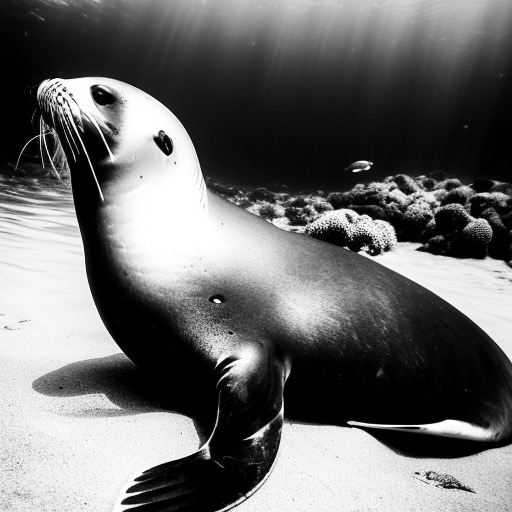}
    \end{minipage}%
    \begin{minipage}[t]{0.13\textwidth}
        \centering
        \subcaption[]{$x_\text{edit}$}
        \includegraphics[width=\textwidth]{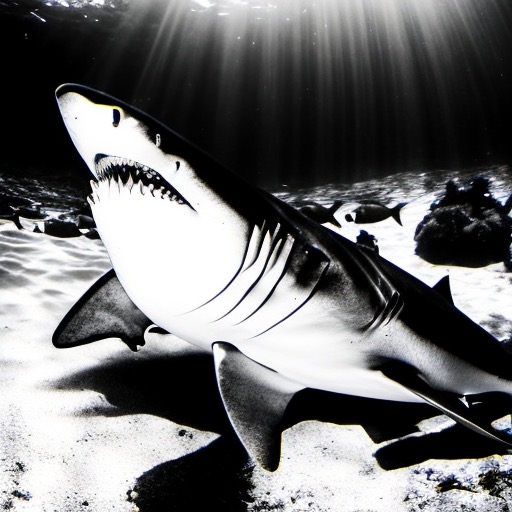}
    \end{minipage}%
    \begin{minipage}[t]{0.13\textwidth}
        \centering
        \subcaption[]{$y$}
        \includegraphics[width=\textwidth]{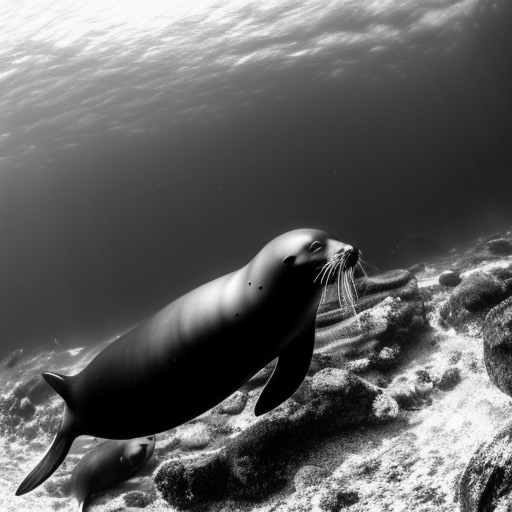}
    \end{minipage}%
    \begin{minipage}[t]{0.13\textwidth}
        \centering
        \subcaption[]{\method}
        \includegraphics[width=\textwidth]{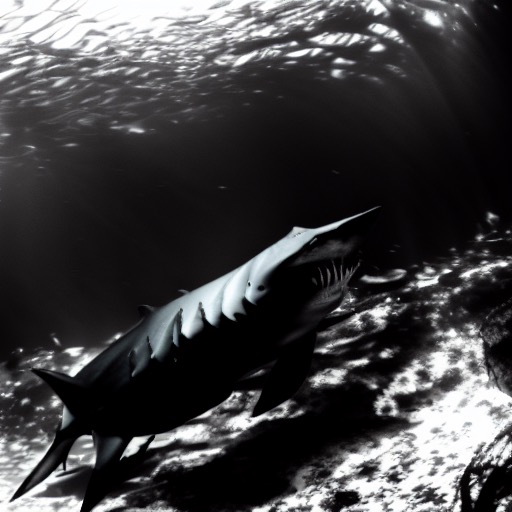}
    \end{minipage}%
    \begin{minipage}[t]{0.13\textwidth}
        \centering
        \subcaption[]{VISII}
        \includegraphics[width=\textwidth]{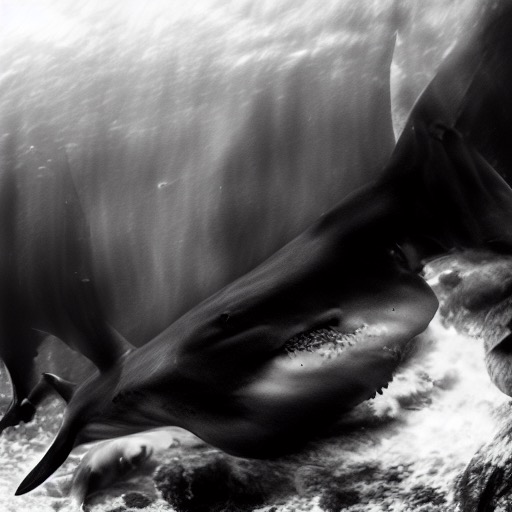}
    \end{minipage}%
    \begin{minipage}[t]{0.13\textwidth}
        \centering
        \subcaption[]{VISII w\ Text}
        \includegraphics[width=\textwidth]{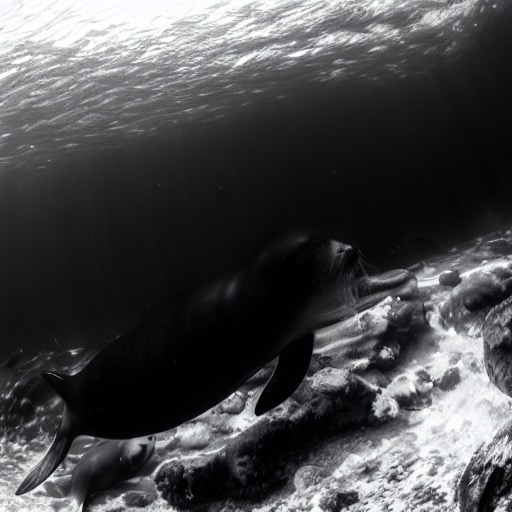}
    \end{minipage}%
    \begin{minipage}[t]{0.13\textwidth}
        \centering
        \subcaption[]{IP2P w\ Text}
        \includegraphics[width=\textwidth]{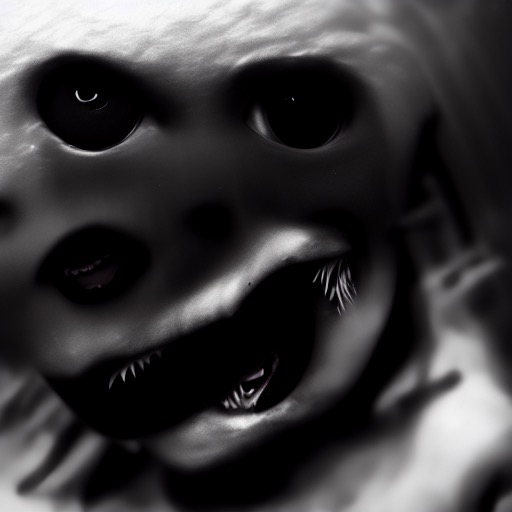}
    \end{minipage}%
    \par

    \begin{minipage}[t]{0.13\textwidth}
        \centering
        \includegraphics[width=\textwidth]{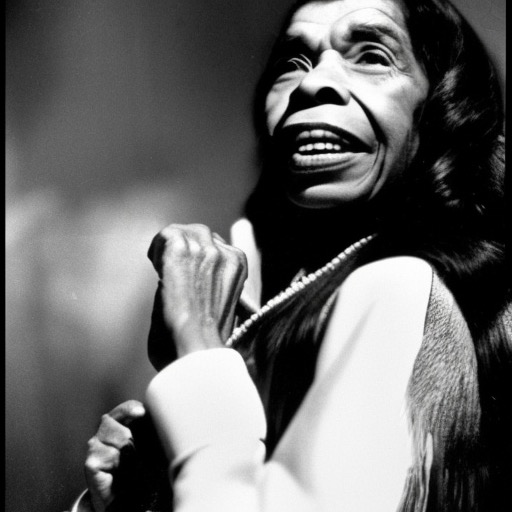}
    \end{minipage}%
    \begin{minipage}[t]{0.13\textwidth}
        \centering
        \includegraphics[width=\textwidth]{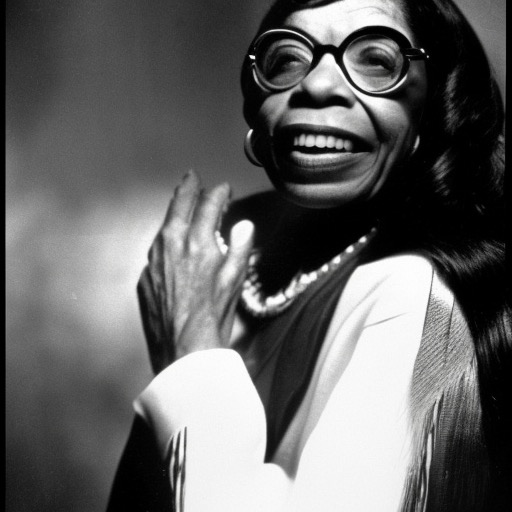}
    \end{minipage}%
    \begin{minipage}[t]{0.13\textwidth}
        \centering
        \includegraphics[width=\textwidth]{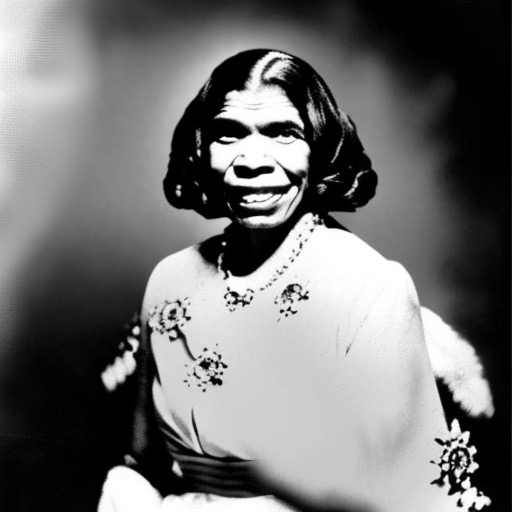}
    \end{minipage}%
    \begin{minipage}[t]{0.13\textwidth}
        \centering
        \includegraphics[width=\textwidth]{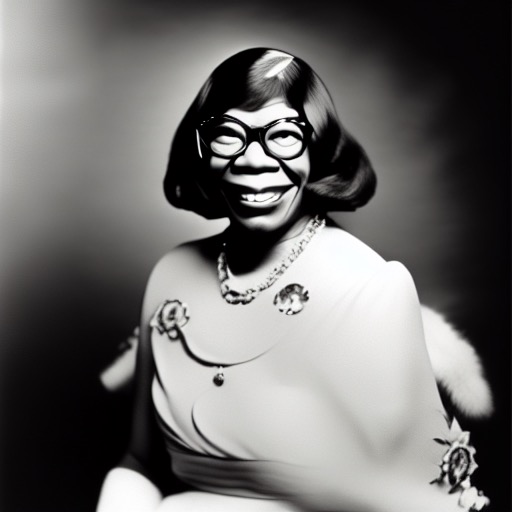}
    \end{minipage}%
    \begin{minipage}[t]{0.13\textwidth}
        \centering
        \includegraphics[width=\textwidth]{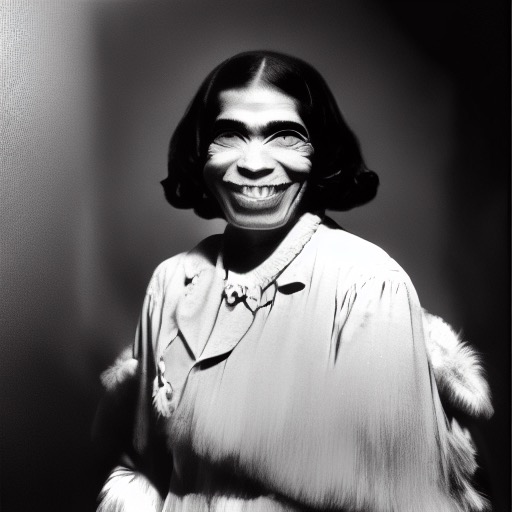}
    \end{minipage}%
    \begin{minipage}[t]{0.13\textwidth}
        \centering
        \includegraphics[width=\textwidth]{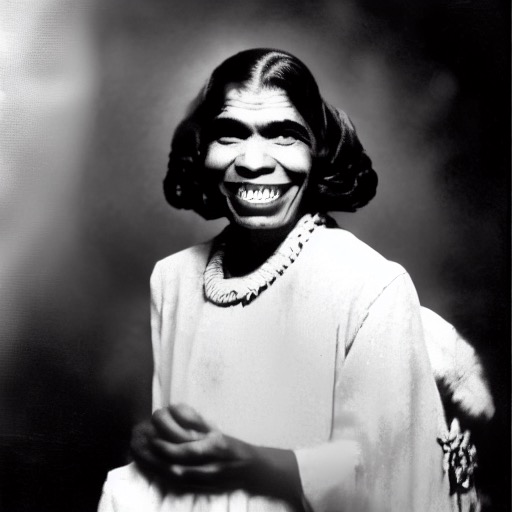}
    \end{minipage}%
    \begin{minipage}[t]{0.13\textwidth}
        \centering
        \includegraphics[width=\textwidth]{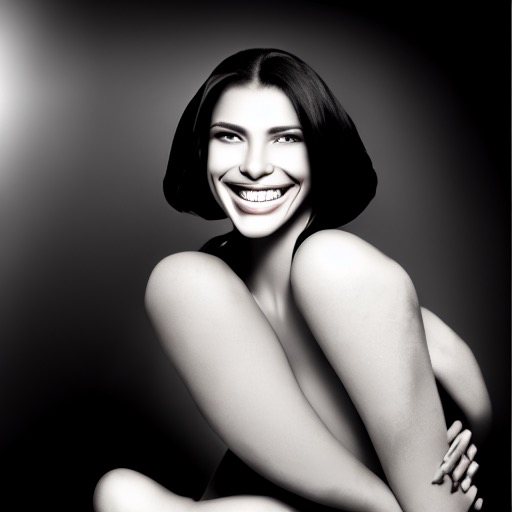}
    \end{minipage}%
    \par

    \begin{minipage}[t]{0.13\textwidth}
        \centering
        \includegraphics[width=\textwidth]{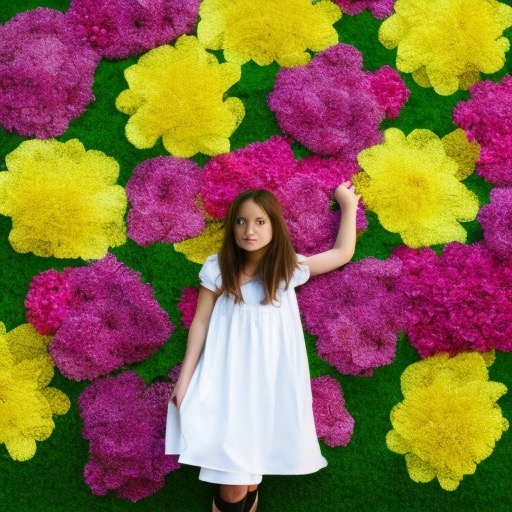}
    \end{minipage}%
    \begin{minipage}[t]{0.13\textwidth}
        \centering
        \includegraphics[width=\textwidth]{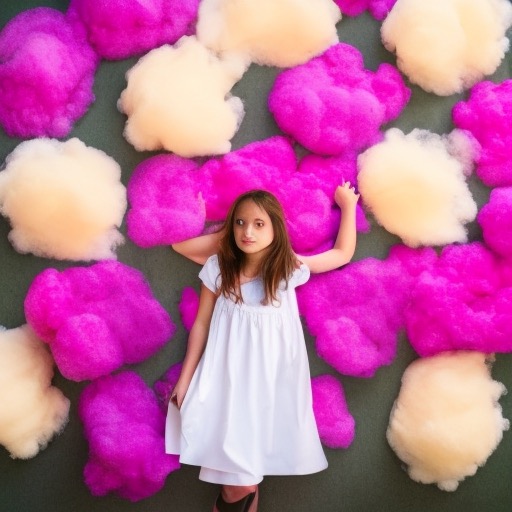}
    \end{minipage}%
    \begin{minipage}[t]{0.13\textwidth}
        \centering
        \includegraphics[width=\textwidth]{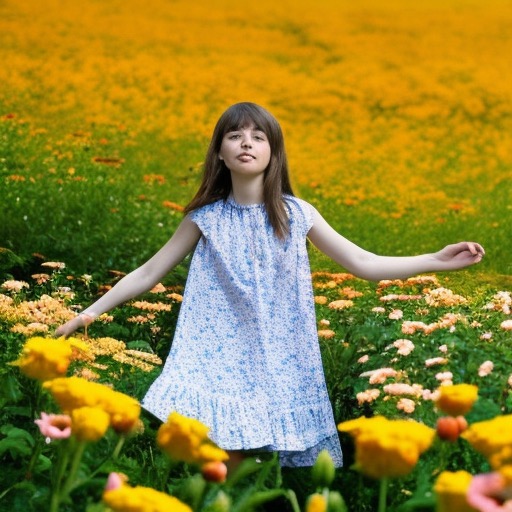}
    \end{minipage}%
    \begin{minipage}[t]{0.13\textwidth}
        \centering
        \includegraphics[width=\textwidth]{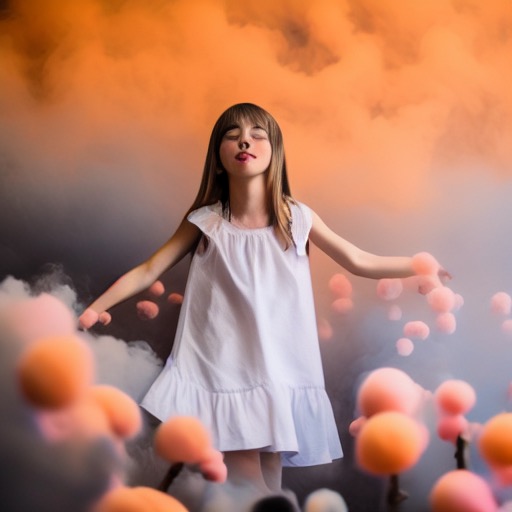}
    \end{minipage}%
    \begin{minipage}[t]{0.13\textwidth}
        \centering
        \includegraphics[width=\textwidth]{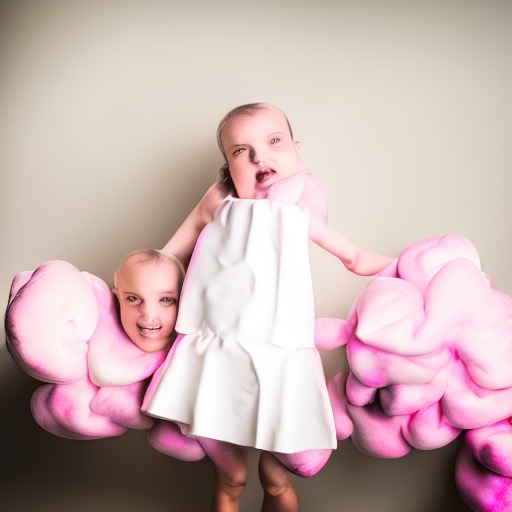}
    \end{minipage}%
    \begin{minipage}[t]{0.13\textwidth}
        \centering
        \includegraphics[width=\textwidth]{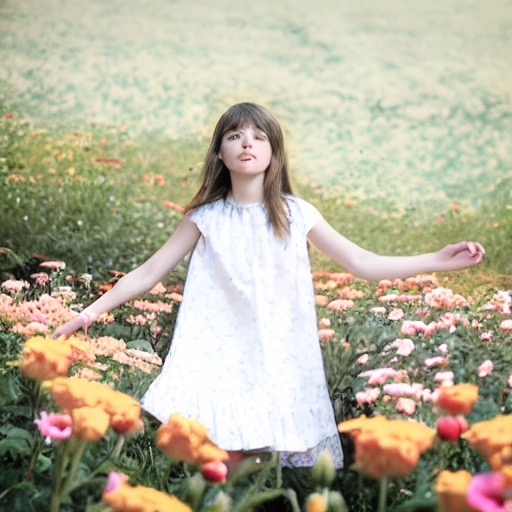}
    \end{minipage}%
    \begin{minipage}[t]{0.13\textwidth}
        \centering
        \includegraphics[width=\textwidth]{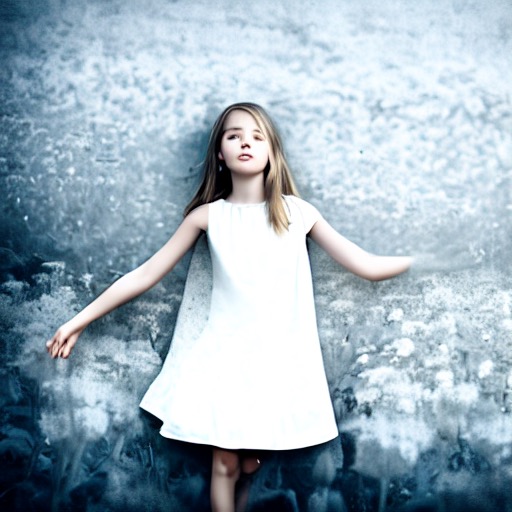}
    \end{minipage}%
    \par

    \begin{minipage}[t]{0.13\textwidth}
        \centering
        \includegraphics[width=\textwidth]{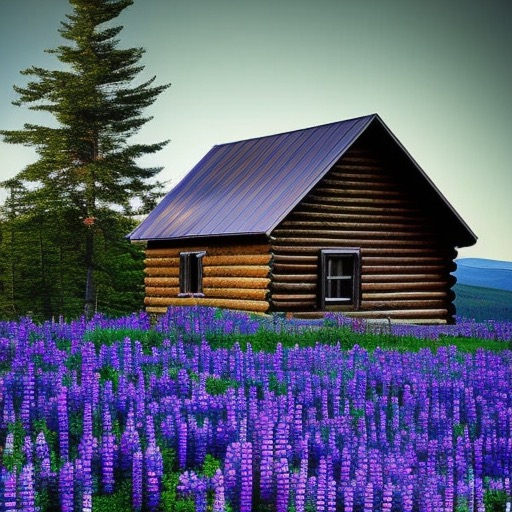}
    \end{minipage}%
    \begin{minipage}[t]{0.13\textwidth}
        \centering
        \includegraphics[width=\textwidth]{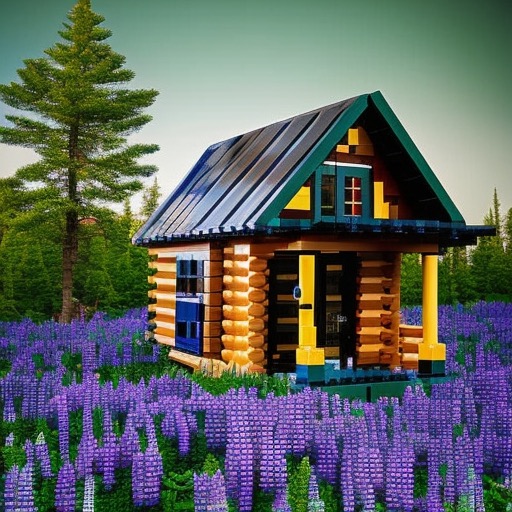}
    \end{minipage}%
    \begin{minipage}[t]{0.13\textwidth}
        \centering
        \includegraphics[width=\textwidth]{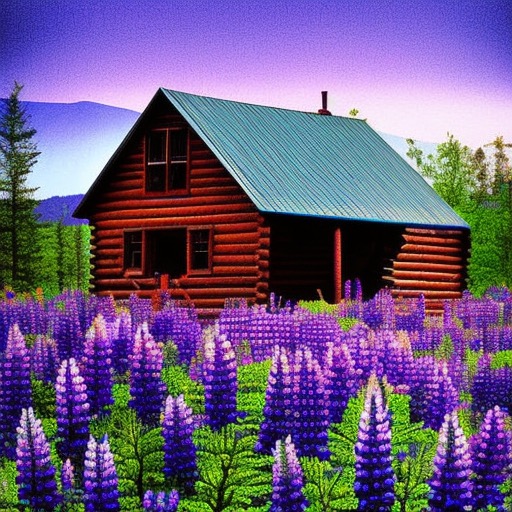}
    \end{minipage}%
    \begin{minipage}[t]{0.13\textwidth}
        \centering
        \includegraphics[width=\textwidth]{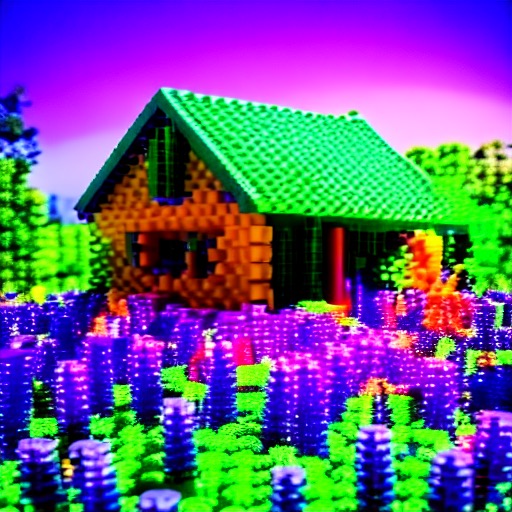}
    \end{minipage}%
    \begin{minipage}[t]{0.13\textwidth}
        \centering
        \includegraphics[width=\textwidth]{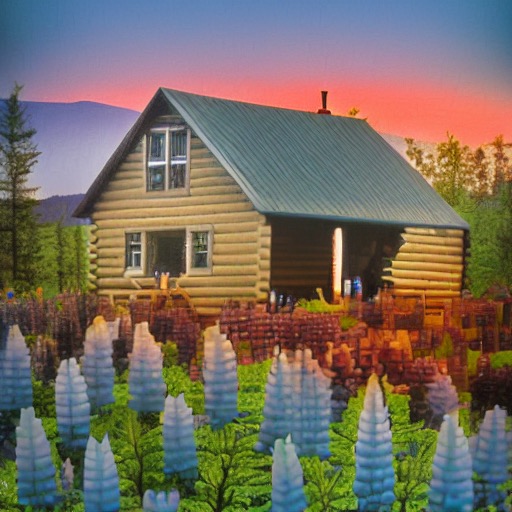}
    \end{minipage}%
    \begin{minipage}[t]{0.13\textwidth}
        \centering
        \includegraphics[width=\textwidth]{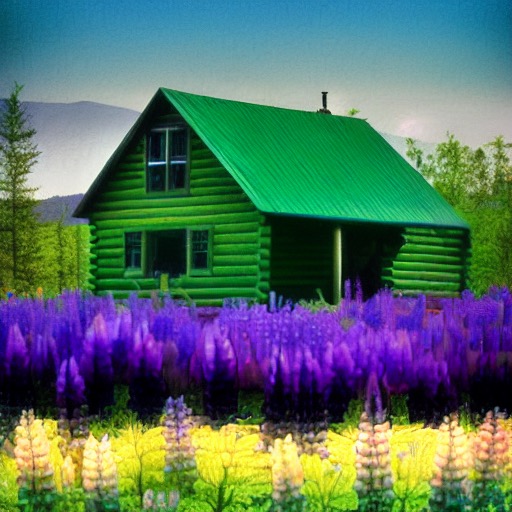}
    \end{minipage}%
    \begin{minipage}[t]{0.13\textwidth}
        \centering
        \includegraphics[width=\textwidth]{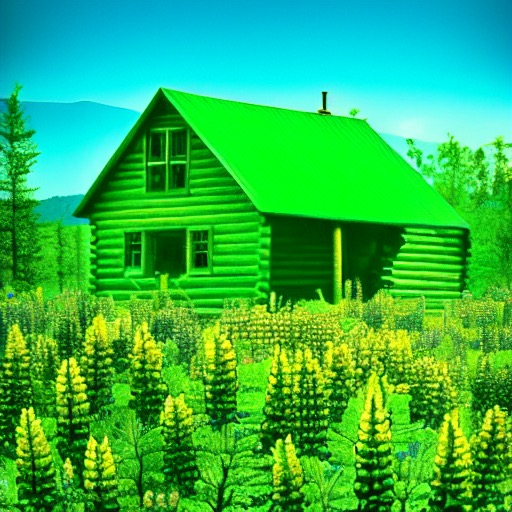}
    \end{minipage}%
    \par

    \begin{minipage}[t]{0.13\textwidth}
        \centering
        \includegraphics[width=\textwidth]{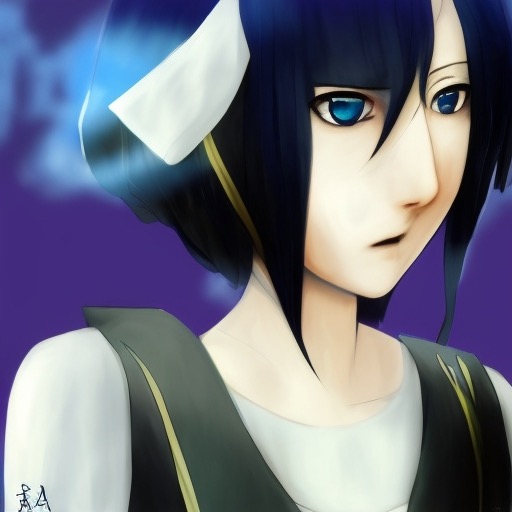}
    \end{minipage}%
    \begin{minipage}[t]{0.13\textwidth}
        \centering
        \includegraphics[width=\textwidth]{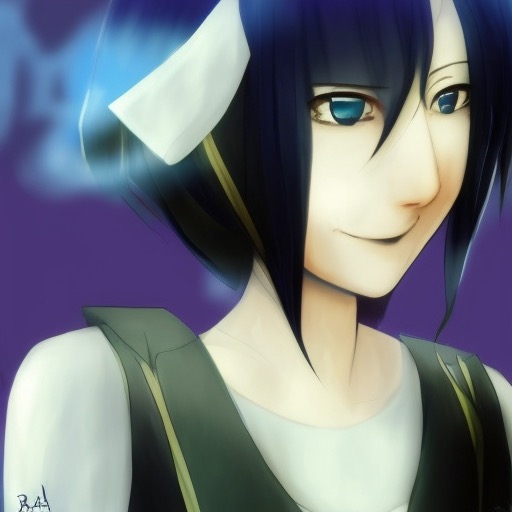}
    \end{minipage}%
    \begin{minipage}[t]{0.13\textwidth}
        \centering
        \includegraphics[width=\textwidth]{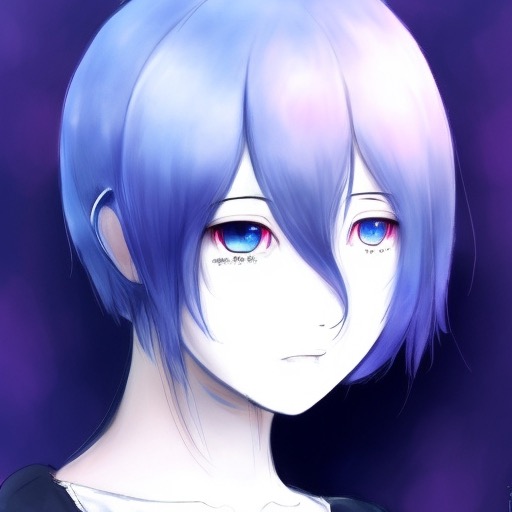}
    \end{minipage}%
    \begin{minipage}[t]{0.13\textwidth}
        \centering
        \includegraphics[width=\textwidth]{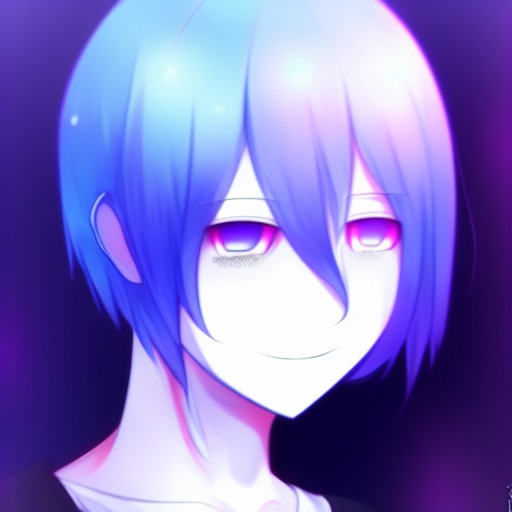}
    \end{minipage}%
    \begin{minipage}[t]{0.13\textwidth}
        \centering
        \includegraphics[width=\textwidth]{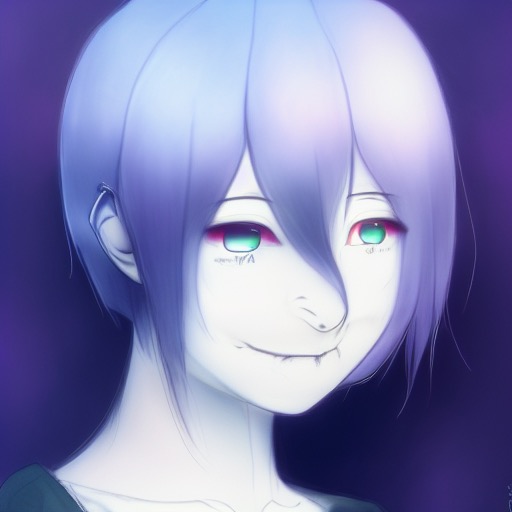}
    \end{minipage}%
    \begin{minipage}[t]{0.13\textwidth}
        \centering
        \includegraphics[width=\textwidth]{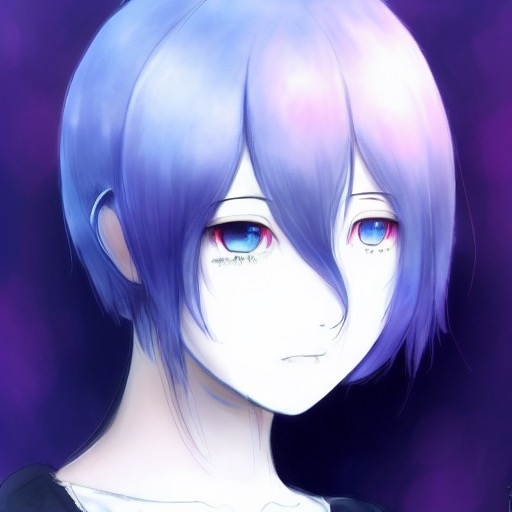}
    \end{minipage}%
    \begin{minipage}[t]{0.13\textwidth}
        \centering
        \includegraphics[width=\textwidth]{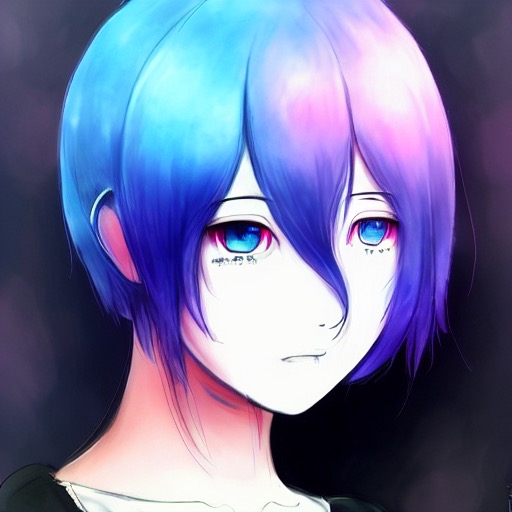}
    \end{minipage}%
    \par

    \begin{minipage}[t]{0.13\textwidth}
        \centering
        \includegraphics[width=\textwidth]{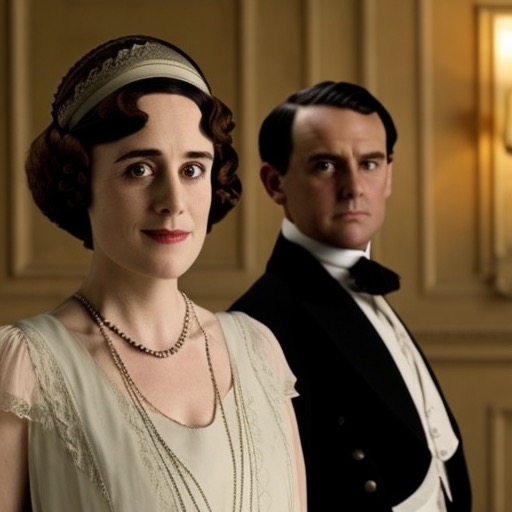}
    \end{minipage}%
    \begin{minipage}[t]{0.13\textwidth}
        \centering
        \includegraphics[width=\textwidth]{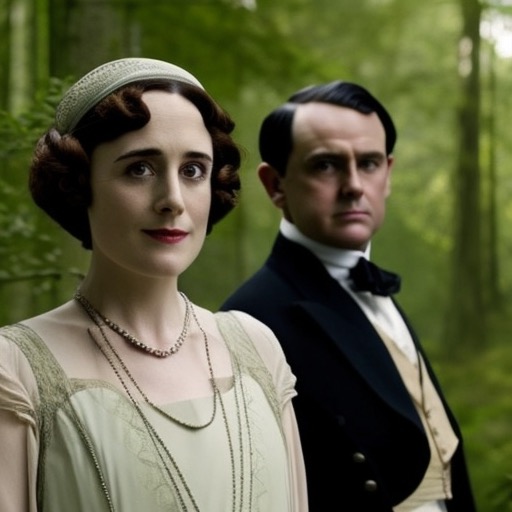}
    \end{minipage}%
    \begin{minipage}[t]{0.13\textwidth}
        \centering
        \includegraphics[width=\textwidth]{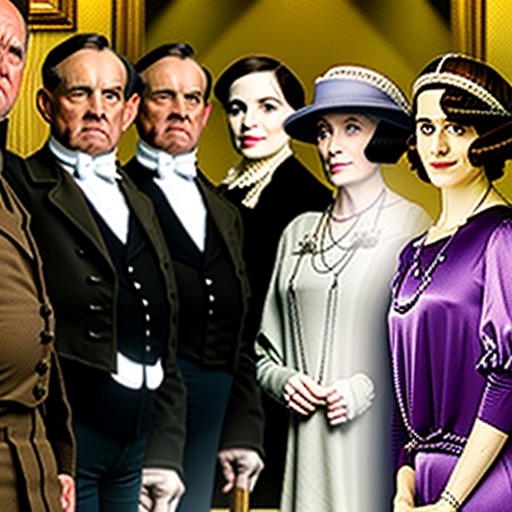}
    \end{minipage}%
    \begin{minipage}[t]{0.13\textwidth}
        \centering
        \includegraphics[width=\textwidth]{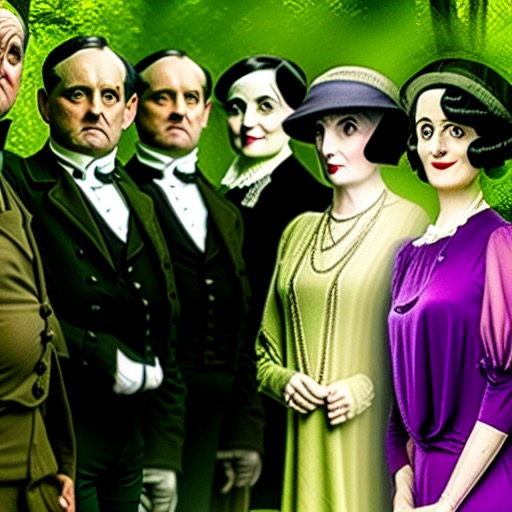}
    \end{minipage}%
    \begin{minipage}[t]{0.13\textwidth}
        \centering
        \includegraphics[width=\textwidth]{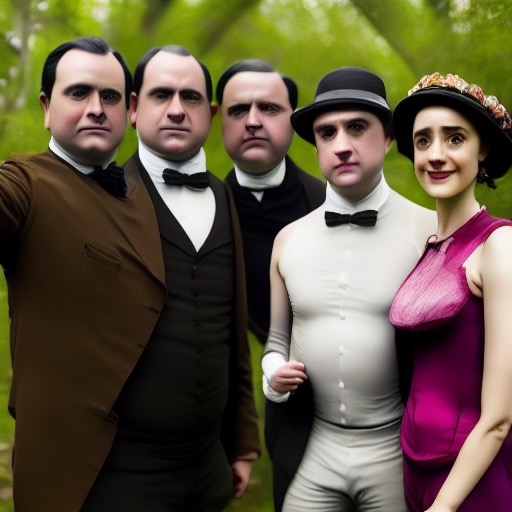}
    \end{minipage}%
    \begin{minipage}[t]{0.13\textwidth}
        \centering
        \includegraphics[width=\textwidth]{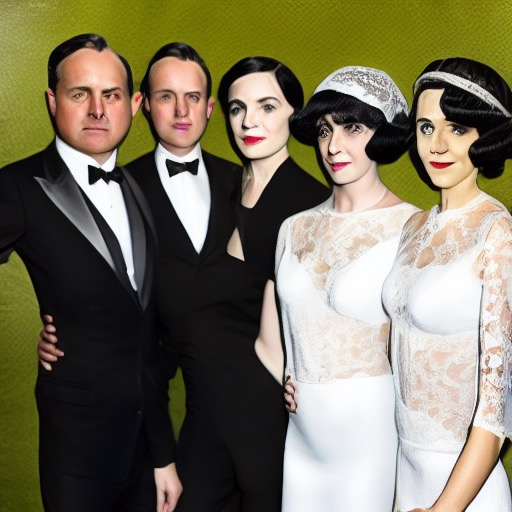}
    \end{minipage}%
    \begin{minipage}[t]{0.13\textwidth}
        \centering
        \includegraphics[width=\textwidth]{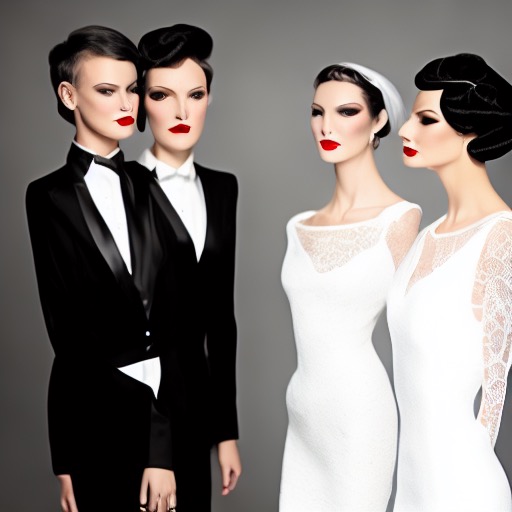}
    \end{minipage}%
    \par

    \begin{minipage}[t]{0.13\textwidth}
        \centering
        \includegraphics[width=\textwidth]{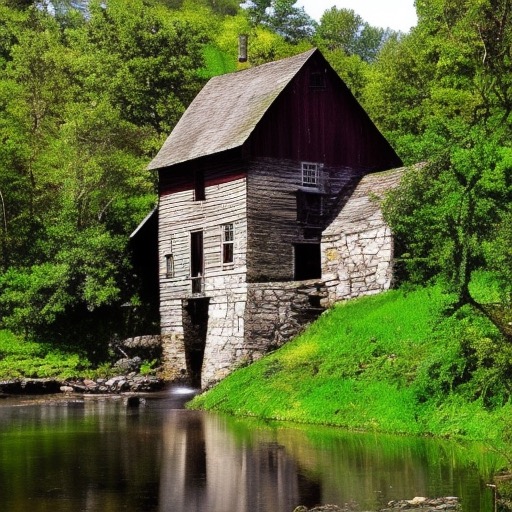}
    \end{minipage}%
    \begin{minipage}[t]{0.13\textwidth}
        \centering
        \includegraphics[width=\textwidth]{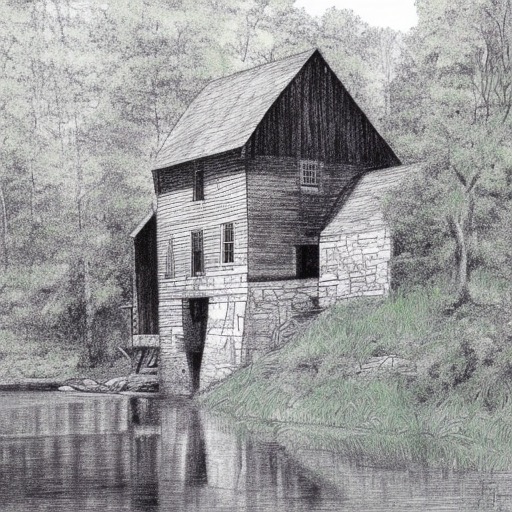}
    \end{minipage}%
    \begin{minipage}[t]{0.13\textwidth}
        \centering
        \includegraphics[width=\textwidth]{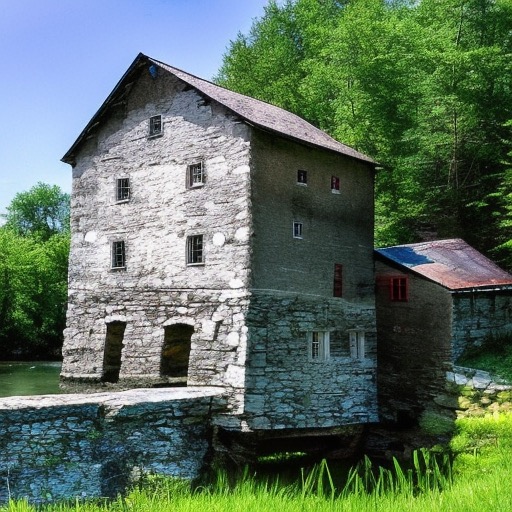}
    \end{minipage}%
    \begin{minipage}[t]{0.13\textwidth}
        \centering
        \includegraphics[width=\textwidth]{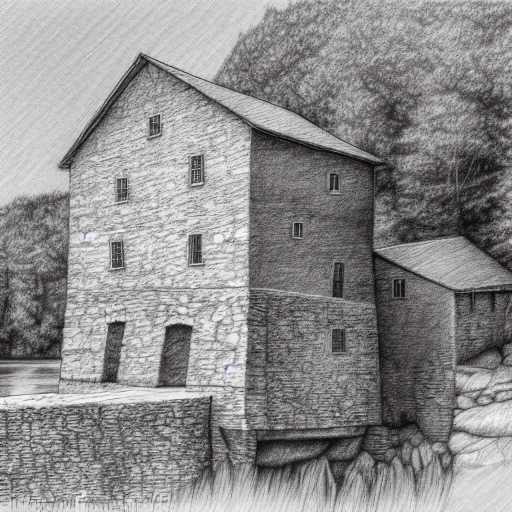}
    \end{minipage}%
    \begin{minipage}[t]{0.13\textwidth}
        \centering
        \includegraphics[width=\textwidth]{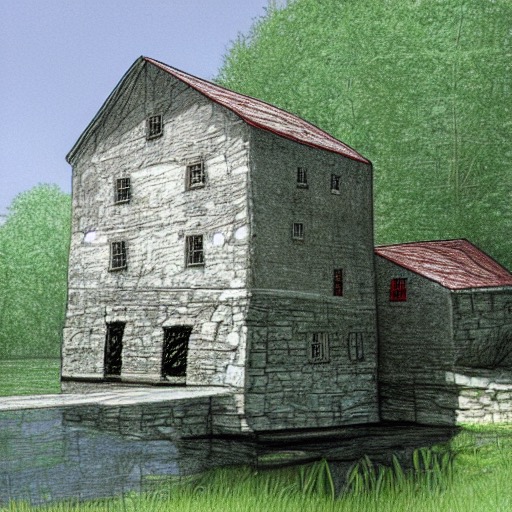}
    \end{minipage}%
    \begin{minipage}[t]{0.13\textwidth}
        \centering
        \includegraphics[width=\textwidth]{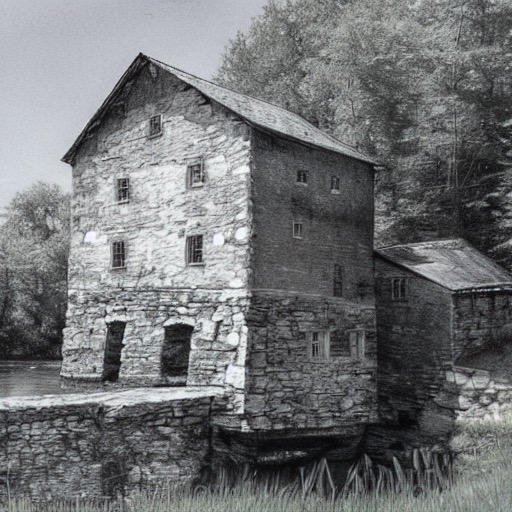}
    \end{minipage}%
    \begin{minipage}[t]{0.13\textwidth}
        \centering
        \includegraphics[width=\textwidth]{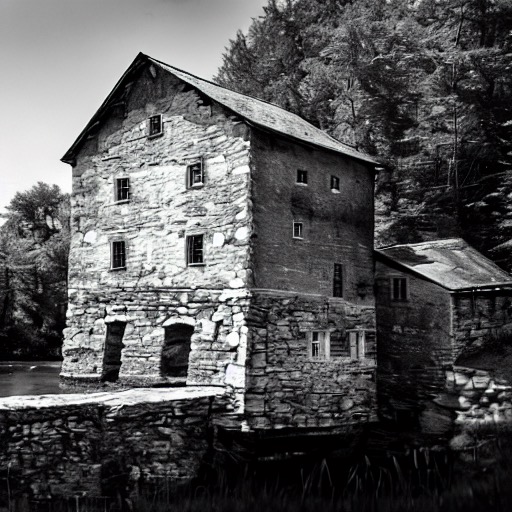}
    \end{minipage}%
    \par
    \caption{Overview of additional qualitative comparisons: We show additional results across different edit types. \method clearly outperforms the baselines consistently, by both maintaining the structure of the test image $y$ and being faithful to the edit illustrated in the exemplar pair. View at high magnification to observe subtle edits.\label{fig:addn-qual2}}
\end{figure*}

\end{document}